\pdfoutput=1
\documentclass[10pt,twocolumn,letterpaper]{article}

\usepackage[pagenumbers]{cvpr}
\usepackage{xcolor}

\usepackage{graphicx}
\usepackage{amsmath}
\usepackage{amssymb}
\usepackage{booktabs}
\usepackage{xcolor}
\usepackage{multirow}
\usepackage{tikz}       %
\usepackage{bm}

\makeatletter
\renewcommand\paragraph{\@startsection{paragraph}{4}{\z@}{0pt}{-1em}%
  {\normalfont\normalsize\bfseries}}
\makeatother

\DeclareMathAlphabet{\mathsfit}{\encodingdefault}{\sfdefault}{m}{sl}
\SetMathAlphabet{\mathsfit}{bold}{\encodingdefault}{\sfdefault}{bx}{n}
\newcommand{\R}{{\mathbb{R}}}
\newcommand{\E}{{\mathbb{E}}}
\newcommand{\sT}{{\mathbb{T}}}       %
\newcommand{\tT}{{\mathbf{T}}}       %
\newcommand{\tD}{{\mathbf{D}}}       %
\newcommand{\tS}{{\mathbf{S}}}       %
\newcommand{\mB}{{\bm{B}}}           %
\newcommand{\mC}{{\bm{C}}}           %
\newcommand{\mH}{{\bm{H}}}           %
\newcommand{\mQ}{{\bm{Q}}}
\newcommand{\mK}{{\bm{K}}}
\newcommand{\mV}{{\bm{V}}}
\newcommand{\mW}{{\bm{W}}}
\newcommand{\mI}{{\bm{I}}}
\newcommand{\vx}{{\bm{x}}}           %
\newcommand{\vz}{{\bm{z}}}           %
\newcommand{\vv}{{\bm{v}}}           %
\newcommand{\veps}{{\bm{\epsilon}}}  %
\newcommand{\vtheta}{{\bm{\theta}}}
\newcommand{\vzero}{{\bm{0}}}
\newcommand{\vone}{{\bm{1}}}            %

\newlength{\tw}
\newcommand{\tsrc}[1]{\begin{picture}(1,1)%
  \put(0,0.5){\includegraphics[width=0.5\tw]{figures/teaser/#1_src0.jpg}}\put(0.5,0.5){\includegraphics[width=0.5\tw]{figures/teaser/#1_src1.jpg}}%
  \put(0,0){\includegraphics[width=0.5\tw]{figures/teaser/#1_src2.jpg}}\put(0.5,0){\includegraphics[width=0.5\tw]{figures/teaser/#1_src3.jpg}}\end{picture}}
\newcommand{\tres}[2]{\begin{picture}(1,1)%
  \put(0,0){\includegraphics[width=\tw]{figures/teaser/#1_res#2.jpg}}\put(0,0){\includegraphics[width=0.32\tw]{figures/teaser/#1_drv#2.jpg}}\end{picture}}
\newsavebox{\grpbox}
\newcommand{\grpgraphics}[2][]{\leavevmode
  \sbox{\grpbox}{\includegraphics[#1]{#2}}%
  \pdfxform attr{/Group <</S /Transparency /I false /K false>>}\grpbox
  \pdfrefxform\pdflastxform}

\definecolor{cvprblue}{rgb}{0.21,0.49,0.74}
\usepackage[pagebackref,breaklinks,colorlinks,allcolors=cvprblue]{hyperref}
\hypersetup{pdftitle={MYSTIQUE: Few-Shot Facial Identity Transfer with Performance Preservation},
  pdfauthor={Shihan Cheng, David Hyde, Junting Deng, Michael Bao, Paul Debevec, Roshni Cooper, Soshi Shimada}}

\usepackage{capt-of}

\def\confName{CVPR}
\def\confYear{2027}

\title{MYSTIQUE: Few-Shot Facial Identity Transfer with Performance Preservation}

\author{
Shihan Cheng$^{1}$\quad David Hyde$^{1}$\quad Junting Deng$^{2}$\quad Michael Bao$^{3}$\\
Paul Debevec$^{3}$\quad Roshni Cooper$^{3}$\quad Soshi Shimada$^{3}$\\[0.3em]
$^{1}$Vanderbilt University\qquad $^{2}$Carnegie Mellon University\qquad $^{3}$Netflix\\[0.1em]
{\tt\small \char123shihan.cheng, david.hyde.1\char125@vanderbilt.edu\quad juntingd@andrew.cmu.edu}\\
{\tt\small \char123mbao, debevec, roshnic, soshis\char125@netflix.com}
}

\begin{document}

\twocolumn[{%
  \renewcommand\twocolumn[1][]{##1}%
  \maketitle
  \begin{center}
    \setlength{\tw}{0.1215\linewidth}\setlength{\unitlength}{\tw}
    \setlength{\tabcolsep}{0.4pt}\renewcommand{\arraystretch}{1}
    \begin{tabular}{@{}cccc@{\hspace{3pt}}cccc@{}}
      \tsrc{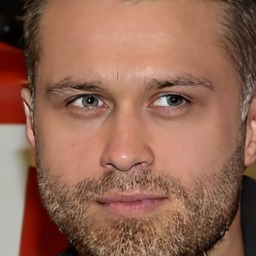} & \tres{lg}{1} & \tres{lg}{2} & \tres{lg}{3} & \tsrc{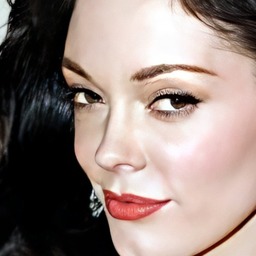} & \tres{oc}{1} & \tres{oc}{2} & \tres{oc}{3} \\[0.8pt]
      \tsrc{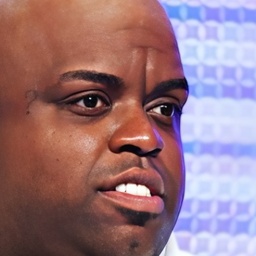} & \tres{ex}{1} & \tres{ex}{2} & \tres{ex}{3} & \tsrc{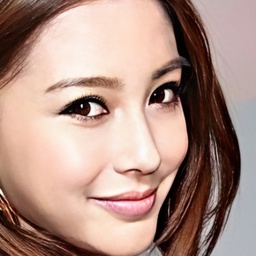} & \tres{hr}{1} & \tres{hr}{2} & \tres{hr}{3}
    \end{tabular}
    \captionof{figure}{\textbf{Identity transfer with MYSTIQUE.} MYSTIQUE takes the identity from the target's photos and keeps
    the expression, lighting and occluders of the driving image. Users can also edit the image during the transfer.
    Each group shows four of the target's photos on the left and three results on the right, each with its driving image
    as an inset. The examples cover hard colored and directional lighting (top left), occluders such as flowers and a camera
    (top right), different expressions, also from people of another age or sex (bottom left), and edits of the hair and
    accessories (bottom right).}
    \label{fig:teaser}
  \end{center}
  \vspace{0.5em}
}]
\begin{abstract}
In visual effects, identity transfer replaces the identity in a driving image with
that of a target person, while preserving the expression, lighting, pose and scene of that image. Paired
ground truth is rarely available, so many methods are trained by
self-reconstruction, where the model can learn a shortcut by copying identity
cues from the driving image. This shortcut causes identity leakage at
inference, where the driving image shows a different person. Our key
observation is that leakage can enter through any input that reads the driving
image. We therefore present MYSTIQUE, a diffusion-based identity transfer method that takes
identity from a set of the target's photos and reads expression, lighting and
scene through separate streams. It learns to disentangle identity from the
other streams through multi-stage training, with disentanglement strategies at
their inputs: a random block mask whose boundaries do not follow the face
outline, self-augmentation, which randomly changes the face shape in the
expression input and keeps its expression, and face randomization, which
replaces the person in the expression and lighting inputs while the ground
truth stays the driving image. At inference, we also warp the lighting input
onto the face shape of a first pass, which reduces face-shape leakage through
the lighting stream. Since face-recognition scores and human
judgments each have limits, we evaluate identity with both, together
with face-shape metrics on rendered faces of known shape. In a user study,
raters judge MYSTIQUE's results more like the target than those of each compared
method. Among these methods, MYSTIQUE also recovers the target's face
shape most accurately, and its recognition score gains the most as more photos
of the target are given.

\end{abstract}

\section{Introduction}
\label{sec:intro}
Facial identity transfer replaces the identity in a \emph{driving image} with that of a
\emph{target} person while preserving the pose, expression, gaze, lighting,
hair, occluders and background of the driving image. In visual effects (VFX)
for film and television, applications include stunt doubles, reshoots and digital
doubles, where the target actor's likeness must be reproduced while the original
performance is preserved~\cite{naruniec2020vfx,livingswap2026}. Such shots can
contain strong or colored lighting that the transferred face must follow. Many
photos of the target actor are also often available, and combining them can
improve identity fidelity.

In practice, the result often keeps identity cues of the person in the driving
image, which we call \emph{identity leakage}. It comes from how identity transfer
methods are trained. Direct supervision would need images of two different people with the
same expression, lighting and scene, and real-world datasets rarely contain such
pairs. Many feed-forward methods are therefore trained by self-reconstruction,
where the model rebuilds a driving image from photos of the same
person~\cite{yuan2023reliableswap,han2024faceadapter,baliah2025reface,lee2024selfswapper}. Since the driving image already shows the
right identity, the model can learn a shortcut by copying identity cues from the
inputs that read it for expression, lighting and scene. At inference, the
driving image shows a different person, and the same shortcut brings that
person's identity into the result. Yet the model needs these inputs for
the expression and lighting that the result should keep.

Existing methods take identity from a separate input, but their other inputs
still read the driving image and can carry its identity~\cite{ren2023wscswap}. GAN-based methods~\cite{li2020faceshifter,chen2020simswap,inswapper,rosberg2023facedancer,groshev2022ghost}
inject a face-recognition embedding of the target's photo and repaint the inner
face of the driving image. Diffusion-based methods~\cite{zhao2023diffswap,baliah2025reface,han2024faceadapter,ye2025dreamid,guo2025dreamidv}
regenerate the face with a diffusion model conditioned on an identity embedding
of the target's photo. Face-Adapter~\cite{han2024faceadapter} counters this
leakage at inference: its attribute input encodes the whole driving image, so it
uses the identity tokens of the driving image as a negative condition. In our
experiments, the results of these methods, including Face-Adapter, still tend to
follow the jawline of the driving image.

\begin{figure*}[t]
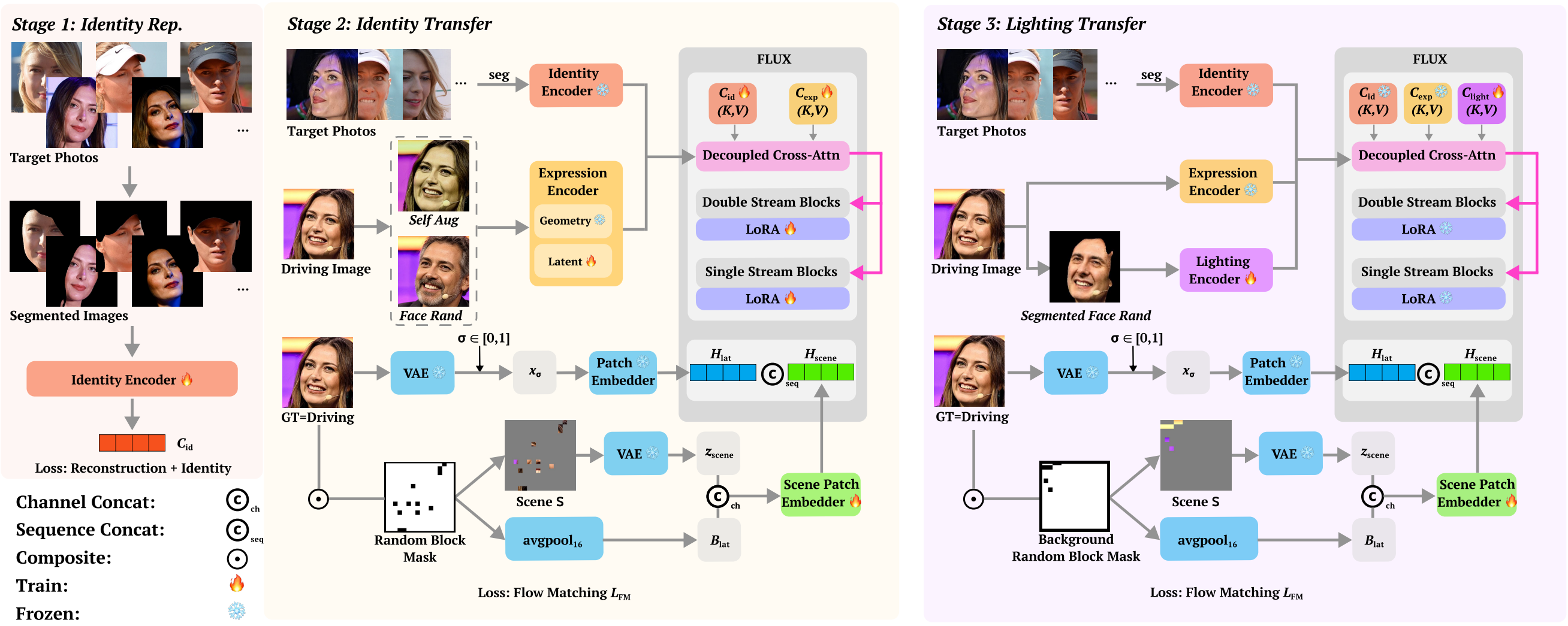

  \centering
  \grpgraphics[width=\linewidth]{figures/fig2_training.pdf}
  \caption{\textbf{Training in three stages.} Stage~1 learns an identity
  representation from photos of a person; its second step, through the frozen
  transformer, is not drawn. Stage~2 learns identity transfer with the identity encoder
  and the geometry branch frozen. Stage~3 learns lighting transfer with the identity
  and expression streams frozen. The identity, expression and lighting tokens
  enter every block through decoupled cross-attention, with the projection of each
  encoder drawn as part of its keys and values $(\mK,\mV)$; the scene tokens
  $\mH_\mathrm{scene}$ are appended to the noisy latent tokens $\mH_\mathrm{lat}$.
  Self Aug, Face Rand, seg, GT: self-augmentation, face randomization, segmentation,
  ground truth.}
  \label{fig:arch}
\end{figure*}

Our key observation is that leakage can enter through any input that reads the
driving image, so we reduce it at each of them. To this end, MYSTIQUE decouples
identity, expression and lighting into separate streams of a frozen diffusion
transformer~\cite{flux2klein}, and the scene stream reads the driving image with
the face masked out (Fig.~\ref{fig:arch}). It then learns to disentangle
identity from the other inputs by multi-stage training. Stage~1 learns an
identity representation in a separate model that reconstructs a person from
their photos, so the identity tokens are independent of the driving image.
Stage~2 learns identity transfer on top of the frozen identity encoder, and
Stage~3 learns lighting transfer.
Stage~2 and Stage~3 reduce leakage at these inputs with
disentanglement strategies. The scene stream reads the driving image through a
random block mask, whose boundaries do not follow the face outline. The
expression stream reads the driving image with its face shape randomly changed
and its expression kept, which we call \emph{self-augmentation}. The expression
and lighting streams also use \emph{face randomization}: their inputs show
another person, while the ground truth stays the driving image, so copying
identity from these inputs helps less. For the expression input, an image
generator invents a person of a different sex, age and facial structure. For the
lighting input, an existing identity transfer model~\cite{inswapper} changes the
identity and keeps the lighting. As its results retain part of the face shape, at
inference we warp the lighting input onto the face shape of a first pass without
lighting.

A single photo shows the target in one pose, expression and lighting, and
reveals little about how the face looks in others. MYSTIQUE therefore reads an
unordered set of the target's photos, and its identity encoder, which follows
the architecture of Omni-ID~\cite{qian2024omniid}, combines them into identity
tokens. In our experiments, added photos raise the recognition score of
MYSTIQUE more than averaging over the photos does in other methods. With less
leakage from the other inputs, the added photos also bring the face shape closer
to the target.
A second problem is that identity is hard to measure.
Whether two faces show the same person is a complex human judgment that draws on face shape, the individual features and how
they are arranged~\cite{maurer2002configural,jozwik2022face}. Face-recognition
embeddings~\cite{deng2019arcface,kim2022adaface} capture part of this judgment.
They can score a result highly even when its mouth and lower face shape differ from
the target~\cite{yuan2023reliableswap}, and they respond to hairstyle and hair color,
which the result should keep from the driving image~\cite{shiohara2023blendface}.
Several works on identity transfer add user studies or new identity
metrics~\cite{li2020faceshifter,yuan2023reliableswap,guo2025dreamidv}. Following
them, we report human preference along with recognition scores, and add
face-shape metrics on rendered faces of known shape.

In summary, our contributions are:
\begin{itemize}\itemsep2pt \parskip0pt \parsep0pt
\item MYSTIQUE, an identity transfer method that learns to disentangle identity from the inputs
  that read the driving image. It learns its streams by multi-stage training, and
  disentanglement strategies reduce leakage at each of these inputs.
\item Identity from a set of the target's photos. Added photos raise the
  recognition score more than averaging does in other methods, and with less
  leakage, they also bring the face shape closer to the target.
\item An identity evaluation by raters and recognizers, with face-shape metrics on
  rendered faces of known shape. In a user study, raters judge MYSTIQUE's
  results more like the target than those of each compared method, and MYSTIQUE
  recovers the target's face shape most accurately among them, without a 3D model
  of the target's face.
\end{itemize}

\section{Related Work}
\label{sec:related}

\paragraph{Identity disentanglement.}
Self-reconstruction lets a model copy identity from the driving image, and prior
work reduces this in two ways. The first builds cross-identity training pairs. The
input that shows another person is made by reenactment and
blending~\cite{yuan2023reliableswap}, by an existing identity transfer
model~\cite{huang2024cscs,shao2024vividface,ye2025dreamid,livingswap2026}, or by a video model
driven by landmarks~\cite{guo2025dreamidv}. When the pairs should also differ in
face shape, a further step is added, such as warping~\cite{huang2024cscs,ye2025dreamid} or 3DMM
retargeting~\cite{guo2025dreamidv}. The second way reduces the identity cues in what
the model reads. BlendFace~\cite{shiohara2023blendface} trains its identity encoder
on images whose attributes are replaced, so that hairstyle and similar attributes
do not affect it. In reenactment, X-NeMo~\cite{zhao2025xnemo} compresses pose
and expression into a 1D latent, trained on driving images with randomly changed
color and face shape, so that it carries little identity.
Omni-ID~\cite{qian2024omniid} keeps a small part of the image to be rebuilt visible
when training its identity encoder, so that this part leaks little identity.
MYSTIQUE uses both ways: the random block mask and self-augmentation follow the
second, and face randomization the first. Face randomization is a form of domain
randomization~\cite{tobin2017domain}: the input shows another person, and the
ground truth stays the driving image. For the expression input, an image generator
invents a person of a different sex, age and facial structure, so the face shape
changes without a further step.

\paragraph{Identity transfer.}
DeepFaceLab~\cite{perov2020deepfacelab} trains one model per pair of people.
Subject-agnostic GAN-based methods inject a face-recognition embedding of the
target into the features of the driving
image~\cite{li2020faceshifter,chen2020simswap,inswapper,rosberg2023facedancer,groshev2022ghost}.
Diffusion-based methods~\cite{kim2022diffface,zhao2023diffswap,baliah2025reface,han2024faceadapter,ye2025dreamid}
followed, several of which cast the task as conditional
inpainting~\cite{zhao2023diffswap,baliah2025reface,han2024faceadapter}, and recent
ones extend it to video with diffusion
transformers~\cite{guo2025dreamidv,livingswap2026}. ID-Constrained~\cite{he2026idconstrained}
trains with identity conditions alone before adding the others; MYSTIQUE learns
identity in a separate model. To control face shape, many of
these methods add explicit geometry of the target, such as 3DMM coefficients or 3D
landmarks~\cite{wang2021hififace,zhao2023diffswap,han2024faceadapter,shao2024vividface,wang2025dynamicface}.
MYSTIQUE uses no 3D geometry of the target and takes face shape from its photos.
Several photos of the target are used by multi-reference identity
transfer~\cite{wu2020particularity}, by a patented
StyleGAN-based method~\cite{kim2025multisource}, which attends to them to be
robust to poor ones, and by GSwap~\cite{gswap2026}, which replaces the whole head and optimizes a
model per subject and per video. The embedding of a GAN-based method can also be averaged
over photos, as we do for the baselines in Sec.~\ref{sec:exp}. Outside identity
transfer, photo sets are common: personalization methods
average recognition embeddings~\cite{wang2024instantid,papantoniou2024arc2face} or
stack image embeddings~\cite{li2024photomaker}, few-shot reenactment pools
several frames of the subject~\cite{zakharov2019fewshot,ha2020marionette},
reference-based restoration attends to several photos of the
subject~\cite{hsiao2024refldm,liu2025faceme}, and Omni-ID~\cite{qian2024omniid}
encodes a photo set into a fixed number of tokens. MYSTIQUE reads the target's
photos as a set. With less leakage from the other inputs, added photos also bring
the face shape closer to the target.

\section{Method}
\label{sec:method}

MYSTIQUE reads the target's photos and the driving image through separate streams
of a frozen diffusion transformer (Sec.~\ref{sec:method:arch}) and trains them in
three stages, so that the identity tokens are learned from photos of a person alone
(Sec.~\ref{sec:method:training}). Because the other streams read the driving image,
three disentanglement strategies (Sec.~\ref{sec:method:strategies}) and two-pass
inference (Sec.~\ref{sec:method:inference}) reduce the identity leakage through
them.

\subsection{Architecture}
\label{sec:method:arch}
\begin{figure*}[t]
  \centering
  \input{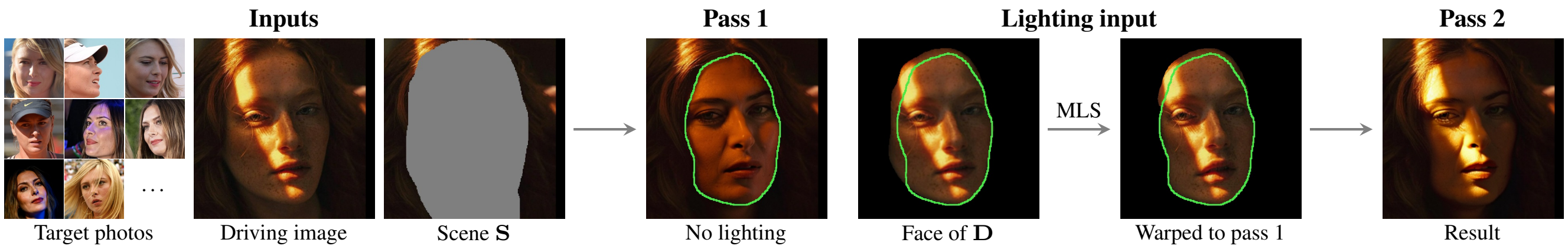}
  \caption{\textbf{Two-pass inference.} Pass~1 runs without the lighting
  stream. The face of the driving image is warped to the face shape of the
  pass-1 result (green outline), except at the hairline, and serves as the lighting input of pass~2,
  which otherwise repeats pass~1.}
  \label{fig:inference}
\end{figure*}
Given a set $\sT=\{\tT_i\}_{i=1}^{M}$ of $M$ photos of the target and a driving
image $\tD$, MYSTIQUE generates an image that shows the target's identity with the
pose, expression, lighting and scene of $\tD$. All images are face crops: each
photo is resized to $\tT_i\in\R^{448\times448\times3}$, and the driving image to
$\tD\in\R^{H\times W\times3}$, which is also the resolution of the result
($H=W=256$; Supp.~\ref{sec:suppl:allmetrics} and~\ref{sec:suppl:shape} report a $512$\,px model). We formulate the task as
inpainting: a binary mask $\mB\in\{0,1\}^{H\times W}$ marks the face region of
$\tD$, which the model removes and generates again. The generation runs in the
FLUX.2 Klein 9B diffusion transformer~\cite{flux2klein}, whose pretrained weights
stay frozen; a low-rank update~\cite{hu2022lora} on the attention projections of
every block adapts it to identity transfer. Fig.~\ref{fig:arch} shows the
architecture in each training stage.

\paragraph{Encoders.}
Three encoders turn the inputs into tokens of width $d_\mathrm{c}=1280$. The
identity encoder reads the $M$ photos of the target, each masked by a
face-parsing model~\cite{yu2018bisenet} to the face, ears and neck. These are the
regions that the model replaces at inference, so hair does not enter the identity
tokens $\mC_\mathrm{id}\in\R^{256\times d_\mathrm{c}}$. Their number does not
depend on $M$, so $M$ can be chosen freely at inference. The expression encoder
reads the driving image through two branches. The geometry branch predicts $128$
implicit 3D keypoints of the face, following the principle of
LivePortrait~\cite{guo2024liveportrait}, in a separate model that we pretrain
ourselves (Supp.~\ref{sec:suppl:impl:arch}) and keep frozen; a linear layer maps the 3D
position of each keypoint to one token. The latent branch, the expression encoder
of X-NeMo~\cite{zhao2025xnemo}, reads a latent expression descriptor from the face
image, which linear layers expand into $128$ tokens of width $d_\mathrm{c}$;
Stage~2 fine-tunes it from its pretrained weights. Layer normalization and
averaging fuse the two branches into $\mC_\mathrm{exp}\in\R^{128\times d_\mathrm{c}}$,
which follows expression more closely than either branch alone
(Sec.~\ref{sec:exp:ablations}). The lighting encoder reads the driving image
masked to the skin, brows, eyes, nose, mouth and glasses, without the neck, ears or
hair. A small convolutional network maps it to a $16\times16$ grid, whose cells
become the lighting tokens $\mC_\mathrm{light}\in\R^{256\times d_\mathrm{c}}$,
each aligned with one latent token.

\paragraph{Conditioning.}
The four streams enter the transformer in two ways: the identity, expression and
lighting tokens through cross-attention, and the scene as part of the token
sequence. The transformer denoises a latent of the FLUX.2 Klein autoencoder, a
$16\times16$ grid with $128$ channels, which it reads as $N=256$ tokens with hidden
states $\mH_\mathrm{lat}\in\R^{N\times d}$, $d=4096$. As in
IP-Adapter~\cite{ye2023ipadapter}, each token stream
$k\in\{\mathrm{id},\mathrm{exp},\mathrm{light}\}$ has a projection
$\phi_k:\R^{d_\mathrm{c}}\to\R^{d}$, shared by all blocks, and in each block its own
keys $\mK_k=\phi_k(\mC_k)\mW_\mathrm{K}^k$ and values
$\mV_k=\phi_k(\mC_k)\mW_\mathrm{V}^k$, with
$\mW_\mathrm{K}^k,\mW_\mathrm{V}^k\in\R^{d\times d}$. The attention outputs of the
streams are added to the hidden states of the noisy latent:
\begin{equation}
  \mH_\mathrm{lat} \;\leftarrow\; \mH_\mathrm{lat} + \sum_{k} \operatorname{Attn}(\mQ,\mK_k,\mV_k),
  \label{eq:attn}
\end{equation}
where $\mQ\in\R^{N\times d}$ are the queries of the noisy latent tokens in that
block. For the lighting stream, rotary position embeddings~\cite{su2024roformer}
place each key at the center of its cell and each query at its own position, so
the attention depends on relative position and each output location can attend
to the lighting at the same location of the input. The scene
$\tS=(\vone-\mB)\odot\tD$ is the driving image with the face removed. The
autoencoder encodes it into $\vz_\mathrm{scene}$, which is concatenated with
$\mB_\mathrm{lat}$, the mask $\mB$ average-pooled to the latent resolution. A
trainable copy of the patch embedding, the scene patch embedding, maps the result to
scene tokens $\mH_\mathrm{scene}\in\R^{N\times d}$. These tokens are appended to the
noisy latent tokens along the sequence and share their spatial positions, so the
scene stays aligned with the result. A fixed text prompt, given in Supp.~\ref{sec:suppl:impl:arch},
completes the sequence.

\subsection{Multi-stage training}
\label{sec:method:training}
\begin{figure*}[t]
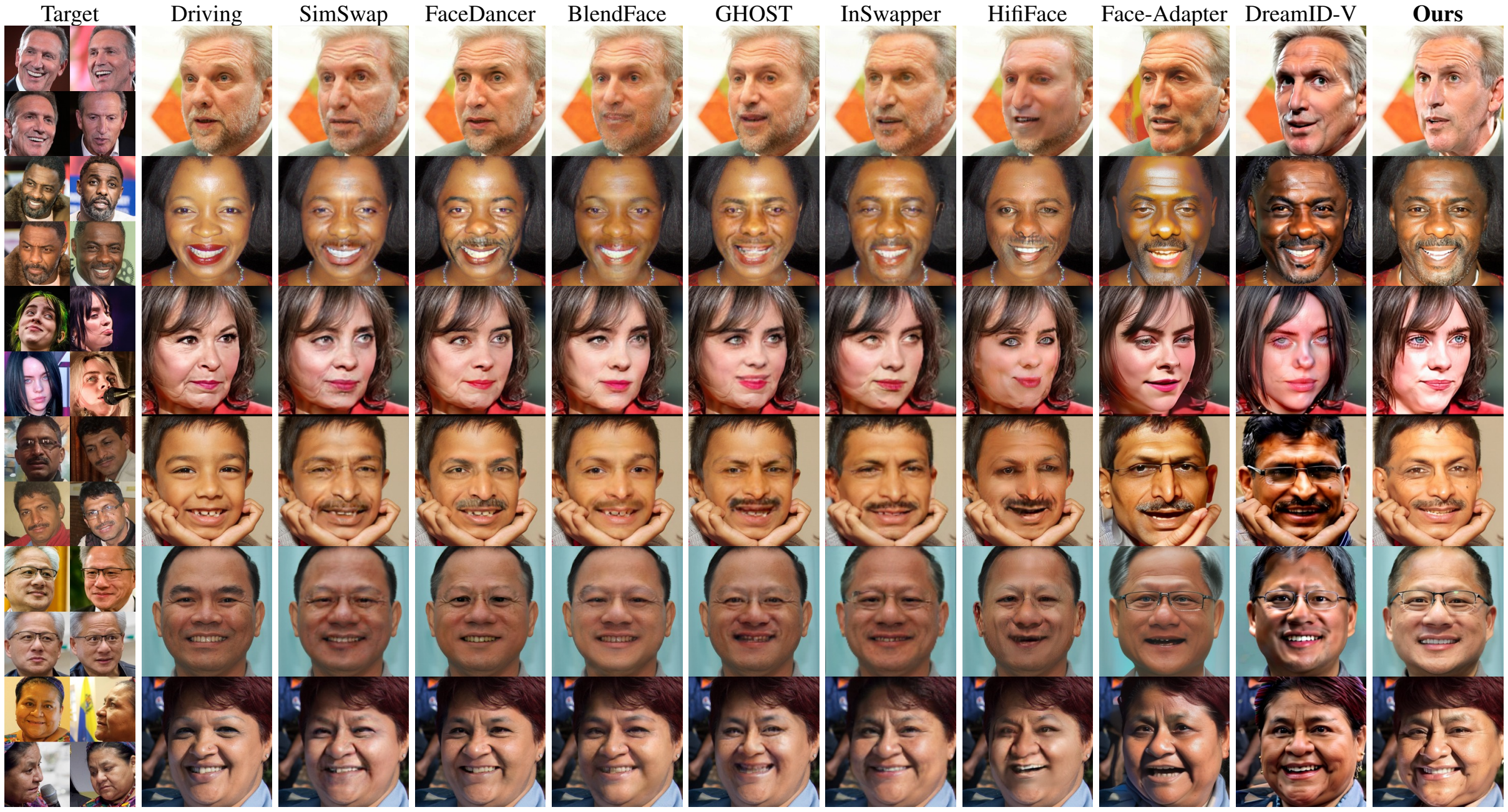

  \centering
  \def\bcw{0.0790\linewidth}
  \newcommand{\bcell}[2]{\includegraphics[width=\bcw]{figures/celeb/#1/#2.jpg}}
  \newcommand{\bhead}[1]{\makebox[\bcw][c]{\scriptsize #1}}
  \setlength{\tabcolsep}{1pt}\renewcommand{\arraystretch}{0}
  \begin{tabular}{@{}ccccccccccc@{}}
    \bhead{Target} & \bhead{Driving} & \bhead{SimSwap} & \bhead{FaceDancer} & \bhead{BlendFace} & \bhead{GHOST} & \bhead{InSwapper} & \bhead{HifiFace} & \bhead{Face-Adapter} & \bhead{DreamID-V} & \bhead{\textbf{Ours}} \\[1pt]
    \bcell{c1}{ref} & \bcell{c1}{driver} & \bcell{c1}{simswap} & \bcell{c1}{facedancer} & \bcell{c1}{blendface} & \bcell{c1}{ghost} & \bcell{c1}{inswapper} & \bcell{c1}{hififace} & \bcell{c1}{faceadapter} & \bcell{c1}{dreamidv} & \bcell{c1}{ours} \\
    \bcell{c2}{ref} & \bcell{c2}{driver} & \bcell{c2}{simswap} & \bcell{c2}{facedancer} & \bcell{c2}{blendface} & \bcell{c2}{ghost} & \bcell{c2}{inswapper} & \bcell{c2}{hififace} & \bcell{c2}{faceadapter} & \bcell{c2}{dreamidv} & \bcell{c2}{ours} \\
    \bcell{c3}{ref} & \bcell{c3}{driver} & \bcell{c3}{simswap} & \bcell{c3}{facedancer} & \bcell{c3}{blendface} & \bcell{c3}{ghost} & \bcell{c3}{inswapper} & \bcell{c3}{hififace} & \bcell{c3}{faceadapter} & \bcell{c3}{dreamidv} & \bcell{c3}{ours} \\
    \bcell{c4}{ref} & \bcell{c4}{driver} & \bcell{c4}{simswap} & \bcell{c4}{facedancer} & \bcell{c4}{blendface} & \bcell{c4}{ghost} & \bcell{c4}{inswapper} & \bcell{c4}{hififace} & \bcell{c4}{faceadapter} & \bcell{c4}{dreamidv} & \bcell{c4}{ours} \\
    \bcell{c5}{ref} & \bcell{c5}{driver} & \bcell{c5}{simswap} & \bcell{c5}{facedancer} & \bcell{c5}{blendface} & \bcell{c5}{ghost} & \bcell{c5}{inswapper} & \bcell{c5}{hififace} & \bcell{c5}{faceadapter} & \bcell{c5}{dreamidv} & \bcell{c5}{ours} \\
    \bcell{c6}{ref} & \bcell{c6}{driver} & \bcell{c6}{simswap} & \bcell{c6}{facedancer} & \bcell{c6}{blendface} & \bcell{c6}{ghost} & \bcell{c6}{inswapper} & \bcell{c6}{hififace} & \bcell{c6}{faceadapter} & \bcell{c6}{dreamidv} & \bcell{c6}{ours} \\
  \end{tabular}
  \caption{\textbf{Comparison with prior work.} Each row shows four photos of the
  target, the driving image and the results. MYSTIQUE's results look more like the
  target, including its face shape, and keep the hair,
  expression, lighting and background of the driving image.}
  \label{fig:baselines}
\end{figure*}
\label{sec:method:curriculum}
We train MYSTIQUE by self-reconstruction: the driving image and the target's
photos are different photos of the same person, and the model rebuilds the driving
image, which serves as the ground truth.
Let $\vx_0\in\R^{n}$ be the flattened latent of the ground
truth, with $n=16\cdot16\cdot128$. With $\veps\sim\mathcal{N}(\vzero,\mI)$, a noise
level $\sigma\in[0,1]$ drawn from a logit-normal distribution~\cite{esser2024sd3},
and $\vx_\sigma=(1-\sigma)\vx_0+\sigma\veps$, Stages~2 and~3 minimize the
flow-matching loss~\cite{lipman2023flowmatching,esser2024sd3}
\begin{equation}
  L_\mathrm{FM} \;=\; \E_{\vx_0,\veps,\sigma}
  \big\lVert \vv_\vtheta(\vx_\sigma,\sigma) - (\veps - \vx_0) \big\rVert_2^2 ,
  \label{eq:fm}
\end{equation}
where $\vv_\vtheta$ is the velocity that the model predicts from the noisy latent
and the streams, and $\vtheta$ are its trainable parameters. The loss covers
the noisy latent tokens only.

\label{sec:method:identity}
The three stages of Fig.~\ref{fig:arch} each start from the model of the previous
stage. \textbf{Stage~1} learns an identity representation that does not depend on
the driving image. Following the architecture of Omni-ID~\cite{qian2024omniid}, we
train our own identity encoder to reconstruct photos of a person through a masked
decoder. Besides the reconstruction losses, an identity loss compares the
AdaFace~\cite{kim2022adaface} embeddings of each reconstruction and of its unmasked
photo by cosine similarity. We then train the encoder further through the frozen
transformer with the flow-matching loss of Eq.~\ref{eq:fm}: the transformer
generates other photos of the person from the identity tokens, and this step also
trains the cross-attention of the identity stream. Since no driving image enters
Stage~1, the identity tokens learn to carry the identity, including the face shape,
without help from a driving image.

\textbf{Stage~2} learns identity transfer on top of this representation. The identity
encoder and the geometry branch stay frozen, while the latent branch, the
projections and cross-attention of the identity and expression streams, the scene patch
embedding and the low-rank update are trained. Each sample draws the number of
photos $M$ at random between $1$ and $24$, so the model learns to use any number of
photos. Stage~2 uses no lighting stream.

\textbf{Stage~3} then learns lighting
transfer: it adds the lighting stream and trains it together with the scene patch
embedding, while the identity and expression streams and the low-rank update stay
frozen, so this stage does not change what they encode.

\subsection{Disentanglement strategies}
\label{sec:method:strategies}
\label{sec:method:performance}
\label{sec:method:leak}
\label{sec:method:mask}
The expression, lighting and scene streams read the driving image, which in
self-reconstruction is the ground truth itself. Each of these inputs
can therefore pass identity cues of the ground truth to the result: the geometry
branch encodes its face shape, a mask that follows the face reveals its outline to
the scene stream, and the lighting input shows the face itself. Copying these cues is a
shortcut that lowers the training loss, but at inference it brings the face shape
of the person in the driving image into the result. Stages~2 and~3 reduce this leakage with three
disentanglement strategies.

\paragraph{Random block mask.}
Inpainting-based methods usually generate the face inside a face mask, both in
training~\cite{baliah2025reface} and at inference, and the outline of such a mask passes the face
shape of the driving image on to the result~\cite{han2024faceadapter}. REFace and
SelfSwapper~\cite{lee2024selfswapper} perturb this mask in training. Stage~2
instead trains with a random block mask $\mB$, which hides a random
$95$--$99\%$ of the $16\times16$-pixel blocks of the driving image, one block per
latent token. The hidden region does not follow the face outline, so the mask
carries little information about the face shape. In Stage~3, the visible blocks are
drawn only from the background, so the scene does not show the lighting on the
person, and the model reads it from the lighting stream.

\paragraph{Self-augmentation.}
The expression stream should pass on the expression and pose of the driving image,
but not its face shape. Stage~2 therefore gives the expression encoder a copy of the
driving image whose face shape is randomly changed and whose expression is kept:
moving least squares~\cite{schaefer2006mls} moves the face outline by up to $12\%$,
and the aspect ratio of the face is scaled by a random factor in $[0.8,1.25]$. The
ground truth stays the original driving image. Copying the face shape of the
expression input then helps less in rebuilding the ground truth, while following
its expression and pose still helps.
\begin{table}[t]
  \centering\footnotesize
  \setlength{\tabcolsep}{2.6pt}
  \caption{\textbf{Comparison on the identity benchmark} ($240$ pairs). Each
  method runs at its best setting (Sec.~\ref{sec:exp:setup}). \emph{Pref.}:
  share of identity judgements that prefer our result over the row's (\%),
  from the user study (Sec.~\ref{sec:exp:eval}).
  \emph{ArcFace}: a real photo of the target scores $0.677$;
  italic values exceed it and are not ranked. Bold: best.}
  \label{tab:main}
  \begin{tabular}{@{}l c c ccc c@{}}
    \toprule
    Method & Pref. & ArcFace$\uparrow$ & Expr.$\downarrow$ & Pose$\downarrow$ & Light$\uparrow$ & FaceScore$\uparrow$ \\
    \midrule
    SimSwap~\cite{chen2020simswap}          & 97  & 0.500 & 0.243 & \textbf{2.03} & \textbf{0.96} & 3.67 \\
    FaceDancer~\cite{rosberg2023facedancer} & 81  & 0.394 & \textbf{0.205} & 2.38 & \textbf{0.96} & 4.39 \\
    BlendFace~\cite{shiohara2023blendface}  & 92  & 0.487 & 0.231 & 4.42 & 0.90 & 3.73 \\
    GHOST~\cite{groshev2022ghost}           & 99  & 0.548 & 0.337 & 4.60 & 0.93 & 4.02 \\
    InSwapper~\cite{inswapper}              & 98  & \textit{0.715} & 0.258 & 2.70 & 0.93 & 3.92 \\
    HifiFace~\cite{wang2021hififace}        & 100 & 0.528 & 0.345 & 3.14 & 0.90 & 4.02 \\ %
    Face-Adapter~\cite{han2024faceadapter}  & 73  & 0.501 & 0.332 & 4.38 & 0.83 & 4.09 \\
    DreamID-V~\cite{guo2025dreamidv}        & 79  & 0.550 & 0.372 & 7.52 & 0.52 & 4.42 \\
    \midrule
    \textbf{Ours}                           & --  & \textbf{0.601} & 0.253 & 3.99 & 0.83 & \textbf{4.62} \\
    \bottomrule
  \end{tabular}

  \vspace{8pt}
  {\setlength{\tabcolsep}{3.2pt}
  \caption{\textbf{Face shape on the shape benchmark} ($348$ pairs). Metrics as
  in Sec.~\ref{sec:exp:eval}. Each method receives $24$ renders of the target and
  runs at its setting of Table~\ref{tab:main}. Bold: best. \emph{Driving image}:
  the driving image itself, scored as a result.}
  \label{tab:shape}
  \begin{tabular}{@{}lccc@{\hspace{10pt}}lccc@{}}
    \toprule
    Method & $s_\mathrm{gt}\downarrow$ & Leak$\downarrow$ & 3D$\downarrow$ & Method & $s_\mathrm{gt}\downarrow$ & Leak$\downarrow$ & 3D$\downarrow$ \\
    \midrule
    SimSwap    & 0.125 & 0.95 & 1.83 & InSwapper    & 0.116 & 0.81 & 1.48 \\
    FaceDancer & 0.122 & 0.91 & 1.85 & HifiFace     & 0.138 & 0.78 & 2.03 \\
    BlendFace  & 0.119 & 0.86 & 1.99 & Face-Adapter & 0.138 & 0.60 & 1.86 \\
    GHOST      & 0.132 & 0.90 & 2.21 & DreamID-V    & 0.149 & 0.52 & 1.67 \\
    \midrule
    \textbf{Ours} & \textbf{0.083} & \textbf{0.34} & \textbf{1.35} & \emph{Driving image} & 0.128 & 1.00 & 3.49 \\
    \bottomrule
  \end{tabular}
  }
\end{table}

\begin{figure}[t]
  \centering
  \def\icw{0.245\linewidth}
  \newcommand{\icell}[1]{\includegraphics[width=\icw]{figures/showcase/#1.jpg}}
  \newcommand{\ihead}[1]{\parbox[b]{\icw}{\centering\scriptsize #1}}
  \setlength{\tabcolsep}{0.5pt}\renewcommand{\arraystretch}{0}
  \begin{tabular}{@{}cccc@{}}
    \ihead{Target} & \ihead{Driving} & \ihead{Mask} & \ihead{\textbf{Ours}} \\[2pt]
    \icell{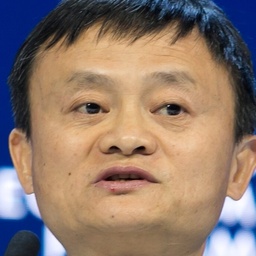} & \icell{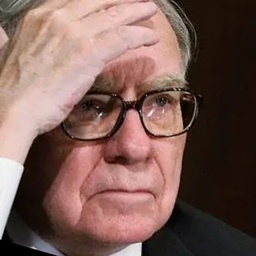} & \icell{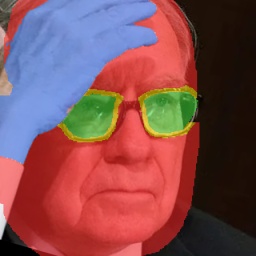} & \icell{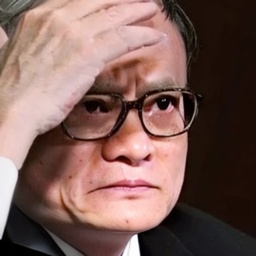} \\
  \end{tabular}
  \caption{\textbf{Occluders and glasses.} MYSTIQUE generates the face around the
  hand and behind the glasses, and still follows the lighting. In the
  mask, red regions are generated, the hand (blue) and the frame (yellow) stay from
  the driving image, and behind the clear lenses (green) the eyes are generated
  again.}
  \label{fig:showcase}
\end{figure}

\paragraph{Face randomization.}
Self-augmentation changes the face shape in the expression input, but this input,
like the lighting input, still shows the person of the ground truth, whereas at
inference both show someone other than the target. Face randomization brings this
situation into training: the expression and lighting inputs show another person,
while the ground truth stays the driving image, so copying identity from them moves
the result toward the wrong person, which the loss penalizes. For
the expression input in Stage~2, an image generator is asked to turn the driving image into a
person of a different sex, age and facial structure, and to keep the expression,
pose, lighting, skin tone and the rest of the image. For the lighting input in Stage~3, an existing
identity transfer model~\cite{inswapper} replaces the identity with a randomly
chosen one and keeps the lighting. Its results retain part of the face shape of
the ground truth, which two-pass inference keeps out of our results
(Sec.~\ref{sec:method:inference}).

\subsection{Inference}
\label{sec:method:inference}
\label{sec:method:light}
\paragraph{Two-pass inference.}
Two-pass inference keeps the lighting stream, trained on inputs with face-shape
leakage, from passing the face shape of the person in the driving image to the
result (Fig.~\ref{fig:inference}). The first pass
runs without the lighting stream, which the model supports because each stream is
dropped at random in training (Supp.~\ref{sec:suppl:impl:train}), so the face shape of this result
does not depend on the lighting input. We then warp the face of the driving image
onto this result with moving least squares~\cite{schaefer2006mls}, and the second
pass repeats the first with the warped face as the lighting input, which carries
the lighting of the driving image on the face shape of the first pass.

\paragraph{Mask and background.}
At inference, the mask $\mB$ decides which parts of the image are generated and
which stay from the driving image. By default, $\mB$ is a face-parsing
mask~\cite{yu2018bisenet} over the face, ears and neck, the regions that the
identity encoder reads (Sec.~\ref{sec:method:arch}), so the hair stays from the
driving image. Occluders in front of the face, such as hands or jewelry, are left
out of the mask and stay as well, and so does the frame of glasses; behind clear
lenses, the eyes are generated again (Fig.~\ref{fig:showcase}). To edit the hair, a
face+hair mask also covers the hair, and a prompt that describes the edit guides the
new hair, as in Fig.~\ref{fig:teaser}. Outside the mask, the latent is replaced at
every sampling step by the latent of the driving image, noised to the current
level, as in blended latent diffusion~\cite{avrahami2023blendedlatent}, so
everything outside the mask matches the driving image.

\section{Experiments}
\label{sec:exp}

\subsection{Setup}
\label{sec:exp:setup}

\paragraph{Implementation details.}
Unless noted otherwise, results use the $256$\,px model, the face mask and
$M{=}24$ photos of the target. We train by self-reconstruction on face photos of about
$30$k people, grouped by person with a face-recognition model~\cite{kim2022adaface}.
We sample with $28$ Euler steps and a guidance scale of $4$ for each of the
identity, expression and lighting streams
(Supp.~\ref{sec:suppl:impl:inference} and~\ref{sec:suppl:apps:scale}).
\paragraph{Baselines.}
We compare with twelve methods. The main paper reports eight of them: the
GAN-based SimSwap~\cite{chen2020simswap}, FaceDancer~\cite{rosberg2023facedancer},
BlendFace~\cite{shiohara2023blendface}, GHOST~\cite{groshev2022ghost},
InSwapper~\cite{inswapper} and HifiFace~\cite{wang2021hififace}, and the
diffusion-based Face-Adapter~\cite{han2024faceadapter} and
DreamID-V~\cite{guo2025dreamidv}. Supp.~\ref{sec:suppl:allmetrics} reports the other four,
DiffSwap~\cite{zhao2023diffswap}, REFace~\cite{baliah2025reface},
DiffFace~\cite{kim2022diffface} and E4S~\cite{liu2023e4s}, together with two
commercial image models and how each method is run. Each baseline runs at its
best setting: the methods that take one identity embedding receive its average
over the target's photos, which raises their recognition scores
(Sec.~\ref{sec:exp:msweep}).

\begin{figure}[t]
  \centering
  \grpgraphics[width=\linewidth]{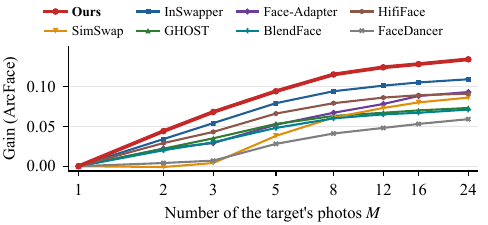}
  \caption{\textbf{More photos of the target.} Gain in identity (ArcFace,
  mean cosine to the $24$ held-out photos) over $M{=}1$, on the $240$ pairs of
  the identity benchmark. The baselines average their identity inputs over the $M$
  photos; MYSTIQUE reads the photos as a set. Absolute scores are in
  Supp.~\ref{sec:suppl:msweep}.}
  \label{fig:msweep}
\end{figure}
\subsection{Evaluation}
\label{sec:exp:eval}

We built two benchmarks for this paper: an identity benchmark of real photos, on
which human raters judge identity, and a shape benchmark of rendered faces, whose
true face shape is known.

\paragraph{Identity benchmark.}
The identity benchmark has $240$ targets from VGGFace2-HQ~\cite{cao2018vggface2,chen2024simswappp},
a public dataset that we do not train on. Each method receives $24$ photos of the target, and
$24$ other photos are held out for scoring identity. Each target is paired with a
driving image of another person from the set. The pairs form four groups of $60$.
In one group, the two people share sex and ethnicity and are close in age. In each
of the other three, they differ in one of these attributes (Supp.~\ref{sec:suppl:idbench}).

\paragraph{Identity measurement.}
The usual identity score, the cosine similarity between face-recognition
embeddings of the result and the target, has known limits. It reflects whether
two photos show the same person, not how closely two faces resemble each
other~\cite{kumagai2025perface}. It can miss fine detail, because recognizers
are trained to be robust to image quality~\cite{kim2022adaface}, and it responds
little to face shape~\cite{wang2021hififace,huang2024cscs}. It also favors methods
trained with an identity loss from the same recognizer~\cite{shiohara2023blendface}.
Recent work therefore learns face similarity from human
judgments~\cite{kumagai2025perface,huang2026alignface}. Human judgments have
limits too: in our study, the agreement between raters beyond chance is
$\kappa=0.34$, and the likeness and realism answers match in $87\%$ of the
trials (Supp.~\ref{sec:suppl:human}). Given the limits of both, we report both.
In each trial of the study, a rater sees three of the
target's photos, the driving image and two unlabeled results in
counterbalanced order, ours and one baseline's, and picks the one more like
the target and the more realistic one. The study ran in two batches of $30$
raters ($42$ people), each judging about $50$ trials per batch. The $1{,}726$ trials against the eight baselines of Table~\ref{tab:main}
use targets of the identity benchmark and celebrity targets; Supp.~\ref{sec:suppl:human}
reports the other methods. The identity score is ArcFace~\cite{deng2019arcface}
against the $24$ held-out photos, which none of our losses uses; a real photo of
the target scores $0.677$. The two can disagree: InSwapper keeps most
of the jawline of the driving image (Table~\ref{tab:shape}) yet scores $0.715$,
while raters prefer our results over InSwapper's in $98\%$ of the trials.

\begin{figure*}[t]
  \centering
  \def\acw{0.136\linewidth}
  \newcommand{\acell}[1]{\includegraphics[width=\acw]{figures/ablation/1511/#1.jpg}}
  \newcommand{\ahead}[1]{\parbox[b]{\acw}{\centering\scriptsize #1}}
  \setlength{\tabcolsep}{1pt}\renewcommand{\arraystretch}{0}
  \begin{tabular}{@{}cc@{\hspace{4pt}}ccc@{\hspace{4pt}}cc@{}}
    & & \multicolumn{3}{c}{\scriptsize Stage~2, lighting off} & \multicolumn{2}{c}{\scriptsize Stage~3, lighting on} \\[3pt]
    \cmidrule(lr){3-5}\cmidrule(l){6-7}
    \ahead{Ground truth} & \ahead{Driving} & \ahead{Face mask} & \ahead{Random block mask} & \ahead{+ both (full model)} & \ahead{One pass, no warp} & \ahead{Two passes with warp} \\[2pt]
    \acell{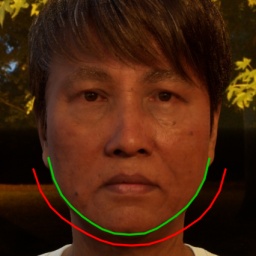} & \acell{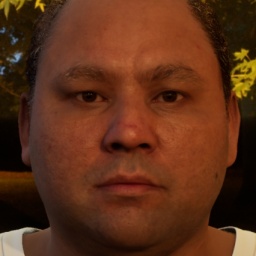} & \acell{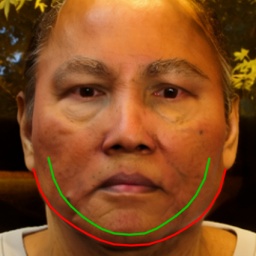} & \acell{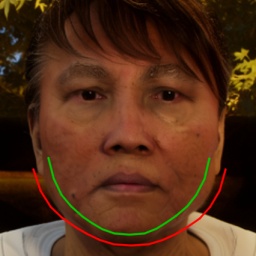} & \acell{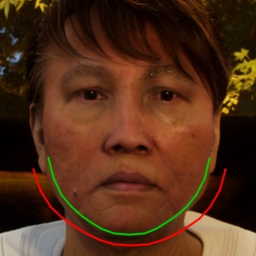} & \acell{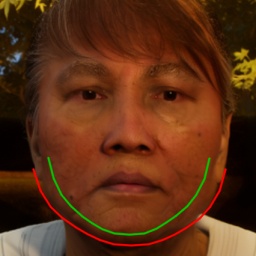} & \acell{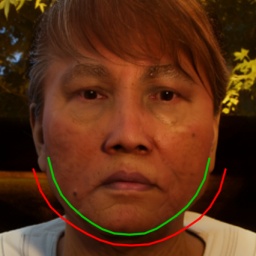} \\
  \end{tabular}
  \caption{\textbf{Face shape in the ablations.} Results for one pair of
  characters of the shape benchmark, made with the variants of Table~\ref{tab:ablation}.
  The first column is the ground truth: the target rendered with the camera,
  expression and lighting of the driving image. Each variant receives
  $24$ other renders of the target. Lines: jawline of the ground truth (green)
  and of the driving image (red).}
  \label{fig:ablation}
\end{figure*}
\paragraph{Shape benchmark.}
A real photo gives no ground truth for how the target's face would look in the
driving image, so we render faces of known shape. The shape benchmark has $29$
MetaHuman~\cite{metahuman} characters rendered in Unreal Engine~5. Each character
is the target of $12$ pairs whose driving images show other characters, which gives
$348$ pairs. The ground truth is the target rendered with the camera, expression
and lighting of the driving image, and each method receives $24$ other renders of
the target. We report three face-shape metrics (Supp.~\ref{sec:suppl:shape:bench}). The contour distance
$s_\mathrm{gt}$ is the mean distance between the jawlines of the result and the
ground truth. The leak measures how much the jawline of the result follows that of
the driving image, from $0$ (not at all) to $1$ (copied). The 3D error compares the
face shapes that Deep3DFaceRecon~\cite{deng2019accurate} reconstructs from the
result and the ground truth.

\paragraph{Expression, pose, lighting and realism.}
On the identity benchmark, we also measure how closely each result follows the
driving image. Expression is the distance between the valence and arousal that
EmoNet~\cite{toisoul2021emonet} predicts for the two images, and pose the geodesic
angle between the head rotations that HopeNet~\cite{ruiz2018hopenet} predicts.
Lighting is the correlation between the low-pass luminance of the two images inside
the face region. Realism is FaceScore~\cite{liao2024facescore}, a learned reward
for face quality that needs no reference. Three other expression measures, and
6DRepNet~\cite{hempel2022sixdrepnet} for pose, rank the methods similarly
(Supp.~\ref{sec:suppl:allmetrics}).

\subsection{Comparison with prior work}
\label{sec:exp:comparison}

Table~\ref{tab:main} compares the methods on the identity benchmark and
Table~\ref{tab:shape} on the shape benchmark. Fig.~\ref{fig:baselines} shows
examples.

\paragraph{Identity.}
Raters pick our result as more like the target in $90\%$ of the trials, from $73\%$
against Face-Adapter to $100\%$ against HifiFace. Under ArcFace, our results score
$0.601$, the highest after InSwapper, whose score lies above that of a real
photo of the target (Sec.~\ref{sec:exp:eval}).

\paragraph{Face shape.}
Six baselines follow the jawline of the driving image closely, with a leak of
$0.78$ to $0.95$ (Fig.~\ref{fig:baselines}). Face-Adapter and DreamID-V follow
it less (leak $0.60$ and $0.52$), but their contour distances ($0.138$ and
$0.149$) exceed that of the driving image ($0.128$), as does that of HifiFace
($0.138$). These results change the jawline of the driving image, but not toward
the target. Our results have the lowest contour distance, $0.083$ against $0.116$
for the closest baseline, InSwapper, and the lowest leak, $0.34$. They also have
the lowest 3D error, $1.35$ against $1.48$ for InSwapper.

\paragraph{Expression, pose, lighting and realism.}
Our expression and pose errors lie within the range of the baselines: the EmoNet
distance is $0.253$, against $0.205$ to $0.372$, and the pose error $3.99^\circ$,
against $2.03$ to $4.60^\circ$ and $7.52^\circ$ for DreamID-V. The lighting stream
raises our lighting score from $0.66$ to $0.83$ (Sec.~\ref{sec:exp:ablations}).
Scored against the driving image, these measures favor results that keep its face
(Supp.~\ref{sec:suppl:exprfloor}). FaceScore rates the realism of our results at
$4.62$, the highest in Table~\ref{tab:main}.

\subsection{Effect of the number of target photos}
\label{sec:exp:msweep}
More photos raise the recognition score of every baseline that takes an identity
embedding, and ours the most (Fig.~\ref{fig:msweep}). Each method receives the
first $M$ photos of a fixed order that varies the pose; the baselines average their
identity inputs, and DreamID-V, which takes one photo, is left out. From $M{=}1$ to
$24$, the baselines gain $0.059$ to $0.109$ in ArcFace, and ours, which combines
the photos as a set, gains $0.134$.
The added photos bring our face shape closer to the target as well: the contour
distance falls from $0.090$ to $0.083$. This needs little leakage. The contour
distance of InSwapper changes by less than $0.001$, and that of our model without
the strategies (first row of Table~\ref{tab:ablation}) stays near the driving
image ($0.124$ against $0.128$), although its ArcFace score rises from $0.470$ to
$0.599$.

\subsection{Ablations}
\label{sec:exp:ablations}
Table~\ref{tab:ablation} ablates the disentanglement strategies
(Sec.~\ref{sec:method:strategies}), the expression encoder and two-pass
inference (Sec.~\ref{sec:method:inference}). The strategies and two-pass
inference act mainly on the face shape, and the expression encoder on the
expression and pose.

\begin{table}[t]
  \centering\footnotesize
  \setlength{\tabcolsep}{2.0pt}
  \caption{\textbf{Ablations.} Each row is one model. Left: identity,
  expression, pose and lighting on the identity benchmark, as in
  Table~\ref{tab:main}. Right: face shape on the shape benchmark, as in
  Table~\ref{tab:shape}. The two groups use different data.}
  \label{tab:ablation}
  \begin{tabular}{@{}l cccc @{\hspace{4pt}} ccc@{}}
    \toprule
    & \multicolumn{4}{c}{Identity benchmark} & \multicolumn{3}{c}{Shape benchmark} \\
    \cmidrule(r{4pt}){2-5} \cmidrule(l{2pt}){6-8}
    Variant & Id.$\uparrow$ & Expr.$\downarrow$ & Pose$\downarrow$ & Light$\uparrow$ & $s_\mathrm{gt}\downarrow$ & Leak$\downarrow$ & 3D$\downarrow$ \\
    \midrule
    \multicolumn{8}{@{}l}{\emph{Stage 2, lighting off}} \\
    Face mask & 0.599 & 0.268 & 3.44 & 0.74 & 0.124 & 0.82 & 1.51 \\
    Random block mask & 0.610 & 0.248 & 3.68 & 0.67 & 0.098 & 0.54 & 1.37 \\
    \quad + self-augmentation & 0.610 & 0.256 & 3.98 & 0.66 & 0.082 & 0.38 & 1.34 \\
    \quad + face randomization & 0.618 & 0.256 & 3.78 & 0.67 & 0.087 & 0.45 & 1.28 \\
    \quad + both (full model) & 0.616 & 0.256 & 3.97 & 0.66 & 0.081 & 0.33 & 1.26 \\
    \addlinespace[2pt]
    Geometry branch only & 0.634 & 0.375 & 5.11 & 0.57 & 0.093 & 0.23 & 1.47 \\
    Latent branch only & 0.613 & 0.271 & 4.52 & 0.63 & 0.082 & 0.33 & 1.30 \\
    \midrule
    \multicolumn{8}{@{}l}{\emph{Stage 3, lighting on}} \\
    One pass, no warp & 0.599 & 0.248 & 3.42 & 0.87 & 0.095 & 0.60 & 1.45 \\
    Two passes with warp & 0.601 & 0.253 & 3.99 & 0.83 & 0.083 & 0.34 & 1.35 \\
    \bottomrule
  \end{tabular}
\end{table}

\paragraph{Face shape.}
Each strategy, and then two-pass inference, lowers the leak
(Fig.~\ref{fig:ablation}). Trained with the face-parsing mask of inference, the
model follows the outline of this mask (leak $0.82$). The random block mask lowers the
leak to $0.54$, and self-augmentation and face randomization together to $0.33$.
The lighting stream raises the lighting score from $0.66$ to $0.83$, but in a
single pass it brings the face shape of the driving image back (leak $0.60$); two
passes keep the leak at $0.34$. Identity and expression change little along these
rows; the pose error grows by about $0.5^\circ$, in part because the pose estimator
also responds to face shape (Supp.~\ref{sec:suppl:exprfloor}).

\paragraph{Expression and pose.}
The two branches of the expression encoder work best together. The geometry
branch alone raises the expression and pose errors to $0.375$ and $5.11^\circ$;
the latent branch alone raises the expression error to $0.271$ and lowers the
lighting score to $0.63$. Neither brings the face shape closer to the target.

\section{Conclusion}
\label{sec:conclusion}
We presented MYSTIQUE, an identity transfer method that learns to disentangle identity from
the inputs that read the driving image. It decouples identity, expression and
lighting into separate streams of a frozen diffusion transformer, and reduces
leakage through these inputs with multi-stage training, three disentanglement
strategies and two-pass inference. The face shape then comes mainly from the
target's photos, and added photos bring it closer to the target. In a user study,
raters judge our results more like the target than those of each compared
method, and on rendered faces, ours come closest to the target's face shape.

\section*{LLM Usage Statement}
We used Claude to write and run code for training and evaluation, and to draft
and polish the text under the authors' direction. No figure was generated by an
AI model, apart from the results and face-randomized images that the paper
describes. Within the method and evaluation, Qwen2.5-VL names occluders, a
closed-source image model creates training inputs, a multimodal GPT model
labels ethnicity to form the benchmark pairs, and GPT and Gemini image models
create images for a comparison in the supplement.
{
    \small\fontsize{9}{10.8}\selectfont
    \bibliographystyle{ieeenat_fullname}
    \bibliography{main}
}

\clearpage
\maketitlesupplementary
\setcounter{section}{0}
\renewcommand{\thesection}{\Alph{section}}
\renewcommand{\theHsection}{S\Alph{section}}%
\renewcommand{\floatpagefraction}{0.6}
\renewcommand{\dblfloatpagefraction}{0.6}

\section{Method and implementation details}
\label{sec:suppl:impl}
This section gives the details of Sec.~\ref{sec:method} and
Sec.~\ref{sec:exp:setup} that are needed to reproduce MYSTIQUE, in the order of
Sec.~\ref{sec:method}. They refer to the $256$\,px model unless noted otherwise;
Sec.~\ref{sec:suppl:impl:inference} ends with the changes for the $512$\,px
model.

\subsection{Architecture}
\label{sec:suppl:impl:arch}
Table~\ref{tab:suppl:encoders} lists the input, the encoder and the tokens of
each stream, and the stage that trains it. The paragraphs below follow its rows.

\begin{table*}[t]
  \centering\footnotesize
  \setlength{\tabcolsep}{4pt}
  \caption{\textbf{The streams of MYSTIQUE} ($256$\,px model). Tokens: number
  $\times$ width. The scene tokens enter the sequence of the transformer; the
  other streams enter every block through cross-attention.}
  \label{tab:suppl:encoders}
  \begin{tabular}{@{}l p{4.1cm} p{4.1cm} c p{2.7cm}@{}}
    \toprule
    Stream & Input & Encoder & Tokens & Trained in \\
    \midrule
    Identity & $M$ photos of the target, $448$\,px; face, ears and neck & CLIP ViT-H/14 and attention pooling & $256\times1280$ & Stage~1 \\
    Expression (geometry) & driving image, $256$\,px & keypoint extractor, pretrained separately & $128\times1280$ & frozen; embedding in Stage~2 \\
    Expression (latent) & driving image, $224$\,px & X-NeMo expression encoder~\cite{zhao2025xnemo} & $128\times1280$ & Stage~2 \\
    Lighting & face of the driving image, $256$\,px & convolutional network on a $16\times16$ grid & $256\times1280$ & Stage~3 \\
    Scene & driving image without the face, and $\mB$ & autoencoder and patch embedding & $256\times4096$ & Stages~2 and~3 \\
    \bottomrule
  \end{tabular}
\end{table*}

\paragraph{Input crops.}
All inputs are crops around the face. A box is fitted to five facial landmarks
(the eyes, the tip of the nose and the corners of the mouth), enlarged by a
factor of $3.0$ and shifted up by $0.2$ of its height; parts outside the image
are filled with black. Photos of the target are resized to $448$\,px and the
driving image to $256$\,px.

\paragraph{Identity encoder.}
The identity encoder follows the architecture of Omni-ID~\cite{qian2024omniid},
and we train it ourselves (Sec.~\ref{sec:suppl:impl:train}). A face-parsing
model~\cite{yu2018bisenet} first sets every pixel of a photo outside the face,
ears and neck to black, so the encoder does not see the hair. A CLIP
ViT-H/14~\cite{radford2021clip,cherti2023openclip}, initialized from its
LAION-2B weights, reads each $448$\,px photo, and we take the $1024$ patch
tokens of its penultimate layer without the class token. The patch tokens of
all $M$ photos form one key and value sequence, and a one-dimensional rotary
embedding~\cite{su2024roformer} of the photo index on the keys tells the photos
apart. $256$ learnable queries then pass through one self-attention block and
two pairs of cross-attention and self-attention blocks, with $16$ heads each,
and a final layer normalization gives $\mC_\mathrm{id}\in\R^{256\times1280}$.
Because the queries are fixed, the number of tokens does not depend on $M$.

\paragraph{Geometry branch.}
The geometry branch extracts $128$ implicit 3D keypoints of the face from the
driving image, in the manner of LivePortrait~\cite{guo2024liveportrait}. We
pretrained it as a separate model, which differs from LivePortrait in two ways:
it predicts $128$ keypoints instead of $21$, and it represents the head rotation
in 6D. In this pretraining, a ViT generator renders the face from the keypoints
and appearance features, and 6DRepNet~\cite{hempel2022sixdrepnet} supervises the
head rotation with a geodesic loss. Two consistency losses are added: an identity
consistency loss penalizes differences between the identity vectors of two
images of the same person, and an expression consistency loss penalizes
differences between the expression vectors of different views of the same
expression in a canonical space.
The data
are CelebV-Text~\cite{yu2023celebvtext}, CelebV-HQ~\cite{zhu2022celebvhq},
DH-FaceVid-1K~\cite{di2025dhfacevid}, MEAD~\cite{wang2020mead} and about $80$k generated face
images, and the encoder was trained for about $1.03$M steps. MYSTIQUE uses the
keypoint extractor only, frozen. Its driven keypoints, normalized to
$[-1,1]$, pass through a linear layer from $3$ to $1280$ dimensions, and a
learnable position embedding per keypoint, initialized at zero, is added.

\paragraph{Latent branch.}
The latent branch reads a latent expression descriptor from the face image with the
expression encoder of X-NeMo~\cite{zhao2025xnemo}, a network of about $65$M
parameters that takes $224$\,px input. We initialize it from the X-NeMo
weights, without its bounding-box conditioning, fine-tune it in Stage~2 and freeze
it in Stage~3. It outputs one $512$-dimensional vector. A linear layer expands
this vector into $128$ tokens of width $16$, as many as the keypoint tokens of the
geometry branch, a fixed sinusoidal position encoding is added, and a second
linear layer, initialized at zero, maps each token to width $1280$. The two
branches are then fused: each set of tokens is layer-normalized, the two are
averaged and the average is layer-normalized again, which gives
$\mC_\mathrm{exp}\in\R^{128\times1280}$. Both branches are on for every
sample.

\paragraph{Lighting encoder.}
The lighting encoder reads the driving image with every pixel outside the face
set to zero; the face here is the skin, brows, eyes, nose, mouth and glasses,
without the ears, neck and hair. Four convolutional stages with stride $2$
($32$, $64$, $128$ and $256$ channels, group normalization and SiLU) reduce the
$256$\,px input to a $16\times16$ grid, so each cell covers one latent token.
Each cell is layer-normalized and projected by a shared linear layer to width
$1280$, and a learnable position embedding per cell is added, which gives
$\mC_\mathrm{light}\in\R^{256\times1280}$.

\paragraph{Cross-attention.}
The identity, expression and lighting tokens enter all $32$ blocks of the
transformer, its $8$ double-stream and $24$ single-stream blocks. The shared
projection $\phi_k$ of each stream is a two-layer MLP with a GELU, from $1280$
to $4096$ dimensions, followed by layer normalization. In each block, linear
layers of size $4096\times4096$ give the keys and values of the stream, and the
keys receive an RMS normalization per head. The queries and the output
projection are those of the transformer and stay frozen. The value layers start
at zero, so a new stream does not change the output of the model when its
training begins. The outputs of the streams are added with weight $1$ to the
hidden states of the noisy latent tokens and to no other tokens.

\paragraph{Scene tokens and text.}
The face region of the scene is filled with mid-gray, which is $0$ in the input
range $[-1,1]$ of the autoencoder, so $\tS=(\vone-\mB)\odot\tD$ holds there. The
mask is average-pooled to the $16\times16$ latent grid and appended to the $128$
latent channels as a $129$th channel. A linear layer from $129$ to $4096$
dimensions, initialized from the patch embedding of the transformer with a zero
column for the mask, gives the scene tokens. They carry a rotary time offset of
$1$, which separates them from the noisy latent tokens at the same spatial
positions, and their modulation uses the timestep of a clean image. A fixed
prompt, ``a photorealistic portrait of a human face'', completes the sequence in
training and at inference, and it is never dropped.

\paragraph{Low-rank update.}
The low-rank update has rank $64$ and scaling factor $64$. In the double-stream
blocks, it adapts the query, key, value and output projections of the image
tokens; in the single-stream blocks, it adapts the fused input projection and
the output projection. The projections of the text tokens are not adapted.

\subsection{Training}
\label{sec:suppl:impl:train}
Table~\ref{tab:suppl:train} gives the schedule of each stage. Each stage starts
from the weights of the previous one, and all stages draw $M$ uniformly from
$\{1,\ldots,24\}$, once per step on each GPU.

\begin{table*}[t]
  \centering\footnotesize
  \setlength{\tabcolsep}{4pt}
  \caption{\textbf{Training schedule} ($256$\,px model). Batch: samples per step
  over all GPUs. LR: learning rate. All stages use mixed precision
  (bfloat16).}
  \label{tab:suppl:train}
  \begin{tabular}{@{}l l r l l l p{5.6cm}@{}}
    \toprule
    Stage & Step & Steps & LR & Batch & GPUs & Trained \\
    \midrule
    1, identity & reconstruction & $925$k & $10^{-4}$ & $16$ people & $8$ H200 & identity encoder, masked decoder \\
    & flow matching & $410$k & $10^{-5}$ & $8$ & $8$ H200 & identity encoder, identity cross-attention \\
    \addlinespace[2pt]
    2, identity transfer & fixed face mask & $100$k & $10^{-5}$ & $16$ & $8$ B200 & latent branch; projections and cross-attention of the identity and expression streams; scene patch embedding; low-rank update \\
    & \parbox[t]{2.6cm}{\raggedright random block mask and both strategies} & $100$k & $10^{-5}$ & $16$ & $8$ H200 & as in the first step of Stage~2 \\
    \addlinespace[2pt]
    3, lighting & -- & $50$k & $2\times10^{-5}$ & $32$ & $16$ H200 & lighting encoder and cross-attention; scene patch embedding \\
    \bottomrule
  \end{tabular}
\end{table*}

\paragraph{Stage 1.}
Stage~1 trains the identity encoder in two steps. In the first step, a masked
decoder reconstructs other photos of the person from the identity tokens, as in
Omni-ID. The loss combines a region-weighted L1 term (weight $1.0$), an
LPIPS~\cite{zhang2018lpips} term with VGG features (weight $0.5$) and an identity term (weight $0.1$),
$1-\cos$ between the AdaFace~\cite{kim2022adaface} embeddings of the
reconstruction and of the photo, both aligned to $112$\,px with five landmarks.
The identity term is off for the first $45$k of the $925$k steps.
Each sample is one person, with $4$ photos to reconstruct and $M$ photos for
the identity encoder. The learning rate warms up over $500$ steps to $10^{-4}$
and decays with a cosine schedule. The run was resumed once, at $775$k steps,
with the schedule extended to $1$M steps, which raised the learning rate to
$1.2\times10^{-5}$ before it decayed again; we use its checkpoint at $925$k steps.
In the second step, the encoder and the identity cross-attention train through
the frozen transformer with the flow-matching loss of Eq.~\ref{eq:fm}: the
transformer generates other photos of the person from the identity tokens,
without a driving image. The CLIP backbone is fine-tuned in both steps. Stage~1
uses our dataset for $99\%$ of its samples and MEAD for $1\%$.

\paragraph{Stage 2.}
Stage~2 trains identity transfer by self-reconstruction on our dataset, in which
the photos are grouped by person (Sec.~\ref{sec:exp:setup}). The driving image
is a photo drawn uniformly from one group, and the photos of the target are up
to $24$ other photos of the same group, drawn in turn from clusters of head
pose; the model receives the first $M$ of them. The driving image itself is
never among the target's photos. The stage runs in two phases of $100$k steps each. The first phase
trains with a face mask, the face-parsing region of the face, ears and neck
dilated by $12$\,px, and without the disentanglement strategies; it is the
model of the first row of Table~\ref{tab:ablation}. The second phase continues
from it with the random block mask, self-augmentation and face randomization
(Sec.~\ref{sec:suppl:impl:strategies}). In both phases, the input of the
latent branch also receives color jitter (brightness, contrast and saturation
$\pm0.2$, hue $\pm0.05$), shifts of up to $\pm3\%$ and scaling of up to
$\pm5\%$, without rotation. These are on in every Stage~2 run, including the
first row of Table~\ref{tab:ablation}, and so are not part of
self-augmentation. Both phases
use a weight decay of $0.01$, $500$ warmup steps and gradient clipping at
$1.0$.
\paragraph{Stage 3.}
Stage~3 adds the lighting stream and trains it for $50$k steps, with $500$
warmup steps and a cosine schedule. The identity and expression streams and the
low-rank update stay frozen. The lighting input is the driving image after face
randomization (Sec.~\ref{sec:suppl:impl:strategies}), with every pixel outside
the face set to zero, and the visible blocks of the scene come from the
background only. Neither self-augmentation nor face randomization changes the
expression input in this stage. Stage~3 uses the dataset and the frame sampling
of Stage~2; the driving image is the last of the frames drawn for a sample.

\paragraph{Stream dropout and guidance.}
Each of the identity, expression and lighting streams is dropped with
probability $0.1$; a dropped stream has its keys and values set to zero. The
three draws are independent, and each is made once per step for the whole
batch. The scene and the text prompt are never dropped. All conditioned streams
are off together in about $1\%$ of the Stage~2 steps (two streams) and $0.1\%$
of the Stage~3 steps (three streams). This trains the prediction without the
lighting stream that the first pass uses, and the unconditional prediction of
classifier-free guidance~\cite{ho2022cfg}, which drops the three streams together
(Sec.~\ref{sec:suppl:impl:inference}). The noise level of
Eq.~\ref{eq:fm} is $\sigma=\mathrm{sigmoid}(u)$ with $u\sim\mathcal{N}(0,1)$.

\subsection{Disentanglement strategies}
\label{sec:suppl:impl:strategies}
\label{sec:suppl:facerand}

\paragraph{Random block mask.}
The random block mask of Stage~2 is drawn anew for every sample. A ratio $r$ is
drawn uniformly from $[0.95,0.99]$ once per step on each GPU, and each of the
$256$ blocks of $16\times16$ pixels is hidden independently with probability
$r$. On average, about $3\%$ of the blocks, about $8$, stay visible, anywhere in
the image and also on the face. Stage~3 hides every block that touches the
person, including the hair, and keeps blocks only from the background, with the
same probabilities, so that the scene shows no lighting on the person and the
model reads it from the lighting stream (Sec.~\ref{sec:method:strategies}). On our tight crops, about $0.8\%$ of the blocks stay
visible, about $2$ per image, and about a quarter of the samples show none.
Training masks do not leave out occluders; this happens only at inference.

\paragraph{Self-augmentation.}
Self-augmentation warps the expression input of every sample of the second
phase of Stage~2, the input of both branches. Moving least
squares~\cite{schaefer2006mls} first moves the outline of the face by a random
factor between $0.88$ and $1.12$, and the aspect ratio of the face is then
scaled by a random factor between $0.8$ and $1.25$, without translation. The
scene, the lighting input and the ground truth are not warped.

\begin{figure}[t]
  \centering
  \includegraphics[width=\linewidth]{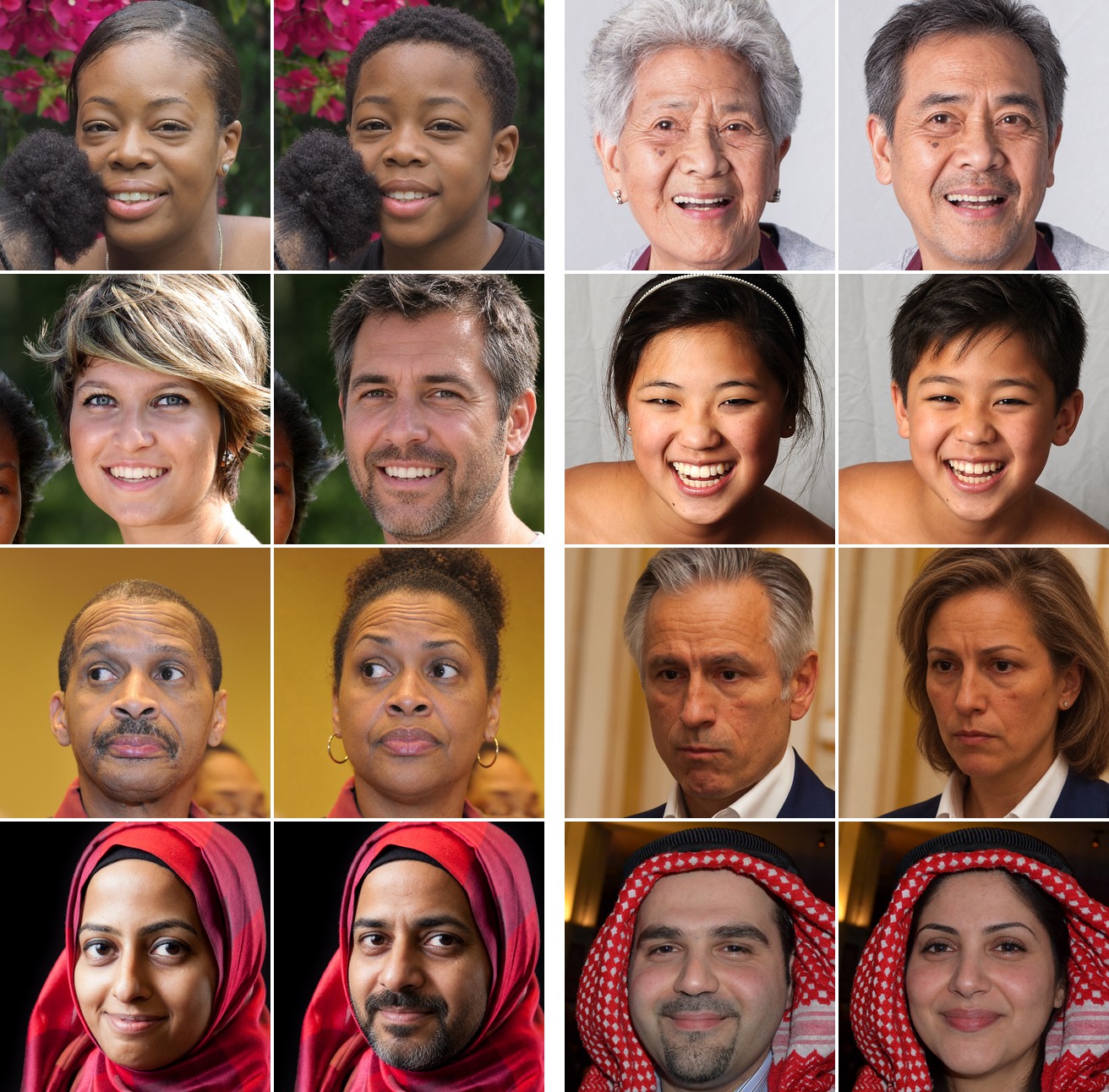}
  \caption{\textbf{Face randomization on public photos.} Each pair shows a
  photo from FFHQ~\cite{karras2019stylegan} (left) and the image that the image
  generator makes from it with the prompt of training (right). The landmark
  check of training keeps all eight. The generated person differs in sex and
  age, while the pose, expression, lighting and background stay close to the
  photo.}
  \label{fig:suppl:facerand}
\end{figure}

\paragraph{Face randomization.}
Face randomization replaces the person in the expression input in $30\%$ of the
samples of the second phase of Stage~2. For these samples, a closed-source image
generator edits the photo into a new person, with the prompt below. It receives
the whole square photo of our dataset, in which the face spans about $18\%$ of
the side, rather than the training crop (about $55\%$), so it has room to change
the face shape. The expression input is then cut from its output with the same
crop as real photos, from landmarks detected on the generated image.
Table~\ref{tab:suppl:facerandcheck} checks that the generated images keep the
expression and pose. The photos of the target and the ground truth
stay real photos of the person in the driving image, so copying identity from
the expression input moves the result away from the ground truth.
The prompt, verbatim:
\begin{quote}\footnotesize
Replace the person in this photograph with a DIFFERENT person of a different sex
and a different apparent age, so that no face-recognition system could match
them to the original. Change the underlying bone structure: the skull and
cheekbone shape, face width and length, jaw and chin shape, nose, lips, eye shape
and spacing, and brow shape. CRITICAL: the new person has EXACTLY the same skin
tone, complexion, undertone and skin colour as the original -- do not lighten,
darken or change the hue of the skin anywhere, and keep every shadow and
highlight on the skin exactly as it is. Keep absolutely everything else
pixel-identical: the head pose and camera angle, the facial expression and every
muscle position (brows, eyelid opening, mouth shape, jaw opening), the gaze
direction, the lighting direction / colour / intensity and all shadows and
highlights on the face, the skin tone and colour grade, the clothing, the
background, the framing and the scale and position of the head in the frame,
and the image sharpness and film grain. Do not re-pose, re-light, re-frame,
crop, zoom, straighten or beautify anything. Output the same resolution.
\end{quote}
We requested $40{,}000$ images, each from a randomly chosen photo of our
dataset; $1{,}487$ were blocked by the content filter of the generator, and $38{,}513$
were generated. A landmark check removes the images in which the generator
moved or replaced the person (the face more than $0.5$ inter-ocular distances
from its original position), and keeps $37{,}883$ of them, of about $30$k
people. The similarity of identity between the photo and the generated image is
recorded but not used to filter. Fig.~\ref{fig:suppl:facerand} shows the
generator on public photos.
For the lighting input of Stage~3, InSwapper~\cite{inswapper} replaces the
identity of every sample with one drawn at random from a fixed pool of $511$
identities of our dataset, and no warp is applied.

Table~\ref{tab:suppl:facerandcheck} checks the generated images that training
used, on $500$ random pairs among the $37{,}883$ that the landmark check keeps.
They change the identity and the face shape: the ArcFace similarity to the photo
is $0.39$ on average, lower than for $99.6\%$ of the pairs of photos of the same
person (mean $0.73$), and the jawline moves by $0.14$ in the units of
$s_\mathrm{gt}$, about as much as a typical warp of self-augmentation. In $15$ of
$16$ random pairs that we inspected, the sex changes. The new person still
shares some features of the photo, as the prompt keeps the skin tone and every
shadow and highlight of the skin, and the generator does not change every
feature: the similarity stays above that of a photo of another person ($0.01$).
They keep the expression and pose: the EmoNet distance, $0.20$, stays
below what this estimator gives for two characters of the shape benchmark with
the same expression and pose ($0.26$, Sec.~\ref{sec:suppl:exprfloor}), and the
pose error, $3.6^\circ$, lies $1.0^\circ$ above theirs. The lighting correlation
is $0.77$, and the skin becomes slightly darker ($-2.4$ in $L^*$). Training on
them does not weaken expression and pose: adding face randomization to the
random block mask (Table~\ref{tab:ablation}) changes the EmoNet distance by
$+0.008$ ($95\%$ interval $[-0.007, 0.022]$) and the 6DRepNet pose error by
$+0.2^\circ$ ($[0.05, 0.43]$), less than a change of face shape alone moves this
estimator ($1.4^\circ$ to $1.6^\circ$, Table~\ref{tab:suppl:shapebias}).
InSwapper keeps the lighting better, but moves the jawline less ($0.09$), so its
results keep part of the face shape of the photo
(Sec.~\ref{sec:method:strategies}).

\begin{table}[t]
  \centering\footnotesize
  \setlength{\tabcolsep}{4pt}
  \caption{\textbf{What face randomization changes and keeps}, on $500$ random
  photos of training. Each column compares the photo with another image: the
  generated image of Stage~2, another photo of the same person, a photo of
  another person, and the InSwapper result of Stage~3. ArcFace: cosine
  similarity. Jawline: distance between the jawlines, in the units of
  $s_\mathrm{gt}$, each image in its own crop. EmoNet and 6DRepNet: expression
  and pose as in Sec.~\ref{sec:suppl:metrics}, with the photo as the driving
  image; for two characters of the shape benchmark with the same expression and
  pose, they give $0.26$ and $2.62^\circ$ (Table~\ref{tab:suppl:shapebias}).
  Light: lighting correlation. $|\Delta L^*|$: absolute difference of skin
  brightness.}
  \label{tab:suppl:facerandcheck}
  \begin{tabular}{@{}l cccc@{}}
    \toprule
    & Generated & Same person & Other person & InSwapper \\
    \midrule
    \multicolumn{5}{@{}l}{\emph{Identity and face shape: should change}} \\
    ArcFace & 0.39 & 0.73 & 0.01 & 0.03 \\
    Jawline & 0.14 & 0.51 & 0.72 & 0.09 \\
    \midrule
    \multicolumn{5}{@{}l}{\emph{Expression, pose and lighting: should stay}} \\
    EmoNet & 0.20 & 0.37 & 0.48 & 0.31 \\
    6DRepNet (deg) & 3.6 & 27.0 & 37.5 & 3.1 \\
    Light & 0.77 & 0.49 & 0.12 & 0.92 \\
    $|\Delta L^*|$ & 3.1 & 5.3 & 13.7 & 1.3 \\
    \bottomrule
  \end{tabular}
\end{table}

\subsection{Inference}
\label{sec:suppl:impl:inference}

\paragraph{Two-pass inference.}
The first pass sets the keys and values of the lighting stream to zero. To
build the lighting input of the second pass, we detect $106$ facial landmarks
with InsightFace and parse the faces with the face-parsing model in the driving
image and in the first-pass result. The control points of the warp are the
interior landmarks, the points where $72$ rays from the center of the face
meet the face boundary, and points on the edges of the ears and neck. A
similarity-based moving least squares warp~\cite{schaefer2006mls} maps the face
of the driving image onto the first-pass result, and the warped image is parsed
again. The lighting input keeps the warped face where it overlaps the face of
the first-pass result or lies above the brows, without the ears. If no face is
found, or fewer than $12$ rays are valid, the second pass reads the face of the
driving image without the warp. The second pass uses the same seed and all
other inputs of the first pass.

\paragraph{Masks and occluders.}
The face mask is the face-parsing region of the face, ears and neck, dilated by
$12$\,px, minus the occluders dilated by $3$\,px. Occluders are found once per
driving image by an open-vocabulary detector chain. Grounding
DINO~\cite{liu2024groundingdino} looks for a
fixed list of objects (a hand, fingers, a microphone, a mask, food, an ice
cream, a phone, a camera, flowers, a straw, a cigarette, a cup, a bottle, a
scarf and a fist). Qwen2.5-VL-7B-Instruct~\cite{bai2025qwen25vl} names further objects that cover or
touch the central face, other than glasses, hair and hats, and other people in
the image; each name is located by the language model and by Grounding DINO at
a low threshold. SAM~2.1~\cite{ravi2025sam2} segments all boxes; for other people, negative clicks
on the eyes, nose and mouth of the person in the driving image keep that face
out. Names of glasses, hair, hats, clothing and face parts are discarded, and
the union of the masks, dilated by $3\times3$ pixels and without the eyes of
the person in the driving image, is the occluder map. The face+hair mask also
covers the hair, with the same dilation, and leaves no occluders out.

This default is automatic and fast. For $21$ of the $240$ pairs of our row in
Table~\ref{tab:main}, the automatic map marked too much or too little, and we
corrected the mask or the occluder map by hand; the other $219$ pairs keep the
automatic map. In film production, an
artist can make such corrections by rotoscoping or with stronger detection and
segmentation models, and the model itself does not change. With the automatic
map for all $240$ pairs, our scores in Table~\ref{tab:main} change by at most
$0.004$ (ArcFace) and $0.02$ (FaceScore), and our rank in each column does not
drop. Our other rows in Tables~\ref{tab:main} and~\ref{tab:ablation} use the same
maps; the baselines use their own defaults.
\paragraph{Glasses.}
In film, actors and stunt doubles often wear glasses, and clear lenses still
show the eyes behind them. Glasses are therefore not handled like the occluders
of the previous paragraph, which stay from the driving image: behind clear
lenses, the eyes must become those of the target. They follow their own rule in
each pass.
The glasses region, frame and lenses included, is generated for the first
$60\%$ of the sampling steps. From then on, its latent is set to the noised latent of the
driving image at each step, and to the clean latent after the last one, so the
frame stays that of the driving image.
Lenses are dark if their mean brightness is below $0.31$ and below $0.72$ times
that of the skin; dark lenses are kept as they are. Behind clear lenses, one
more sampling run, which starts from the output of the pass and uses a new seed,
generates the lens interior again, so the eyes of the result show through. A result
therefore takes two sampling runs, or four with clear lenses. Of the $240$
driving images of the identity benchmark, $10$ show glasses, $9$ of them with
clear lenses.

\paragraph{Sampling.}
We sample with $28$ Euler steps on the shifted noise schedule of FLUX.2, whose
shift follows the length of the sequence of $256$ latent tokens. Guidance
combines the two predictions as
$\vv=\vv_\mathrm{u}+4\,(\vv_\mathrm{c}-\vv_\mathrm{u})$. The unconditional
prediction $\vv_\mathrm{u}$ sets the keys and values of the identity, expression
and lighting streams to zero and replaces the text with the negative prompt
below; both predictions share the scene tokens. At every step, the latent
outside the mask is replaced by the latent of the driving image noised to the
current level with the initial noise, and the clean latent of the driving image
is written back at the end~\cite{avrahami2023blendedlatent}. In our
evaluations, each pair uses one fixed seed in both passes.

\begin{figure}[t]
  \centering
  \setlength{\tabcolsep}{1pt}
  \renewcommand{\arraystretch}{0.6}
  \begin{tabular}{@{}ccc@{}}
    \scriptsize Driving image & \scriptsize Without & \scriptsize With \\
    \includegraphics[width=0.325\linewidth]{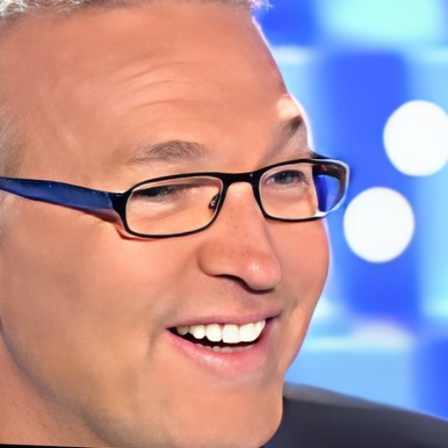} &
    \includegraphics[width=0.325\linewidth]{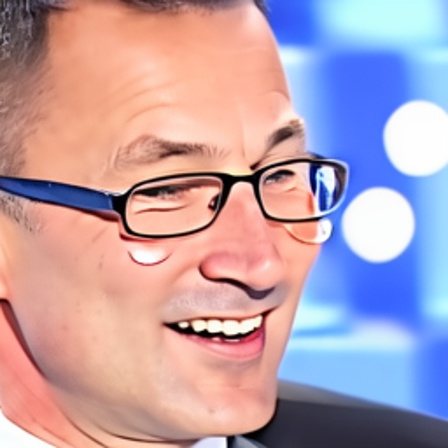} &
    \includegraphics[width=0.325\linewidth]{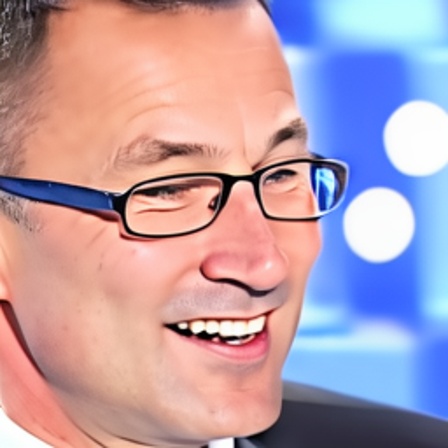} \\
    \includegraphics[width=0.325\linewidth]{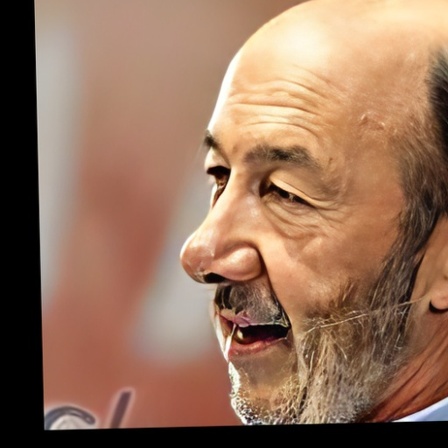} &
    \includegraphics[width=0.325\linewidth]{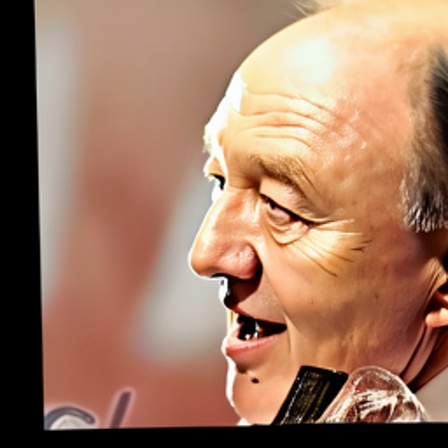} &
    \includegraphics[width=0.325\linewidth]{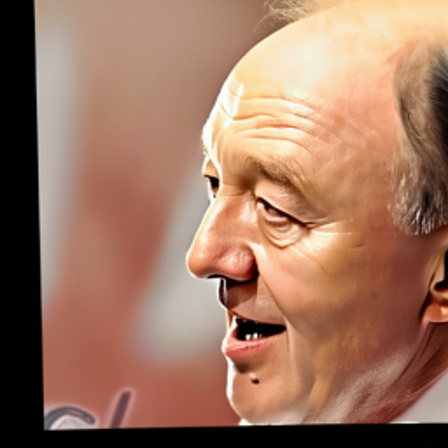} \\
  \end{tabular}
  \caption{\textbf{Negative prompt.} Results of our model without and with the
  negative prompt, with the same seed. Without it, a stain appears on the cheek
  (top) and an object at the mouth (bottom). With it, ArcFace rises from $0.53$
  to $0.58$ and from $0.60$ to $0.64$, the pose error falls from $7.0^\circ$ to
  $5.4^\circ$ and from $5.0^\circ$ to $3.6^\circ$, four to seven times the mean
  change over the identity benchmark, and the EmoNet distance falls from $0.14$
  to $0.09$ and from $0.37$ to $0.24$. Apart from the artifact, the results
  with and without the prompt look the same.}
  \label{fig:suppl:negprompt}
\end{figure}

\paragraph{Negative prompt.}
The negative prompt reads ``face paint, tattoos, camouflage pattern on the
skin, markings, stains, colored patches on the face, text, captions, subtitles,
letters, watermark, logo, Korean characters, hands, fingers, a hand touching the
face, cables, cords''. We found that it lowers the chance that the model paints
artifacts in the generated region, such as marks like tattoos, captions or
hands. The model never sees it in training. On the identity benchmark, removing
it lowers ArcFace from $0.601$ to $0.592$, raises the pose error from $3.99$ to
$4.23$ degrees and lowers the lighting score from $0.83$ to $0.82$, while the
EmoNet distance changes by $0.003$. The prompt names only artifacts, so we read
these changes as fewer artifacts that disturb the scores, not as better identity
transfer (Fig.~\ref{fig:suppl:negprompt}).

\paragraph{Edits.}
An edit replaces the fixed prompt of the conditional prediction with a prompt
that describes the edit, and uses the face+hair mask. The edits of
Fig.~\ref{fig:teaser} use ``a photorealistic portrait of a human face with
tightly curly hair'', ``a photorealistic portrait of a human face with platinum
blonde hair'' and ``a photorealistic portrait of a human face wearing a crown of
fresh flowers in her hair'', with the same negative prompt. Training uses the
fixed prompt only.

\paragraph{Run time.}
A result of the $256$\,px model takes $17$ to $21$\,s on one H200 GPU, averaged
over the identity benchmark, with both passes; results with and without glasses
were not timed apart.

\begin{figure*}[t]
  \centering
  \edef\rcell{\the\dimexpr 0.1555\textwidth\relax}\edef\rgap{\the\dimexpr 0.0024\textwidth\relax}\edef\rbig{\the\dimexpr 0.0073\textwidth\relax}%
  \makebox[0.95\textwidth][l]{\makebox[\rcell][c]{\scriptsize Target}\hspace{\rgap}\makebox[\rcell][c]{\scriptsize Driving}\hspace{\rbig}\makebox[\rcell][c]{\scriptsize $256$\,px}\hspace{\rgap}\makebox[\rcell][c]{\scriptsize $512$\,px}\hspace{\rgap}\makebox[\rcell][c]{\scriptsize $256$\,px, box}\hspace{\rgap}\makebox[\rcell][c]{\scriptsize $512$\,px, box}}\\[2pt]
  \includegraphics[width=0.95\textwidth]{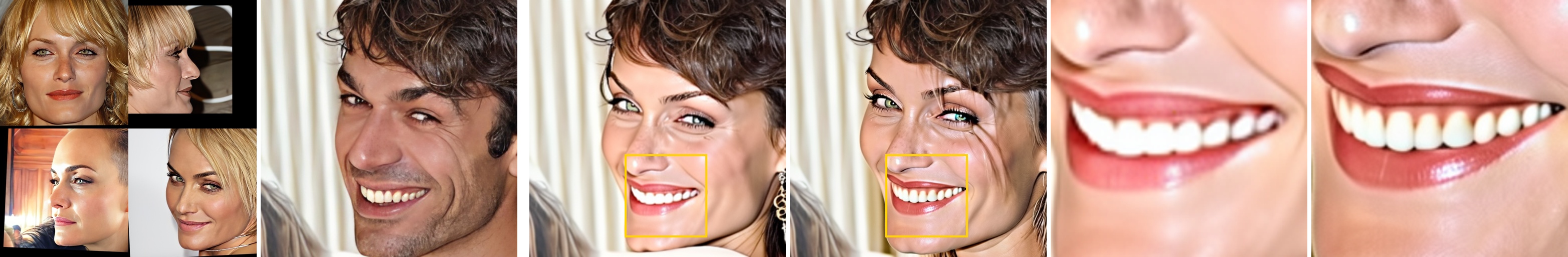}\\[\rgap]
  \includegraphics[width=0.95\textwidth]{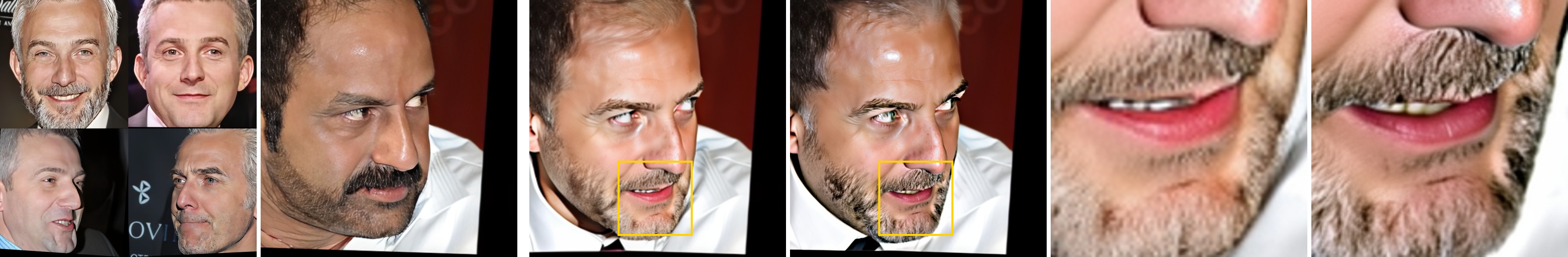}
  \caption{\textbf{The $256$\,px and the $512$\,px model.} Results of both models with the same target, driving image and seed, each at its own resolution; the $256$\,px result is enlarged to $512$\,px (bicubic). The last two columns enlarge the boxed region. The $512$\,px model shows single teeth (top) and single hairs of the beard (bottom). Target: four of its $24$ photos.}
  \label{fig:suppl:res512}
\end{figure*}

\paragraph{The $512$\,px model.}
The $512$\,px model starts from the $256$\,px Stage~2 model and trains for $100$k
steps at a learning rate of $10^{-5}$ with a batch of $16$ on $8$ GPUs, with the
data and the strategies of Stage~2. The rank of the low-rank update is padded
with zeros from $64$ to $256$, since a $512$\,px image has four times as many
latent tokens, and the scaling factor is set to $256$; the blocks of the random
block mask are $32$\,px. Its lighting stream trains for $50$k steps
at a learning rate of $2\times10^{-5}$, with a $32\times32$ lighting grid, so each
cell again covers one latent token of $16\times16$ pixels. At inference, the face mask is dilated by $24$\,px and
occluders by $3$\,px. For scoring, the results of both models are resized to
$448$\,px. Fig.~\ref{fig:suppl:res512} shows them at their own resolution: the
$512$\,px model renders finer detail, such as single teeth and the hairs of a
beard.

\section{Benchmarks}
\label{sec:suppl:bench}

\subsection{Identity benchmark}
\label{sec:suppl:idbench}
\paragraph{Photos.}
The identity benchmark is built from VGGFace2-HQ~\cite{chen2024simswappp}, a version of
VGGFace2~\cite{cao2018vggface2} whose photos are aligned at $512$\,px and
restored with GFPGAN~\cite{wang2021gfpgan}. We take a subset of about $1$k people and
clean each person's photos of identity noise. We build a $k$-nearest-neighbor
graph of the photos ($k{=}8$), with an edge where the mean ArcFace and AdaFace
cosine similarity is at least $0.30$, and keep its largest connected component.
Of these photos, we keep those whose AdaFace similarity to the centroid of the
person's frontal photos is at least $0.75$, with the threshold scaled by yaw.
Each kept photo also has a face-quality score~\cite{chen2024dslfiqa} of at least
$0.5$, a visible nose and mouth, a single face and no sunglasses.

\paragraph{Photos given and held out.}
Each person has $24$ photos that the methods receive and $24$ held-out photos
for scoring identity; all $999$ people of the cleaned subset reach these
numbers. A held-out photo is skipped when its mean cosine similarity to any
given photo is $0.90$ or more, so that the held-out photos are not
near-duplicates of the given ones; where this leaves too few photos, the
threshold rises to $0.95$ and then $0.97$. The given photos follow a fixed
order, and a method with $M$ photos receives the first $M$, so the photos for a
smaller $M$ are part of those for a larger one. The order starts with a
near-frontal photo, picked among the three with the smallest yaw and checked by
eye; the other photos follow in farthest-point order in yaw and pitch, so each
added photo shows the most different pose. Methods that take one photo receive
the near-frontal photo with the highest face-quality score among the $24$
(Sec.~\ref{sec:suppl:baselines}), not the first photo of this order.
The photos come from a source other than our training data, so the two sets
share no images. To check that our results do not depend on people who may also
appear in our training data, we compare the mean face embedding of each target,
over eight of its photos, with the identities of our training data (AdaFace and
ArcFace, averaged). Without the $70$ of the $240$ targets that reach a cosine
similarity of at least $0.5$ to one of them, the ArcFace scores of the methods of
Table~\ref{tab:main} change by less than $0.02$ and their order stays the same
(Table~\ref{tab:suppl:overlap}). Without them and the $32$ of the $176$ celebrity
targets that do, raters prefer our results in $84.4\%$ of the likeness trials
instead of $83.6\%$ (all $14$ methods).
\begin{table}[t]
  \centering\footnotesize
  \setlength{\tabcolsep}{5pt}
  \caption{\textbf{Without the targets closest to our training identities.}
  ArcFace on all $240$ pairs of the identity benchmark, and without the targets
  whose embedding has a cosine similarity of at least $0.5$ ($170$ pairs left) or
  $0.4$ ($165$ left) to an identity of our training data. Italic values exceed a
  real photo of the target.}
  \label{tab:suppl:overlap}
  \begin{tabular}{@{}l ccc@{}}
    \toprule
    Method & All & $<0.5$ & $<0.4$ \\
    \midrule
    SimSwap & 0.500 & 0.496 & 0.495 \\
    FaceDancer & 0.394 & 0.386 & 0.386 \\
    BlendFace & 0.487 & 0.471 & 0.473 \\
    GHOST & 0.548 & 0.545 & 0.545 \\
    InSwapper & \textit{0.715} & \textit{0.714} & \textit{0.714} \\
    HifiFace & 0.528 & 0.525 & 0.529 \\
    Face-Adapter & 0.501 & 0.505 & 0.505 \\
    DreamID-V & 0.550 & 0.558 & 0.558 \\
    \textbf{Ours} & 0.601 & 0.603 & 0.603 \\
    \bottomrule
  \end{tabular}
\end{table}

\paragraph{Conditions.}
The benchmark has $240$ targets in four conditions of $60$, by how much the
target and the person in the driving image differ
(Table~\ref{tab:suppl:conditions}). Sex and age come from
InsightFace~\cite{insightface}, averaged over eight photos of each person, and skin
brightness is the median luminance of the skin that a face-parsing
model~\cite{yu2018bisenet} finds, on a $0$--$255$ scale. Ethnicity is the vote of
a multimodal language model over three photos of each person; we use it only
to form the pairs.
\begin{table}[h]
  \centering\footnotesize
  \caption{\textbf{Conditions of the identity benchmark.} Differences between
  the target and the person in the driving image. Age gap in years; skin
  brightness gap on a $0$--$255$ scale. --: not constrained. $60$ pairs per
  condition.}
  \label{tab:suppl:conditions}
  \begin{tabular}{@{}l c c c c@{}}
    \toprule
    Condition & Sex & Ethnicity & Age gap & Brightness gap \\
    \midrule
    A & same & same & $\le 15$ & $\le 30$ \\
    B & different & same & $\le 15$ & $\le 30$ \\
    C & same & different & $\le 15$ & $\ge 30$ \\
    D & same & same & $\ge 30$ & -- \\
    \bottomrule
  \end{tabular}
\end{table}

\paragraph{Driving images.}
Each target is paired with a photo of another person of the same subset, which
serves as its driving image. Each person appears once across all targets and
driving images. Every method receives the same driving image, a $448$\,px crop
around the face.

\paragraph{Celebrity targets.}
The user study (Sec.~\ref{sec:exp:eval}) also includes $176$ celebrity targets,
whose faces raters may know. We collected their photos from Wikimedia
Commons with a search across fields and ethnicities; each photo keeps its own
license. They have $10$ to $24$
photos each ($60$ of them have $24$) and no held-out photos, so every method
receives all of a celebrity's photos, and the celebrity pairs enter the user
study only. Their driving images show other people, who are not celebrities:
$125$ come from FFHQ~\cite{karras2019stylegan} and $51$ from the driving images of
the identity benchmark, assigned under the rules of the four conditions.

\subsection{Shape benchmark}
\label{sec:suppl:shape:bench}
\begin{figure*}[t]
  \centering
  \edef\ucell{\the\dimexpr 0.0979\textwidth\relax}\edef\ugap{\the\dimexpr 0.0023\textwidth\relax}%
  \makebox[1.00\textwidth][l]{\makebox[\ucell][c]{\scriptsize\shortstack{Render 1}}\hspace{\ugap}\makebox[\ucell][c]{\scriptsize\shortstack{Render 2}}\hspace{\ugap}\makebox[\ucell][c]{\scriptsize\shortstack{Render 3}}\hspace{\ugap}\makebox[\ucell][c]{\scriptsize\shortstack{Render 4}}\hspace{\ugap}\makebox[\ucell][c]{\scriptsize\shortstack{Render 5}}\hspace{\ugap}\makebox[\ucell][c]{\scriptsize\shortstack{Render 6}}\hspace{\ugap}\makebox[\ucell][c]{\scriptsize\shortstack{Render 7}}\hspace{\ugap}\makebox[\ucell][c]{\scriptsize\shortstack{Render 8}}\hspace{\ugap}\makebox[\ucell][c]{\scriptsize\shortstack{Driving}}\hspace{\ugap}\makebox[\ucell][c]{\scriptsize\shortstack{Ground\\truth}}}\\[2pt]
  \includegraphics[width=1.00\textwidth]{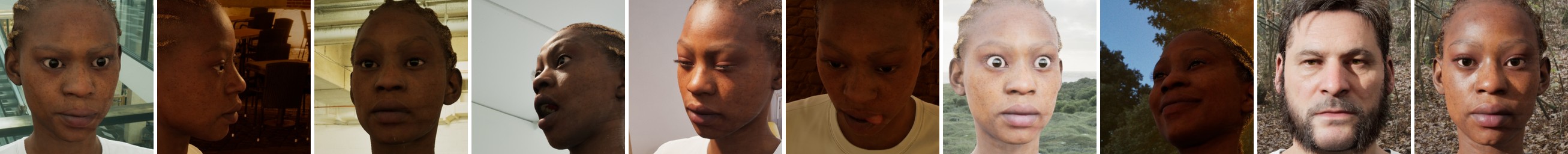}\\[\ugap]
  \includegraphics[width=1.00\textwidth]{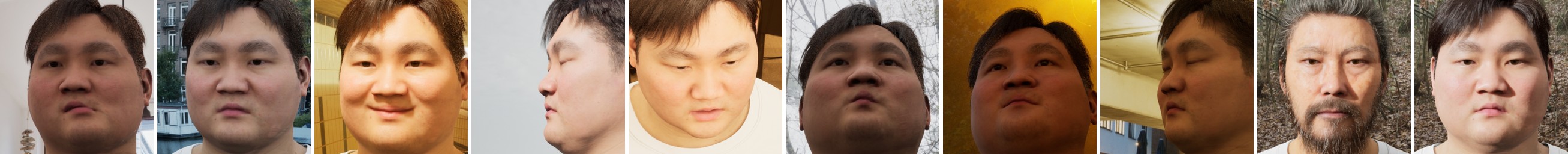}\\[\ugap]
  \includegraphics[width=1.00\textwidth]{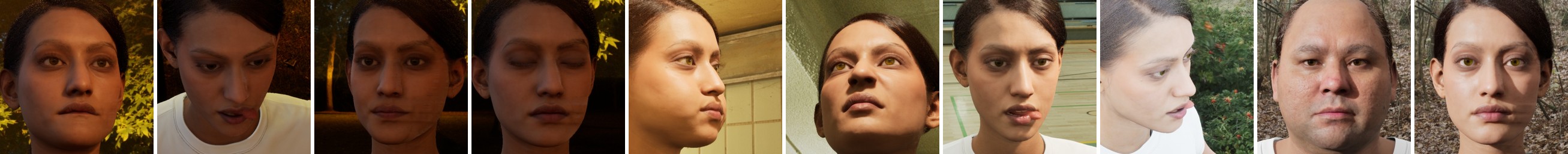}\\[\ugap]
  \includegraphics[width=1.00\textwidth]{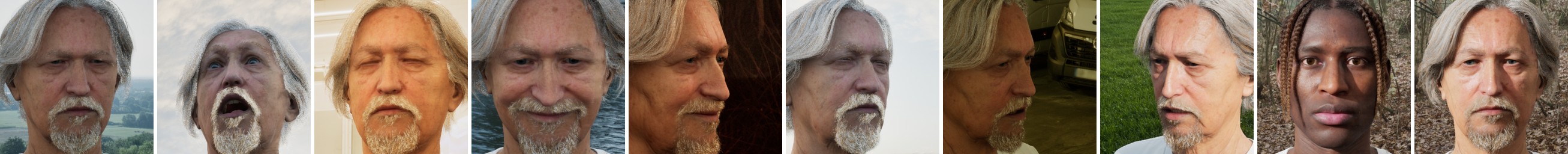}
  \caption{\textbf{The shape benchmark.} Four of the $29$ characters. Left: the first $8$ of the $24$ renders that the methods receive, in their fixed order. Right: a driving image (another character with a neutral expression, facing the camera, under one fixed lighting) and the ground truth of that pair (the target under the same camera, expression and lighting).}
  \label{fig:suppl:ue5data}
\end{figure*}

\paragraph{Characters and renders.}
The shape benchmark has $29$ MetaHuman~\cite{metahuman} characters rendered in
Unreal Engine~5.8. They are stock MetaHuman presets of different sexes, ages and
ethnic groups, built at the highest quality setting. Each character is rendered
at $2048\times1080$\,px with $53$ expressions: a neutral face, $45$ action units
of the Facial Action Coding System~\cite{ekman1976measuring} and $7$ basic
emotions~\cite{ekman1997Universal}. The lighting comes from $30$ HDRI environments
at 4K resolution, indoor and outdoor, by day, in the evening and at night; each is
rendered alone and with an added light source, which gives $60$ lighting
conditions. The cameras cover the front hemisphere with $102$ viewpoints
($\pm90^\circ$ in azimuth and elevation); we use the range of $\pm70^\circ$ in
azimuth and $\pm30^\circ$ in elevation, in which the face stays fully visible.
Each character is the target of $12$ pairs. The driving
images show other characters of the set with a neutral expression, facing the
camera, under one fixed lighting. The ground truth shows the target with the
same camera, expression and lighting. Each method receives $24$ other renders of
the target. Their lighting, expression and camera are drawn at random from
the $60$ lighting conditions, the $45$ action units and $7$ emotions, and the
camera range above. The renders follow a fixed
order: the near-frontal render (head rotation below $20^\circ$) that the
face-quality model rates best comes first, and the others follow in the order
of sampling. Methods that average their identity input use all $24$ renders,
and DreamID-V, which takes one photo, receives the first. For $14$ of the $348$
pairs, the hair of the driving character hid the jawline points of the result,
and we replaced their driving images with other characters of the set.
Fig.~\ref{fig:suppl:ue5data} shows four of the characters with their renders.

\paragraph{Contour distance and leak.}
We detect $21$ points along the jawline with
MediaPipe~\cite{lugaresi2019mediapipe,kartynnik2019facemesh} and leave out the
points that hair covers in the ground truth or in the driving image, so that
every method is scored on the same points. The result, the driving image and the
ground truth share the camera, so we compare them in image coordinates. The
contour distance $s_\mathrm{gt}$ is the mean distance from the jawline points of
the result to the jawline of the ground truth, in units of half the distance
between the eyes. The leak is the slope with which the jawline of the result
follows that of the driving image across the driving images of a target,
measured normal to the jawline of the ground truth ($0$: no leak; $1$: copied).

\paragraph{3D error.}
The 3D error compares the face shapes that
Deep3DFaceRecon~\cite{deng2019accurate} reconstructs from the result and from the
ground truth. Before reconstruction, we paint gray, in both images, the regions
that hair covers in the ground truth or in the driving image. The 3D error is
the mean distance between the two reconstructed faces, with expression and pose
removed, over the vertices outside the painted region, in percent of the
distance between the eyes.

\paragraph{Reference values.}
The same character rendered under other lighting scores $0.024$ in contour
distance and $1.14$ in 3D error, which is the noise level of the two metrics.
The driving image itself scores $0.128$ and $3.49$.

\paragraph{Arrival ratio.}
As a further reading of the same points, the arrival ratio
$s_\mathrm{drv}/(s_\mathrm{drv}+s_\mathrm{gt})$ uses the distance
$s_\mathrm{drv}$ from the jawline points of the result to the jawline of the
driving image. It is $0$ for a result on the jawline of the driving image and $1$
for a result on the jawline of the ground truth.

\paragraph{Versions of the 3D error.}
Two simpler versions of the 3D error compare the $80$ identity coefficients of
the reconstructions instead of the meshes: (a) on the images as they are, and
(b) after the hair region is painted gray. Table~\ref{tab:suppl:d3dversions}
compares the three versions with their noise level, the driving image and a
pure change of face shape. All three separate the driving image from the noise
level, by a factor of $2.3$ to $3.1$. A warp of the jawline alone moves the mesh
distance little above its noise level, while a change of the width-to-height
ratio of the face moves it close to or above the driving image. The 3D error
therefore responds mainly to the proportions of the face, and complements the
contour distance, which reads the jawline.
\begin{table}[h]
  \centering\footnotesize
  \setlength{\tabcolsep}{3.5pt}
  \caption{\textbf{Versions of the 3D error} on the shape benchmark. (a), (b):
  L2 distance between the $80$ identity coefficients of the reconstructions of
  the result and the ground truth, without and with the hair region painted gray.
  (c): the mesh distance of Sec.~\ref{sec:suppl:shape:bench}, in percent of the
  distance between the eyes. Noise: the same character under one other lighting
  ($29$ pairs). Driving: the driving image scored as a result ($348$). Warp: the
  ground truth of each character warped in face shape only, with the jawline
  scaled by $0.88$ or $1.12$, or the width-to-height ratio by $0.80$ or $1.25$
  ($29$ each).}
  \label{tab:suppl:d3dversions}
  \begin{tabular}{@{}l c c cc cc@{}}
    \toprule
    & & & \multicolumn{2}{c}{Jawline warp} & \multicolumn{2}{c}{Ratio warp} \\
    \cmidrule(lr){4-5} \cmidrule(l){6-7}
    Version & Noise & Driving & $0.88$ & $1.12$ & $0.80$ & $1.25$ \\
    \midrule
    (a) coefficients & 2.56 & 6.32 & 2.84 & 2.62 & 4.07 & 3.68 \\
    (b) coefficients, painted & 2.59 & 6.05 & 2.12 & 1.79 & 4.12 & 3.82 \\
    (c) mesh (ours) & 1.14 & 3.49 & 1.46 & 1.22 & 4.03 & 3.38 \\
    \bottomrule
  \end{tabular}
\end{table}

\section{Baselines}
\label{sec:suppl:baselines}
Table~\ref{tab:suppl:baselines} lists the code, checkpoint, working resolution
and use of the target's photos for each method. Every method receives the same
driving image, the $448$\,px crop of Sec.~\ref{sec:suppl:idbench}, and the
original photos of the target. Methods whose face detector needs more context
receive the crop with a black border of $0.25$ to $0.3$ of its size, and their
results are pasted back into the $448$\,px frame. Outputs at another resolution
are resized to $448$\,px.

\begin{table*}[t]
  \centering\footnotesize
  \setlength{\tabcolsep}{4pt}
  \caption{\textbf{Baselines.} Public code and checkpoint of each method, the
  resolution at which it generates the face, and how it uses the target's
  photos. Mean of 24: the identity input is averaged over the $24$ photos of
  Sec.~\ref{sec:suppl:idbench}. One photo: the near-frontal photo that a
  face-quality model~\cite{chen2024dslfiqa} rates best.}
  \label{tab:suppl:baselines}
  \begin{tabular}{@{}l l c l@{}}
    \toprule
    Method & Code and checkpoint & Resolution & Target's photos \\
    \midrule
    SimSwap~\cite{chen2020simswap} & ``people'' model, face-parsing paste-back & $224$ & mean of 24 ArcFace embeddings, renormalized \\
    FaceDancer~\cite{rosberg2023facedancer} & configuration C, HQ & $256$ & mean of 24 ArcFace embeddings \\
    BlendFace~\cite{shiohara2023blendface} & AEI-Net with the BlendFace encoder & $256$ & mean of 24 unit embeddings \\
    GHOST~\cite{groshev2022ghost} & U-Net generator, 2 blocks, no super-resolution & $256$ & mean of 24 embeddings \\
    InSwapper~\cite{inswapper} & \texttt{inswapper\_128}, InsightFace \texttt{buffalo\_l} & $128$ & mean of 24 normalized ArcFace embeddings \\
    HifiFace~\cite{wang2021hififace} & maum-ai reimplementation, trained on VGGFace2 & $256$ & mean of 24 ArcFace embeddings and 3D coefficients \\
    Face-Adapter~\cite{han2024faceadapter} & released checkpoints, \texttt{sd-vae-ft-mse} & $512$ & mean of 24 ArcFace embeddings and 3D coefficients \\
    DreamID-V~\cite{guo2025dreamidv} & Wan-1.3B model with DWPose & $832\times480$ & one photo \\
    \midrule
    DiffSwap~\cite{zhao2023diffswap} & DDIM, $200$ steps, target scale $0.01$ & $256$ & one photo \\
    REFace~\cite{baliah2025reface} & FFHQ configuration & $512$ & one photo \\
    DiffFace~\cite{kim2022diffface} & $4$ iterations & $256$ & one photo \\
    E4S~\cite{liu2023e4s} & official pipeline & $1024$ & one photo \\
    GPT-Image & \texttt{gpt-image-2.5-sunburst}, image-editing endpoint & $1024$ & first 8 photos \\
    Gemini-Pro & \texttt{gemini-3-pro-image} (Nano Banana Pro), batch endpoint & $1024$ & one photo \\

    \bottomrule
  \end{tabular}
\end{table*}

\paragraph{Averaging the identity input.}
SimSwap, FaceDancer, BlendFace, GHOST, InSwapper, HifiFace and Face-Adapter take
the identity embedding of one photo. We extend them to $M$ photos by averaging
the embeddings, normalized as each method normalizes its own; HifiFace and
Face-Adapter also average the 3D identity coefficients that they take with the
embedding. This extension is ours, and it raises their recognition scores
(Sec.~\ref{sec:exp:msweep}), so their best setting uses all $24$ photos.
HifiFace has no official code. We use a public
reimplementation and choose its crop scale, $1.0$, on $40$ held-out pairs with a
rule fixed before scoring.

\paragraph{Methods that take one photo.}
DreamID-V, DiffSwap, REFace, DiffFace, E4S and Gemini-Pro receive one photo: the
photo with the highest face-quality score among those whose head rotation,
estimated by 6DRepNet~\cite{hempel2022sixdrepnet}, is below $20^\circ$, or the
most frontal photo if there is none. DreamID-V is a video model. We repeat the
driving image as a still video of $21$ frames and keep frame $18$. On the shape
benchmark, the methods that average use all $24$ renders and the others the
first render (Sec.~\ref{sec:suppl:shape:bench}).

\paragraph{Commercial image models.}
We prompt two commercial image models to transfer the identity: GPT-Image with the
driving image and the first $8$ photos of the target, and Gemini~3 Pro Image
(Nano Banana Pro) with the driving image and one photo, chosen as above. Both receive the same
prompt, with $n{=}9$ for GPT-Image and $n{=}2$ for Gemini-Pro:
\begin{quote}\small
``Swap the face in image 1 with the person shown in images 2 to $\{n\}$.
Reproduce their face shape, forehead, eyebrows, eyes, nose, mouth, jaw, ears,
skin tone, age and facial hair as faithfully as possible, using all reference
photos. Keep image 1's head pose, gaze, expression, lighting, background and its
hairstyle and hair color (same hair as image 1; do not copy the reference's
hair or headwear). Do not beautify; keep photographic realism.''
\end{quote}
Gemini-Pro receives each image after a caption, ``Image 1 (driving photo, edit
this one):'' and ``Image 2 (target person):''. We chose this prompt among seven
by how well the results kept the hair of the driving image and resembled the
target. For GPT-Image, a sweep
over $1$, $3$, $5$, $8$ and $15$ photos on $48$ pairs showed no gain in identity
beyond $3$ photos, and we use $8$. For Gemini-Pro, three photos gave worse
results than one. Gemini-Pro uses its default sampling settings. Both models also
run on the shape benchmark, GPT-Image with the first $8$ renders of the target and
Gemini-Pro with the first one (Table~\ref{tab:suppl:shapeall}).

\paragraph{Missing results.}
Some methods return no result for a few driving images, and every table reports
the number $n$ of results scored; missing results are neither retried nor imputed.
REFace, DiffFace and E4S miss the same $17$ of the $240$ pairs, where their face
aligner finds no face in the driving image; DiffFace uses the aligner of REFace.
DiffSwap misses $14$, $12$ of them among the same $17$. GPT-Image returned no
image for $16$ pairs; the reason was not recorded for each pair.

\section{Metrics}
\label{sec:suppl:metrics}
This section defines the metrics of Table~\ref{tab:main} and of the supplement,
and shows where each can mislead. All are computed on the $240$ pairs of the
identity benchmark, and each method is scored on the results it returns.

\paragraph{Identity.}
Identity is the mean cosine similarity between the recognition embeddings of
the result and of the $24$ held-out photos of the target. We use three
recognizers: ArcFace~\cite{deng2019arcface} (IResNet-50),
AdaFace~\cite{kim2022adaface} (IR-101) and FaceNet~\cite{schroff2015facenet}
(Inception-ResNet-v1). The FaceNet model we use was trained on VGGFace2~\cite{cao2018vggface2},
from which the targets of the identity benchmark also come. Table~\ref{tab:suppl:idref} gives
two reference points for each, scored the same way. The first is the first given
photo of each target, a real photo of the person and the level of a perfect
result. The second is the same photo scored against the held-out photos of
another target. A result can score above
the real photo, as InSwapper does under AdaFace and ArcFace. Such a score
reflects how the recognizer reads the result, not more resemblance than a photo of
the person, so we set it in italics and do not rank it. AdaFace is part of the
identity loss of Stage~1 (Sec.~\ref{sec:method}), so the main paper reports
ArcFace, which none of our losses uses. On the celebrity targets, where both are
reported, scoring against the given photos instead of held-out ones keeps the
order of the methods (Table~\ref{tab:suppl:celeb}).
\begin{table}[h]
  \centering\footnotesize
  \caption{\textbf{Reference points of the identity scores.} Mean cosine
  similarity to the $24$ held-out photos of the target ($240$ targets).}
  \label{tab:suppl:idref}
  \begin{tabular}{@{}l c c c@{}}
    \toprule
    & AdaFace & ArcFace & FaceNet \\
    \midrule
    Real photo of the target & 0.656 & 0.677 & 0.736 \\
    Photo of another person & 0.012 & 0.011 & 0.012 \\
    \bottomrule
  \end{tabular}
\end{table}

\paragraph{Expression.}
Expression is the L2 distance between the valence and arousal that
EmoNet~\cite{toisoul2021emonet} predicts for the result and for the driving image.
EmoNet reads a square crop of $1.2$ times the face box, resized to $256$\,px,
with the face detected in each image separately. We also report three measures
that compare facial parameters instead of emotion: the expression error of
SMIRK~\cite{retsinas2024smirk}, the mean absolute difference of its $50$
expression, $2$ eyelid and $3$ jaw parameters; the mean absolute difference of
the $52$ MediaPipe blendshapes~\cite{lugaresi2019mediapipe}; and the L2 distance
between the $64$ expression coefficients of
Deep3DFaceRecon~\cite{deng2019accurate}. Over the nine methods of
Table~\ref{tab:main}, they rank the methods similarly to EmoNet, with Spearman
correlations of $0.93$ (SMIRK), $0.97$ (blendshapes) and $0.93$
(Deep3DFaceRecon). Under all four, our results follow the expression less closely
than those of SimSwap and FaceDancer, and more closely than those of GHOST,
HifiFace, Face-Adapter and DreamID-V. Measured against the driving image, these
scores reward a result that keeps the face of the driving image
(Sec.~\ref{sec:suppl:exprfloor}). Two narrower measures agree less: the
mouth opening ($0.72$) and the intensities of $12$ action units from
LibreFace~\cite{chang2024libreface} ($0.72$), which also respond to lighting.

\paragraph{Pose.}
Pose is the geodesic angle between the head rotations predicted for the result and
for the driving image, under two estimators. 6DRepNet~\cite{hempel2022sixdrepnet}
reads the whole $448$\,px frame, resized to $224$\,px. Its weights are the ones
that supervised the head pose of the geometry branch in its pretraining
(Sec.~\ref{sec:method:arch}), so the main paper reports HopeNet~\cite{ruiz2018hopenet}, which
reads a crop around the face box of the driving image, the same crop for both
images. HopeNet's yaw, pitch and roll give the rotation. The two estimators rank
the nine methods of Table~\ref{tab:main} alike (Spearman $0.97$), and both rank
our results fifth.

\paragraph{Lighting.}
Lighting is the Pearson correlation between the low-pass luminance of the result
and of the driving image inside the face. We convert both $256$\,px images to
luma, blur them with a Gaussian of $\sigma{=}8$\,px, and compare them over the
face region that a face-parsing model~\cite{yu2018bisenet} finds in the driving
image: skin, brows, eyes, glasses, nose, mouth and lips, without ears, neck and
hair. The score rewards a result that keeps the face shape and skin of the
driving image: the driving image itself would score $1$, and SimSwap, FaceDancer, BlendFace, GHOST,
InSwapper and HifiFace, which keep most of its jawline (leak $0.78$ to $0.95$),
score $0.90$ to $0.96$. A high score therefore does not show on its own that a
result keeps the lighting while it changes the identity. Over the same region,
Table~\ref{tab:suppl:allmetrics} also gives the absolute difference of the mean
luminance ($\Delta L$, $0$--$255$) and the L2 distance of the mean CIELAB $a^*$
and $b^*$ ($\Delta ab$) between the result and the driving image.

\paragraph{Realism.}
Realism is measured without a reference. FaceScore~\cite{liao2024facescore} is a
learned reward for face quality, and DSL-FIQA~\cite{chen2024dslfiqa} is a
face-quality model, which also filters the photos of the identity benchmark and
picks the photo given to the methods that take one
(Secs.~\ref{sec:suppl:idbench} and~\ref{sec:suppl:baselines}); it scores a crop
around the face box of the driving image. Raters also judge realism in the user study
(Sec.~\ref{sec:suppl:human}).

\subsection{Expression, pose and lighting measures under a shape change}
\label{sec:suppl:exprfloor}
\label{tab:suppl:exprfloor}%
A change of face shape alone moves each estimator of expression and pose that
we use.
Table~\ref{tab:suppl:shapebias} scores images whose expression and pose are the
same by construction. We warp each driving image of the identity benchmark in
face shape only, with the warps of self-augmentation
(Sec.~\ref{sec:method:strategies}), and score it against the original. On the
shape benchmark, we compare two different characters rendered with the same
camera, expression and lighting, and one character under two lightings. A warp
of the jawline by $12\%$ moves EmoNet by $45$ to $56\%$ of the spread between
the methods of Table~\ref{tab:main}, and the three other expression measures by
about $100$ to $140\%$ of their spread. Two different characters with the same
expression, pose and lighting differ under each expression measure by more
than the whole spread between the methods ($1.6$ to $3.7$ times), and under the
two pose estimators by $42$ to $60\%$ of it; this is $2$ to $4$ times as much as
one character under two lightings. A difference between two methods on these measures therefore mixes
expression or pose with the difference in face shape between their results.

The lighting measures respond in the same way (last three rows of
Table~\ref{tab:suppl:shapebias}). Two characters under the same lighting reach a
lighting correlation of only $0.80$, below eight of the nine methods of
Table~\ref{tab:main}, and their mean luminance differs by $30.3$, more than for
any of these methods. This difference follows skin tone: it is $10.4$ for the
third of the pairs with the closest skin tones and $55.0$ for the third with the
farthest, while the correlation changes only from $0.82$ to $0.79$. A result
that changes the person under the same light therefore loses lighting score,
and $\Delta L$ and $\Delta ab$ measure skin tone as much as lighting.

On the shape benchmark, the ground truth is known, so each result can also be
scored against it instead of against the driving image
(Table~\ref{tab:suppl:vsgt}). Against the driving image, the ground truth itself
would rank behind five to eight of the nine methods of Table~\ref{tab:main} on
each measure. Against the ground truth, the order of the methods changes:
Spearman's $\rho$ between the two orders is $0.80$ for EmoNet, $0.30$ for SMIRK,
$-0.23$ for HopeNet, $0.43$ for 6DRepNet and $0.12$ for the lighting
correlation. Our results rank first of the nine on SMIRK, HopeNet and 6DRepNet,
third on lighting and fourth on EmoNet against the ground truth, and fifth,
seventh, sixth, seventh and second against the driving image. The driving images
of this benchmark face the camera with a neutral expression
(Sec.~\ref{sec:suppl:shape:bench}), so these scores cover a narrow range of
expression and pose. Within it, a score against the driving image rewards a
result for keeping the face of the driving image.
\begin{table*}[t]
  \centering\footnotesize
  \setlength{\tabcolsep}{4pt}
  \caption{\textbf{Expression, pose and lighting measures under a change of face
  shape alone.} Each column compares two images with the same expression and pose,
  with the measures of Sec.~\ref{sec:suppl:metrics}. In the lighting rows, Other
  character keeps the lighting and changes the person, and Other lighting keeps
  the person and changes the lighting; Self-augmentation is left out there, since
  its black fill lies partly inside the face region. Jawline: the driving image
  ($240$) with its jawline scaled by $0.88$ or $1.12$ by moving least squares.
  Self-augmentation: three warps per driving image drawn as in training, which
  also change the width-to-height ratio and leave black fill over $13\%$ of
  the frame on average. Other character: two characters of the shape benchmark in
  the same pose, expression, camera and lighting ($348$ pairs). Other lighting:
  one character under two lightings ($29$). Methods: the lowest and highest
  value among the nine methods of Table~\ref{tab:main}. HopeNet: the geodesic angle
  of Table~\ref{tab:main}.}
  \label{tab:suppl:shapebias}
  \begin{tabular}{@{}l cc c cc c@{}}
    \toprule
    & \multicolumn{3}{c}{Identity benchmark, warped} & \multicolumn{2}{c}{Shape benchmark} & \\
    \cmidrule(lr){2-4} \cmidrule(lr){5-6}
    Measure & Jawline $0.88$ & Jawline $1.12$ & Self-augmentation & Other character & Other lighting & Methods \\
    \midrule
    HopeNet (deg) & 2.30 & 2.23 & 3.11 & 3.32 & 1.45 & 2.03--7.52 \\
    6DRepNet (deg) & 1.55 & 1.40 & 1.79 & 2.62 & 1.06 & 1.99--8.18 \\
    EmoNet & 0.094 & 0.075 & 0.123 & 0.263 & 0.095 & 0.205--0.372 \\
    SMIRK & 0.311 & 0.311 & 0.396 & 0.553 & 0.263 & 0.420--0.696 \\
    Blendshapes & 0.026 & 0.023 & 0.027 & 0.050 & 0.013 & 0.036--0.058 \\
    Deep3DFaceRecon & 1.14 & 1.04 & 1.17 & 2.99 & 1.29 & 2.48--3.29 \\
    \midrule
    Lighting correlation $\uparrow$ & 0.943 & 0.962 & -- & 0.802 & 0.930 & 0.52--0.96 \\
    $\Delta L$ & 5.2 & 3.4 & -- & 30.3 & 24.1 & 2.4--23.9 \\
    $\Delta ab$ & 1.20 & 0.46 & -- & 3.93 & 14.77 & 0.82--5.88 \\
    \bottomrule
  \end{tabular}
\end{table*}
\begin{table*}[t]
  \centering\footnotesize
  \setlength{\tabcolsep}{4pt}
  \caption{\textbf{Expression, pose and lighting against the driving image (D)
  or the ground truth (GT)} on the shape benchmark ($348$ pairs), for the nine
  methods of Table~\ref{tab:main}. The first row scores the ground truth against
  the driving image. The face box of HopeNet and the region of the lighting
  correlation come from the image that a result is compared with. Bold: best of
  the nine in each column.}
  \label{tab:suppl:vsgt}
  \begin{tabular}{@{}l cc cc cc cc cc@{}}
    \toprule
    & \multicolumn{2}{c}{EmoNet $\downarrow$} & \multicolumn{2}{c}{SMIRK $\downarrow$}
    & \multicolumn{2}{c}{HopeNet (deg) $\downarrow$} & \multicolumn{2}{c}{6DRepNet (deg) $\downarrow$}
    & \multicolumn{2}{c}{Lighting corr.\ $\uparrow$} \\
    \cmidrule(lr){2-3} \cmidrule(lr){4-5} \cmidrule(lr){6-7} \cmidrule(lr){8-9} \cmidrule(lr){10-11}
    Method & D & GT & D & GT & D & GT & D & GT & D & GT \\
    \midrule
    Ground truth & 0.262 & -- & 0.550 & -- & 3.33 & -- & 2.64 & -- & 0.802 & -- \\
    \midrule
    SimSwap & 0.221 & 0.256 & 0.411 & 0.457 & \textbf{1.60} & 3.42 & \textbf{1.52} & 3.12 & 0.970 & 0.847 \\
    FaceDancer & 0.203 & 0.263 & 0.409 & 0.450 & 1.90 & 3.59 & 1.61 & 3.37 & \textbf{0.975} & 0.850 \\
    BlendFace & \textbf{0.142} & 0.215 & \textbf{0.348} & 0.440 & 2.14 & 3.61 & 1.75 & 3.00 & 0.952 & 0.847 \\
    GHOST & 0.389 & 0.367 & 0.558 & 0.539 & 1.81 & 3.44 & 1.74 & 3.08 & 0.957 & 0.832 \\
    InSwapper & 0.201 & \textbf{0.189} & 0.448 & 0.385 & 2.39 & 3.41 & 1.82 & 3.21 & 0.953 & \textbf{0.880} \\
    HifiFace & 0.384 & 0.413 & 0.776 & 0.736 & 2.71 & 3.49 & 2.49 & 3.43 & 0.945 & 0.826 \\
    Face-Adapter & 0.303 & 0.225 & 0.578 & 0.454 & 3.43 & 3.37 & 2.73 & 3.24 & 0.907 & 0.876 \\
    DreamID-V & 0.363 & 0.320 & 0.606 & 0.401 & 3.52 & 3.53 & 3.77 & 3.96 & 0.497 & 0.592 \\
    Ours & 0.186 & 0.234 & 0.498 & \textbf{0.337} & 2.96 & \textbf{3.28} & 2.10 & \textbf{2.64} & 0.915 & 0.872 \\
    \bottomrule
  \end{tabular}
\end{table*}

\section{User study}
\label{sec:suppl:human}
Human preference is one of our two measures of identity (Sec.~\ref{sec:exp:eval}),
together with recognition scores.
Each trial compares our result with the result of one baseline on the same driving
image. The rater sees three of the target's photos that the methods receive,
the driving image and the two results side by side. The rater then answers two
questions by clicking on one of the results: which result looks more like the
target (likeness), and which looks more realistic (realism). There is no tie
option. The side of our result alternates across the panels of each method and
was fixed when the panels were built, so our result is on the left in half of
them.
The raters read the following instructions (verbatim):
\begin{quote}\small
\emph{Goal.} Evaluate and compare two face-swap methods based on two criteria:
Likeness: How well the generated face preserves the identity of the target
person. Realism: How natural and realistic the generated face looks.
\emph{Image Reference Guide.} Top-Left Images (Green Border): Photos showing the
target person's identity. Top-Right Image (Red Border): The original driving
image providing the facial expression and background. Bottom Row Images: The
generated results labeled LEFT and RIGHT to evaluate.
\emph{Task Instructions.} For each criteria (Likeness Preferred and Realism
Preferred), select ONE image from the bottom row (LEFT or RIGHT). You may select
the same image for both criteria or a different image for each. Do NOT make
choices from the top row images. Do NOT select multiple images per criteria.
\end{quote}

Each method has up to $416$ panels. The $240$ panels of the identity benchmark show
the results scored in Table~\ref{tab:main} and
Table~\ref{tab:suppl:allmetrics}, and the other $176$ show results with celebrity
targets, who have $10$ to $24$ photos each. Methods that returned no result for
some driving images have fewer panels ($387$ at least). Each method runs at its
best setting (Sec.~\ref{sec:suppl:baselines}).

The study ran in two batches on Labelbox. In the first, $30$ raters
judged $50$ panels each, against seven of the eight baselines of
Table~\ref{tab:main}. In the second, $30$ raters, $18$ of whom also took part in
the first batch, judged about $50$ panels each, against HifiFace and the six
other methods of Sec.~\ref{sec:suppl:baselines}. Over both batches, $42$ raters
gave $2{,}983$ likeness and $2{,}988$ realism labels. Labels of one rater are
correlated, so the $95\%$ intervals come from a bootstrap over raters with
$5{,}000$ resamples.
The raters belong to the labeling workforce of the platform. In each batch,
every rater judged the $50$ panels of one project, and each panel of a project
was judged once; the panels were spread over $60$ projects, so $530$ of them
were judged in more than one. Every panel compares our result with that of a
baseline. A label took $79$\,s (first batch)
and $66$\,s (second batch) at the median. Raters clicked the left result in
$1{,}506$ and the right result in $1{,}477$ of the likeness trials.

\begin{table}[t]
  \centering\footnotesize
  \setlength{\tabcolsep}{4pt}
  \caption{\textbf{User study}: share of trials in which raters chose our result
  (\%), with $95\%$ intervals from a bootstrap over raters. Likeness: which result
  looks more like the target. Realism: which result looks more realistic. $n$:
  labels.}
  \label{tab:suppl:human}
  \begin{tabular}{@{}l c c c c@{}}
    \toprule
    Baseline & Likeness & $n$ & Realism & $n$ \\
    \midrule
    SimSwap      & 96.5 [93.1, 99.1] & 200 & 97.5 [94.8, 99.5] & 200 \\
    FaceDancer   & 80.7 [74.2, 86.8] & 207 & 62.1 [54.4, 70.2] & 206 \\
    BlendFace    & 92.4 [88.0, 96.2] & 210 & 92.0 [87.2, 95.9] & 212 \\
    GHOST        & 98.7 [97.1, 100]  & 226 & 99.6 [98.6, 100]  & 226 \\
    InSwapper    & 97.6 [95.3, 99.5] & 210 & 97.6 [94.5, 100]  & 210 \\
    HifiFace     & 100 [100, 100]    & 235 & 100 [100, 100]    & 236 \\
    Face-Adapter & 73.4 [67.5, 79.0] & 229 & 78.6 [73.3, 83.8] & 229 \\
    DreamID-V    & 79.4 [74.4, 84.4] & 209 & 81.3 [75.2, 87.2] & 209 \\
    DiffSwap     & 91.9 [87.5, 95.7] & 197 & 85.8 [77.9, 92.2] & 197 \\
    REFace       & 81.7 [75.1, 87.5] & 218 & 77.5 [69.6, 84.5] & 218 \\
    DiffFace     & 90.6 [85.6, 94.8] & 212 & 85.8 [79.2, 91.7] & 212 \\
    E4S          & 89.1 [84.1, 93.6] & 211 & 84.8 [79.1, 89.8] & 210 \\
    GPT-Image & 37.4 [29.8, 45.6] & 211 & 15.0 [9.5, 21.2] & 214 \\
    Gemini-Pro   & 60.6 [53.2, 67.5] & 208 & 39.2 [31.2, 46.9] & 209 \\
    \midrule
    Eight of Table~\ref{tab:main} & 89.9 [88.2, 91.5] & 1726 & 88.8 [87.2, 90.4] & 1728 \\
    Other six    & 75.1 [71.9, 78.2] & 1257 & 64.4 [60.4, 68.0] & 1260 \\
    All fourteen & 83.6 [81.7, 85.5] & 2983 & 78.5 [75.9, 80.8] & 2988 \\
    \bottomrule
  \end{tabular}
\end{table}

\begin{table*}[t]
  \centering\footnotesize
  \setlength{\tabcolsep}{5pt}
  \caption{\textbf{User study by set of targets}: share of trials in which
  raters chose our result (\%), with $95\%$ intervals from a bootstrap over
  raters. IB: identity benchmark. Celeb.: celebrity targets.}
  \label{tab:suppl:humanset}
  \begin{tabular}{@{}l cc cc@{}}
    \toprule
    & \multicolumn{2}{c}{Likeness} & \multicolumn{2}{c}{Realism} \\
    \cmidrule(lr){2-3} \cmidrule(l){4-5}
    Baselines & IB & Celeb. & IB & Celeb. \\
    \midrule
    Eight of Table~\ref{tab:main} & 90.3 [88.2, 92.1] & 89.3 [86.6, 91.9] & 88.9 [86.9, 91.0] & 88.6 [85.5, 91.3] \\
    Other six    & 75.1 [70.7, 79.3] & 75.1 [71.4, 79.0] & 63.1 [58.9, 66.9] & 66.1 [60.8, 71.3] \\
    All fourteen & 84.0 [81.6, 86.3] & 83.2 [80.4, 85.6] & 78.3 [75.7, 80.7] & 78.8 [75.5, 82.2] \\
    \bottomrule
  \end{tabular}
\end{table*}

Raters judge our result more like the target than the result of each method
(Table~\ref{tab:suppl:human}). Against the eight baselines of
Table~\ref{tab:main}, they choose our result in $89.9\%$ of the trials, from
$73.4\%$ against Face-Adapter to $100\%$ against HifiFace. Against the four
other methods, the share ranges from $81.7\%$ (REFace) to $91.9\%$
(DiffSwap). The two general image models compare differently: raters choose our
result over that of Gemini-Pro in $60.6\%$ of the trials, and the result of
GPT-Image over ours in $62.6\%$. GPT-Image is preferred on both sets of
targets, where our result is chosen in $34.4\%$ (identity benchmark) and $41.6\%$
(celebrities) of the trials. Raters also find the result of GPT-Image more
realistic in $85.0\%$ of the trials, and their likeness answers often follow
their realism answers (below).

The preference holds for every rater. Each of the $30$ raters of the first batch
chooses our result in $75.5\%$ to $96.0\%$ of their trials against the eight
baselines (median $90.0\%$). Each of the $30$ raters of the second batch
chooses it in $55.3\%$ to $90.9\%$ of their trials against the six other methods
(median $74.4\%$). The share is also about the same on the identity benchmark
and on the celebrity targets, whose faces raters may know
(Table~\ref{tab:suppl:humanset}).

Realism, the second question, gives a similar picture for the methods
and a different one for the image models. Against the eight baselines, raters
choose our result as more realistic in $88.8\%$ of the trials. The lowest share
is against FaceDancer ($62.1\%$), and the others range from $78.6\%$ to
$100\%$. Against the image models, raters choose the result of GPT-Image as more
realistic in $85.0\%$ of the trials and that of Gemini-Pro in $60.8\%$.
FaceScore also ranks these two above our results ($5.14$ and $4.98$, against
$4.62$; Table~\ref{tab:suppl:allmetrics}). It ranks DiffFace ($4.72$) and E4S
($4.70$) above ours as well, whereas raters choose our result as more realistic
in $85.8\%$ and $84.8\%$ of the trials against them. Per rater, the realism
share ranges from $78.6\%$ to $95.0\%$ against the eight baselines (first
batch), and from $37.5\%$ to $80.0\%$ against the six other methods, where $26$
of the $30$ raters choose our result in more than half of their trials.

Likeness varies little across the four conditions of the pairs
(Sec.~\ref{sec:suppl:bench}), on both sets of targets: against the eight
baselines, raters choose our result in $90.0\%$ (A), $90.8\%$ (B), $90.0\%$ (C)
and $88.8\%$ (D) of the trials. For realism, the share is lowest in condition A
($85.2\%$, against $89.7\%$ to $90.3\%$ in B--D). Of the panels that more than
one rater judged, two raters chose the same result in $82.0\%$ of the pairs of
raters for likeness ($530$ panels) and in $83.0\%$ for realism ($532$ panels).
Because most labels favor our result, two raters would agree in $72.6\%$ and
$66.3\%$ of the pairs by chance, so the agreement beyond chance ($\kappa$) is
$0.34$ and $0.50$.

The two answers often go together: in $86.6\%$ of the trials, raters pick the
same result for likeness and for realism. When they find our result more
realistic, they choose it as more like the target in $94.7\%$ of these trials
($n=2{,}344$); when they find the other result more realistic, in $43.0\%$
($n=637$). Against GPT-Image, raters find its result more realistic in $179$ of
$211$ trials, and in these they choose ours as more like the target in
$28.5\%$.

\section{Comparison with prior work}
\label{sec:suppl:allmetrics}
\label{tab:suppl:old93}%
Table~\ref{tab:suppl:allmetrics} gives every metric on the identity benchmark for
the twelve compared methods, the two commercial image models and four variants of
our model. Table~\ref{tab:suppl:idcond} splits the identity scores by condition,
and Table~\ref{tab:suppl:celeb} gives the metrics on the celebrity targets of the
user study (Sec.~\ref{sec:suppl:human}). Sec.~\ref{sec:suppl:metrics} defines the
metrics.

\paragraph{Identity.}
Under AdaFace and ArcFace, our results score highest among the results below a real
photo of the target. FaceNet was trained on VGGFace2, from which the targets of
the benchmark also come (Sec.~\ref{sec:suppl:metrics}); under it, InSwapper
($0.706$), Face-Adapter ($0.666$) and DreamID-V ($0.652$) score above ours
($0.647$). With the face+hair mask, our results score
highest under AdaFace and ArcFace and tie with InSwapper under FaceNet. Given one photo instead of $24$, every
averaging baseline loses $0.06$ to $0.11$ in ArcFace, while its expression, pose,
lighting and realism scores stay about the same, so averaging is its best
setting. Our scores change little across the four conditions
(Table~\ref{tab:suppl:idcond}).

\paragraph{Expression, pose, lighting and realism.}
The SMIRK expression error and the blendshape distance rank the methods like the
EmoNet distance of the main paper, and 6DRepNet ranks them like HopeNet.
Our results differ most from the driving image in mean luminance ($\Delta L$
$23.9$, against $2.4$ to $17.3$). $\Delta L$ follows skin tone as much as lighting
(Sec.~\ref{sec:suppl:exprfloor}), and our results make the skin lighter
(Sec.~\ref{sec:limitations}).
Among the methods of Table~\ref{tab:main}, our results have the highest FaceScore;
DiffFace, E4S and the two image models score higher, and GPT-Image also has the
highest DSL-FIQA. On the celebrity targets (Table~\ref{tab:suppl:celeb}),
InSwapper again scores above the others under ArcFace, against the given
photos and against held-out ones.

\begin{table*}[t]
  \centering\scriptsize
  \setlength{\tabcolsep}{2.6pt}
  \caption{\textbf{All metrics on the identity benchmark} ($240$ pairs). Each
  method runs at its best setting; the rows marked $M{=}1$ give the averaging
  baselines one photo of the target instead of $24$. $M$: photos of the target
  given to the method. $n$: results scored; the others returned no image or failed
  face alignment (some estimators score fewer results, see the text). Identity: mean
  cosine similarity to the $24$ held-out photos of the target. Expression: EmoNet
  valence and arousal, SMIRK expression error and MediaPipe blendshapes, each a
  distance to the driving image. Pose: angle between the head rotations of the
  result and of the driving image (degrees). Lighting: correlation of the low-pass
  luminance, difference in mean luminance (0--255) and in mean color ($a^*b^*$)
  inside the face region. Realism: FaceScore and DSL-FIQA, without a reference.
  Rows of ours: resolution and inference mask; the face mask covers the face, ears and neck, and the face+hair
  mask also the hair, which the model then generates; lighting off: the first
  pass only.
  Italic values exceed a real photo of the target and are not ranked. Bold: best
  among the rows at their best setting.}
  \label{tab:suppl:allmetrics}
  \begin{tabular}{@{}l c c ccc ccc cc ccc cc@{}}
    \toprule
    & & & \multicolumn{3}{c}{Identity$\uparrow$} & \multicolumn{3}{c}{Expression$\downarrow$} & \multicolumn{2}{c}{Pose$\downarrow$} & \multicolumn{3}{c}{Lighting} & \multicolumn{2}{c}{Realism$\uparrow$} \\
    \cmidrule(lr){4-6} \cmidrule(lr){7-9} \cmidrule(lr){10-11} \cmidrule(lr){12-14} \cmidrule(l){15-16}
    Method & $M$ & $n$ & AdaFace & ArcFace & FaceNet & EmoNet & SMIRK & Blendsh. & HopeNet & 6DRepNet & Corr.$\uparrow$ & $\Delta L$$\downarrow$ & $\Delta ab$$\downarrow$ & FaceScore & DSL-FIQA \\
    \midrule
    SimSwap & 24 & 240 & 0.399 & 0.500 & 0.573 & 0.243 & 0.422 & \textbf{0.037} & \textbf{2.03} & 2.04 & \textbf{0.96} & \textbf{2.4} & 2.03 & 3.67 & 0.556 \\
    \quad $M{=}1$ & 1 & 240 & 0.326 & 0.414 & 0.506 & 0.241 & 0.424 & 0.036 & 2.08 & 2.04 & 0.96 & 2.3 & 2.01 & 3.64 & 0.555 \\
    FaceDancer & 24 & 240 & 0.314 & 0.394 & 0.618 & \textbf{0.205} & 0.420 & \textbf{0.037} & 2.38 & \textbf{1.99} & \textbf{0.96} & 2.5 & 1.21 & 4.39 & 0.711 \\
    \quad $M{=}1$ & 1 & 240 & 0.260 & 0.335 & 0.544 & 0.208 & 0.421 & 0.037 & 2.31 & 1.97 & 0.96 & 2.3 & 1.15 & 4.39 & 0.711 \\
    BlendFace & 24 & 240 & 0.425 & 0.487 & 0.537 & 0.231 & 0.466 & 0.041 & 4.42 & 3.97 & 0.90 & 6.3 & 1.39 & 3.73 & 0.640 \\
    \quad $M{=}1$ & 1 & 240 & 0.353 & 0.416 & 0.479 & 0.222 & 0.463 & 0.041 & 4.39 & 3.86 & 0.90 & 6.2 & 1.39 & 3.70 & 0.640 \\
    GHOST & 24 & 240 & 0.473 & 0.548 & 0.615 & 0.337 & 0.569 & 0.057 & 4.60 & 4.43 & 0.93 & 4.4 & \textbf{0.82} & 4.02 & 0.570 \\
    \quad $M{=}1$ & 1 & 240 & 0.401 & 0.475 & 0.561 & 0.341 & 0.576 & 0.057 & 4.45 & 4.43 & 0.93 & 4.3 & 0.82 & 4.00 & 0.569 \\
    InSwapper & 24 & 240 & \textit{0.690} & \textit{0.715} & \textbf{0.706} & 0.258 & 0.498 & 0.045 & 2.70 & 2.61 & 0.93 & 2.6 & 1.28 & 3.92 & 0.492 \\
    \quad $M{=}1$ & 1 & 240 & 0.566 & 0.606 & 0.637 & 0.275 & 0.495 & 0.045 & 2.62 & 2.60 & 0.93 & 2.6 & 1.27 & 3.92 & 0.492 \\
    HifiFace & 24 & 240 & 0.444 & 0.528 & 0.588 & 0.345 & 0.696 & 0.057 & 3.14 & 3.04 & 0.90 & 6.2 & 2.11 & 4.02 & 0.682 \\
    \quad $M{=}1$ & 1 & 240 & 0.354 & 0.437 & 0.523 & 0.369 & 0.717 & 0.061 & 3.42 & 3.42 & 0.89 & 6.1 & 2.12 & 3.96 & 0.680 \\
    Face-Adapter & 24 & 240 & 0.446 & 0.501 & 0.666 & 0.332 & 0.624 & 0.052 & 4.38 & 4.33 & 0.83 & 6.9 & 2.59 & 4.09 & 0.811 \\
    \quad $M{=}1$ & 1 & 240 & 0.355 & 0.408 & 0.580 & 0.343 & 0.618 & 0.054 & 4.34 & 3.98 & 0.83 & 7.1 & 2.64 & 4.10 & 0.817 \\
    DreamID-V & 1 & 240 & 0.516 & 0.550 & 0.652 & 0.372 & 0.689 & 0.059 & 7.52 & 8.18 & 0.52 & 15.2 & 5.88 & 4.42 & 0.732 \\
    \midrule
    DiffSwap & 1 & 226 & 0.166 & 0.208 & 0.255 & 0.213 & \textbf{0.416} & \textbf{0.037} & 2.86 & 2.46 & 0.94 & 6.0 & 2.65 & 4.42 & 0.654 \\
    REFace & 1 & 223 & 0.279 & 0.346 & 0.507 & 0.338 & 0.596 & 0.059 & 3.83 & 3.52 & 0.84 & 15.4 & 4.27 & 4.28 & 0.795 \\
    DiffFace & 1 & 223 & 0.315 & 0.381 & 0.527 & 0.325 & 0.531 & 0.052 & 4.07 & 3.79 & 0.93 & 2.7 & 3.43 & 4.72 & 0.634 \\
    E4S & 1 & 223 & 0.171 & 0.218 & 0.385 & 0.376 & 0.631 & 0.066 & 4.79 & 4.63 & 0.81 & 13.5 & 3.00 & 4.70 & 0.790 \\
    GPT-Image & 8 & 224 & 0.369 & 0.414 & 0.567 & 0.295 & 0.557 & 0.049 & 3.94 & 3.75 & 0.86 & 14.9 & 6.15 & \textbf{5.14} & \textbf{0.846} \\
    Gemini-Pro & 1 & 240 & 0.329 & 0.371 & 0.510 & 0.379 & 0.603 & 0.058 & 7.35 & 7.39 & 0.75 & 17.3 & 5.31 & 4.98 & 0.844 \\
    \midrule
    \textbf{Ours}, 256\,px, face & 24 & 240 & 0.573 & 0.601 & 0.647 & 0.253 & 0.550 & 0.042 & 3.99 & 3.51 & 0.83 & 23.9 & 4.23 & 4.62 & 0.668 \\
    \quad 512\,px, face & 24 & 240 & 0.586 & 0.609 & 0.641 & 0.239 & 0.548 & 0.041 & 4.09 & 3.52 & 0.84 & 18.4 & 3.70 & 4.50 & 0.750 \\
    \quad 256\,px, face+hair & 24 & 240 & \textbf{0.618} & \textbf{0.638} & \textbf{0.706} & 0.257 & 0.558 & 0.043 & 4.30 & 3.41 & 0.75 & 29.2 & 5.40 & 4.75 & 0.654 \\
    \quad 256\,px, face, lighting off & 24 & 240 & 0.590 & 0.616 & 0.672 & 0.256 & 0.577 & 0.044 & 3.97 & 3.76 & 0.66 & 25.4 & 7.39 & 4.62 & 0.675 \\
    \midrule
    Real photo of the target & -- & 240 & 0.656 & 0.677 & 0.736 & & & & & & & & & & \\
    Photo of another person & -- & 240 & 0.012 & 0.011 & 0.012 & & & & & & & & & & \\
    Driving image & -- & 240 & & & & & & & & & & & 4.23 & 0.772 \\
    \bottomrule
  \end{tabular}
\end{table*}

\begin{table*}[t]
  \centering\footnotesize
  \setlength{\tabcolsep}{3.2pt}
  \caption{\textbf{Identity on the identity benchmark, per condition.} Mean
  cosine similarity between the result and the $24$ held-out photos of the target.
  A: same sex and ethnicity, age gap of at most $15$ years and skin-brightness
  gap of at most $30$ (0--255). B: as A, but different sex. C: as A, but
  different ethnicity and a skin-brightness gap of at least $30$. D: same sex and
  ethnicity, age gap of at least $30$ years. Sex, age, ethnicity and skin
  brightness are estimated by models. $60$ pairs per condition. $M$, $n$ and the
  rows of ours as in Table~\ref{tab:suppl:allmetrics}. Italic values exceed a real photo of the
  target and are not ranked. Bold: best.}
  \label{tab:suppl:idcond}
  \begin{tabular}{@{}l c c ccccc ccccc ccccc@{}}
    \toprule
    & & & \multicolumn{5}{c}{AdaFace$\uparrow$} & \multicolumn{5}{c}{ArcFace$\uparrow$} & \multicolumn{5}{c}{FaceNet$\uparrow$} \\
    \cmidrule(lr){4-8} \cmidrule(lr){9-13} \cmidrule(l){14-18}
    Method & $M$ & $n$ & A & B & C & D & All & A & B & C & D & All & A & B & C & D & All \\
    \midrule
    SimSwap & 24 & 240 & 0.441 & 0.373 & 0.384 & 0.399 & 0.399 & 0.535 & 0.480 & 0.491 & 0.493 & 0.500 & 0.621 & 0.548 & 0.571 & 0.551 & 0.573 \\
    FaceDancer & 24 & 240 & 0.335 & 0.297 & 0.312 & 0.311 & 0.314 & 0.401 & 0.387 & 0.395 & 0.391 & 0.394 & 0.630 & 0.614 & 0.625 & 0.605 & 0.618 \\
    BlendFace & 24 & 240 & 0.461 & 0.431 & 0.408 & 0.400 & 0.425 & 0.528 & 0.478 & 0.479 & 0.463 & 0.487 & 0.601 & 0.521 & 0.529 & 0.497 & 0.537 \\
    GHOST & 24 & 240 & 0.493 & 0.468 & 0.473 & 0.457 & 0.473 & 0.571 & 0.542 & 0.550 & 0.530 & 0.548 & 0.654 & 0.614 & 0.611 & 0.580 & 0.615 \\
    InSwapper & 24 & 240 & \textit{0.691} & \textit{0.684} & \textit{0.696} & \textit{0.688} & \textit{0.690} & \textit{0.718} & \textit{0.709} & \textit{0.722} & \textit{0.712} & \textit{0.715} & \textit{0.727} & 0.693 & 0.712 & 0.694 & \textbf{0.706} \\
    HifiFace & 24 & 240 & 0.456 & 0.445 & 0.444 & 0.429 & 0.444 & 0.539 & 0.525 & 0.536 & 0.513 & 0.528 & 0.623 & 0.578 & 0.600 & 0.550 & 0.588 \\
    Face-Adapter & 24 & 240 & 0.440 & 0.434 & 0.446 & 0.464 & 0.446 & 0.491 & 0.491 & 0.506 & 0.517 & 0.501 & 0.666 & 0.653 & 0.678 & 0.666 & 0.666 \\
    DreamID-V & 1 & 240 & 0.511 & 0.508 & 0.516 & 0.529 & 0.516 & 0.550 & 0.543 & 0.557 & 0.552 & 0.550 & 0.647 & 0.641 & 0.672 & 0.649 & 0.652 \\
    \midrule
    DiffSwap & 1 & 226 & 0.202 & 0.154 & 0.167 & 0.144 & 0.166 & 0.237 & 0.205 & 0.202 & 0.191 & 0.208 & 0.329 & 0.240 & 0.249 & 0.205 & 0.255 \\
    REFace & 1 & 223 & 0.319 & 0.256 & 0.286 & 0.257 & 0.279 & 0.383 & 0.324 & 0.362 & 0.316 & 0.346 & 0.562 & 0.451 & 0.550 & 0.465 & 0.507 \\
    DiffFace & 1 & 223 & 0.341 & 0.315 & 0.304 & 0.299 & 0.315 & 0.408 & 0.379 & 0.372 & 0.365 & 0.381 & 0.546 & 0.527 & 0.548 & 0.486 & 0.527 \\
    E4S & 1 & 223 & 0.193 & 0.159 & 0.170 & 0.164 & 0.171 & 0.234 & 0.205 & 0.222 & 0.211 & 0.218 & 0.418 & 0.367 & 0.376 & 0.379 & 0.385 \\
    GPT-Image & 8 & 224 & 0.347 & 0.359 & 0.380 & 0.390 & 0.369 & 0.382 & 0.412 & 0.423 & 0.435 & 0.414 & 0.554 & 0.568 & 0.574 & 0.571 & 0.567 \\
    Gemini-Pro & 1 & 240 & 0.345 & 0.320 & 0.334 & 0.318 & 0.329 & 0.376 & 0.365 & 0.382 & 0.363 & 0.371 & 0.527 & 0.516 & 0.509 & 0.490 & 0.510 \\
    \midrule
    \textbf{Ours}, 256\,px, face & 24 & 240 & 0.581 & 0.556 & 0.578 & 0.577 & 0.573 & 0.609 & 0.587 & 0.609 & 0.601 & 0.601 & 0.653 & 0.631 & 0.658 & 0.646 & 0.647 \\
    \quad 512\,px, face & 24 & 240 & 0.592 & 0.577 & 0.587 & 0.589 & 0.586 & 0.616 & 0.601 & 0.612 & 0.607 & 0.609 & 0.650 & 0.628 & 0.656 & 0.632 & 0.641 \\
    \quad 256\,px, face+hair & 24 & 240 & \textbf{0.607} & \textbf{0.618} & \textbf{0.621} & \textbf{0.623} & \textbf{0.618} & \textbf{0.632} & \textbf{0.640} & \textbf{0.645} & \textbf{0.636} & \textbf{0.638} & \textbf{0.690} & \textbf{0.707} & \textbf{0.721} & \textbf{0.706} & \textbf{0.706} \\
    \quad 256\,px, face, lighting off & 24 & 240 & 0.592 & 0.575 & 0.595 & 0.598 & 0.590 & 0.622 & 0.602 & 0.623 & 0.618 & 0.616 & 0.672 & 0.650 & 0.690 & 0.675 & 0.672 \\
    \midrule
    Real photo of the target & -- & 240 & 0.642 & 0.650 & 0.663 & 0.668 & 0.656 & 0.671 & 0.669 & 0.687 & 0.682 & 0.677 & 0.724 & 0.731 & 0.750 & 0.739 & 0.736 \\
    \bottomrule
  \end{tabular}
\end{table*}

\begin{table*}[t]
  \centering\footnotesize
  \setlength{\tabcolsep}{4pt}
  \caption{\textbf{Metrics on the celebrity targets of the user study} ($176$
  pairs). Most celebrities have few photos besides the ones given to the methods,
  so ArcFace is computed twice: against the given photos (Pool), which favors
  methods that copy them, and against held-out photos (Held-out), only for the
  $40$ celebrities with at least four. Other columns as in
  Table~\ref{tab:suppl:allmetrics}; $n$ counts the results scored by ArcFace (Pool).
  Bold: best.}
  \label{tab:suppl:celeb}
  \begin{tabular}{@{}l c c cc cc c c cc@{}}
    \toprule
    & & & \multicolumn{2}{c}{ArcFace$\uparrow$} & \multicolumn{2}{c}{Expression$\downarrow$} & Pose$\downarrow$ & Lighting$\uparrow$ & \multicolumn{2}{c}{Realism$\uparrow$} \\
    \cmidrule(lr){4-5} \cmidrule(lr){6-7} \cmidrule(l){10-11}
    Method & $M$ & $n$ & Pool & Held-out & EmoNet & SMIRK & HopeNet & Corr. & FaceScore & DSL-FIQA \\
    \midrule
    SimSwap & 24 & 176 & 0.513 & 0.501 & 0.277 & 0.421 & \textbf{1.70} & \textbf{0.96} & 4.34 & 0.533 \\
    FaceDancer & 24 & 176 & 0.404 & 0.403 & 0.240 & 0.429 & 1.81 & 0.95 & 5.12 & 0.696 \\
    BlendFace & 24 & 176 & 0.550 & 0.544 & \textbf{0.226} & \textbf{0.362} & 1.86 & 0.92 & 4.56 & 0.624 \\
    GHOST & 24 & 176 & 0.562 & 0.572 & 0.378 & 0.551 & 2.27 & 0.92 & 4.62 & 0.545 \\
    InSwapper & 24 & 176 & \textbf{0.733} & \textbf{0.724} & 0.274 & 0.491 & 2.27 & 0.91 & 4.57 & 0.469 \\
    HifiFace & 24 & 176 & 0.521 & 0.521 & 0.371 & 0.700 & 2.61 & 0.89 & 4.33 & 0.671 \\
    Face-Adapter & 24 & 176 & 0.495 & 0.466 & 0.402 & 0.635 & 3.72 & 0.81 & 4.66 & 0.816 \\
    DreamID-V & 1 & 176 & 0.599 & 0.588 & 0.409 & 0.667 & 5.50 & 0.47 & 4.69 & 0.655 \\
    \midrule
    DiffSwap & 1 & 173 & 0.217 & 0.227 & 0.259 & 0.418 & 2.09 & 0.93 & 4.86 & 0.634 \\
    REFace & 1 & 173 & 0.334 & 0.328 & 0.405 & 0.625 & 3.22 & 0.82 & 4.77 & 0.783 \\
    DiffFace & 1 & 173 & 0.360 & 0.341 & 0.432 & 0.589 & 3.31 & 0.92 & 5.09 & 0.609 \\
    E4S & 1 & 173 & 0.269 & 0.257 & 0.437 & 0.665 & 4.19 & 0.71 & 4.95 & 0.785 \\
    GPT-Image & 8 & 163 & 0.461 & 0.447 & 0.347 & 0.594 & 4.23 & 0.82 & \textbf{5.48} & \textbf{0.831} \\
    Gemini-Pro & 1 & 172 & 0.479 & 0.461 & 0.408 & 0.643 & 5.43 & 0.70 & 5.30 & 0.827 \\
    \midrule
    \textbf{Ours} & 24 & 176 & 0.619 & 0.603 & 0.258 & 0.568 & 3.18 & 0.83 & 5.11 & 0.703 \\
    \midrule
    Driving image & -- & 176 & & & & & & & 5.15 & 0.775 \\
    \bottomrule
  \end{tabular}
\end{table*}

\subsection{Qualitative comparison}
\label{sec:suppl:qual}
\begin{figure*}[p]
  \centering
  \edef\qcell{\the\dimexpr 0.0859\textwidth\relax}\edef\qgap{\the\dimexpr 0.0020\textwidth\relax}%
  \edef\qlab{\the\dimexpr 0.030\textwidth\relax}\edef\qrow{\the\dimexpr 0.9650\textwidth\relax}%
  \makebox[\qlab]{}\hspace{\qgap}\makebox[\qrow][l]{\makebox[\qcell][c]{\scriptsize Target}\hspace{\qgap}\makebox[\qcell][c]{\scriptsize Driving}\hspace{\qgap}\makebox[\qcell][c]{\scriptsize SimSwap}\hspace{\qgap}\makebox[\qcell][c]{\scriptsize FaceDancer}\hspace{\qgap}\makebox[\qcell][c]{\scriptsize BlendFace}\hspace{\qgap}\makebox[\qcell][c]{\scriptsize GHOST}\hspace{\qgap}\makebox[\qcell][c]{\scriptsize InSwapper}\hspace{\qgap}\makebox[\qcell][c]{\scriptsize HifiFace}\hspace{\qgap}\makebox[\qcell][c]{\scriptsize Face-Adapter}\hspace{\qgap}\makebox[\qcell][c]{\scriptsize DreamID-V}\hspace{\qgap}\makebox[\qcell][c]{\scriptsize \textbf{Ours}}}\\[2pt]
  \begin{minipage}[c]{\qlab}\centering\small\textbf{A}\end{minipage}\hspace{\qgap}\begin{minipage}[c]{\qrow}\includegraphics[width=\qrow]{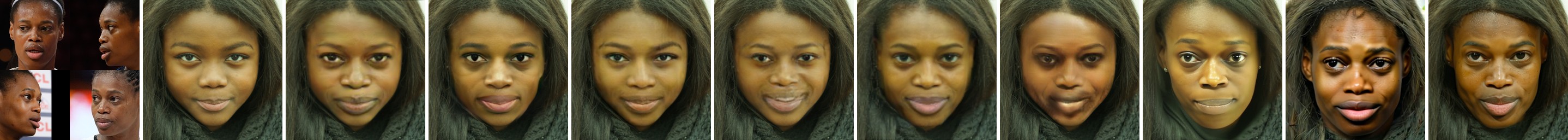}\\[\qgap]\includegraphics[width=\qrow]{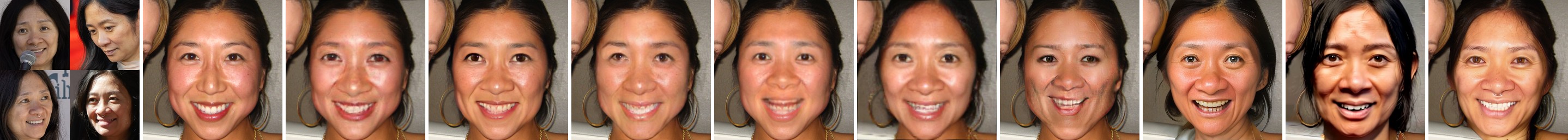}\\[\qgap]\includegraphics[width=\qrow]{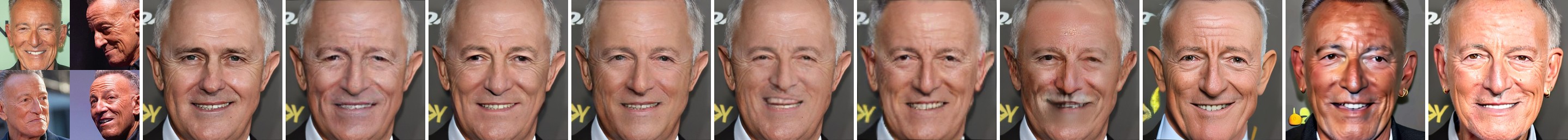}\end{minipage}\\[4pt]
  \begin{minipage}[c]{\qlab}\centering\small\textbf{B}\end{minipage}\hspace{\qgap}\begin{minipage}[c]{\qrow}\includegraphics[width=\qrow]{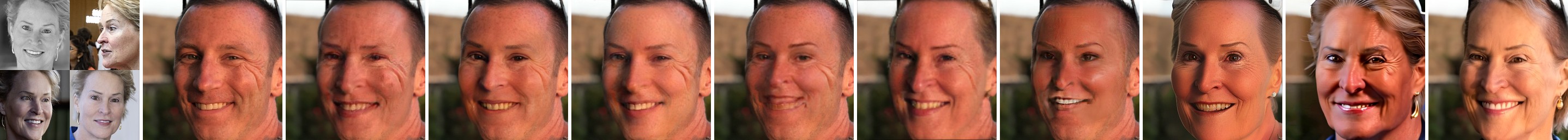}\\[\qgap]\includegraphics[width=\qrow]{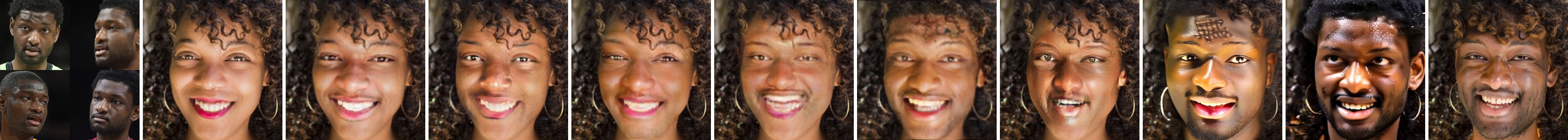}\\[\qgap]\includegraphics[width=\qrow]{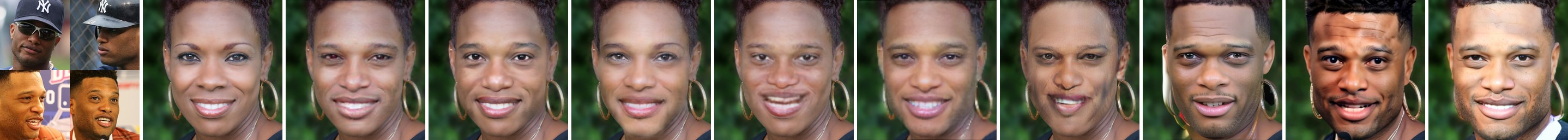}\end{minipage}\\[4pt]
  \begin{minipage}[c]{\qlab}\centering\small\textbf{C}\end{minipage}\hspace{\qgap}\begin{minipage}[c]{\qrow}\includegraphics[width=\qrow]{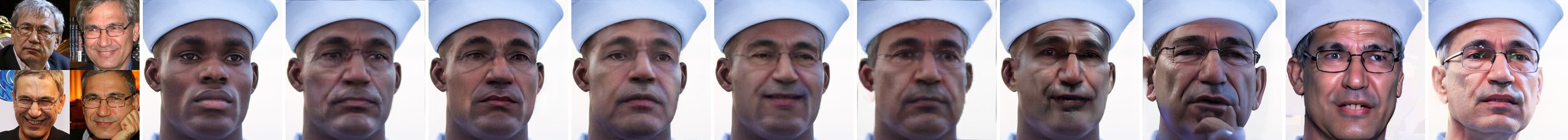}\\[\qgap]\includegraphics[width=\qrow]{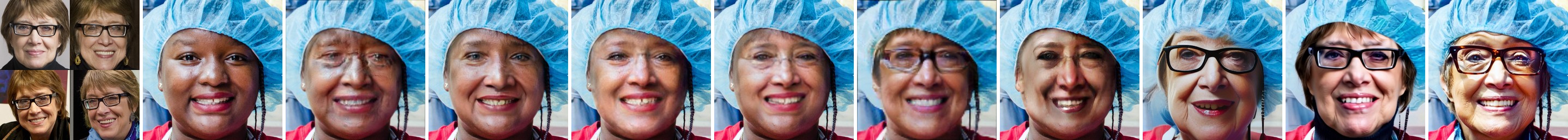}\\[\qgap]\includegraphics[width=\qrow]{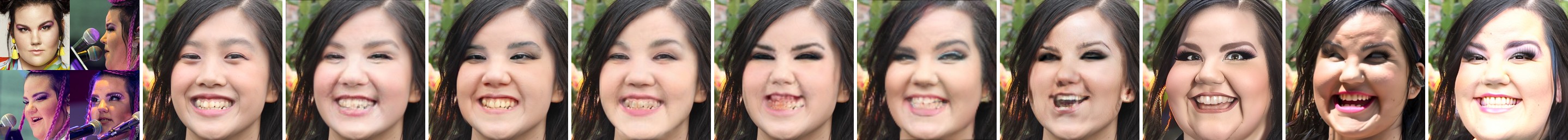}\end{minipage}\\[4pt]
  \begin{minipage}[c]{\qlab}\centering\small\textbf{D}\end{minipage}\hspace{\qgap}\begin{minipage}[c]{\qrow}\includegraphics[width=\qrow]{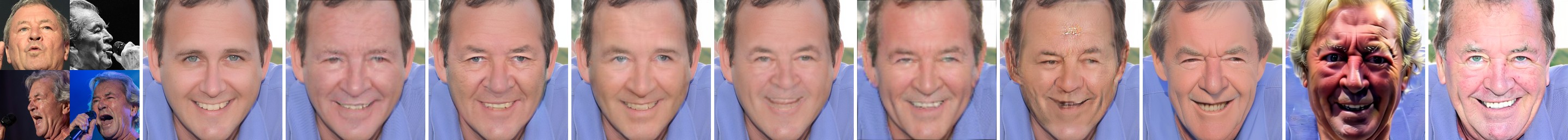}\\[\qgap]\includegraphics[width=\qrow]{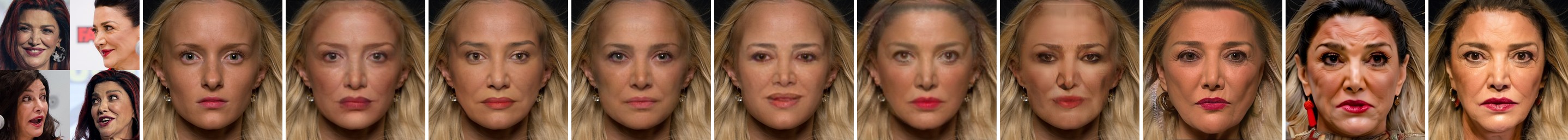}\\[\qgap]\includegraphics[width=\qrow]{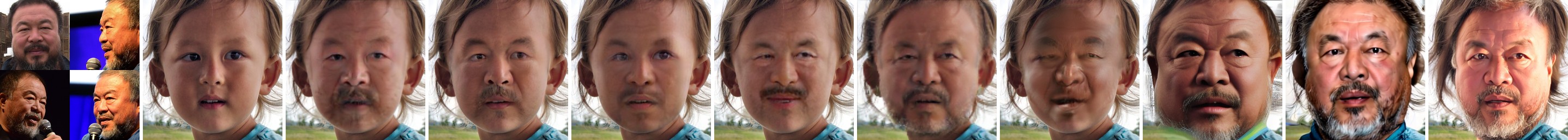}\end{minipage}\\[0pt]
  \caption{\textbf{Comparison with the eight methods of Table~\ref{tab:main}}, on $12$ pairs of the user study, three per condition of Table~\ref{tab:suppl:idcond} (A: matched; B: different sex; C: different ethnicity and skin brightness; D: age gap of at least $30$ years). Target: four of the photos given to the methods. Each method runs at its best setting. We picked these pairs among the $416$ for a range of targets.}
  \label{fig:suppl:qual8}
\end{figure*}

\begin{figure*}[p]
  \centering
  \edef\qcell{\the\dimexpr 0.0859\textwidth\relax}\edef\qgap{\the\dimexpr 0.0020\textwidth\relax}%
  \edef\qlab{\the\dimexpr 0.030\textwidth\relax}\edef\qrow{\the\dimexpr 0.7892\textwidth\relax}%
  \makebox[\qlab]{}\hspace{\qgap}\makebox[\qrow][l]{\makebox[\qcell][c]{\scriptsize Target}\hspace{\qgap}\makebox[\qcell][c]{\scriptsize Driving}\hspace{\qgap}\makebox[\qcell][c]{\scriptsize DiffSwap}\hspace{\qgap}\makebox[\qcell][c]{\scriptsize REFace}\hspace{\qgap}\makebox[\qcell][c]{\scriptsize DiffFace}\hspace{\qgap}\makebox[\qcell][c]{\scriptsize E4S}\hspace{\qgap}\makebox[\qcell][c]{\scriptsize GPT-Image}\hspace{\qgap}\makebox[\qcell][c]{\scriptsize Gemini-Pro}\hspace{\qgap}\makebox[\qcell][c]{\scriptsize \textbf{Ours}}}\\[2pt]
  \begin{minipage}[c]{\qlab}\centering\small\textbf{A}\end{minipage}\hspace{\qgap}\begin{minipage}[c]{\qrow}\includegraphics[width=\qrow]{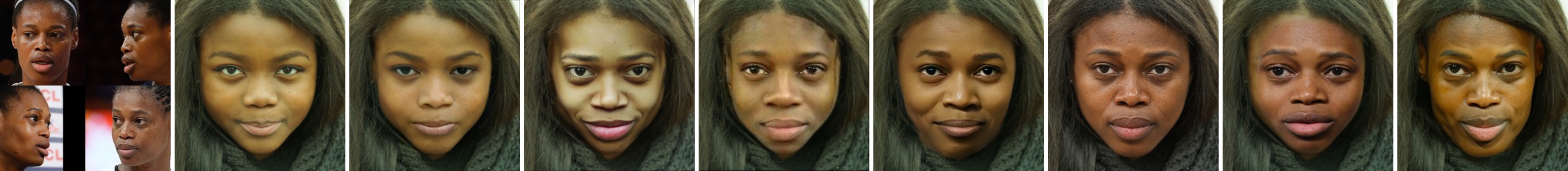}\\[\qgap]\includegraphics[width=\qrow]{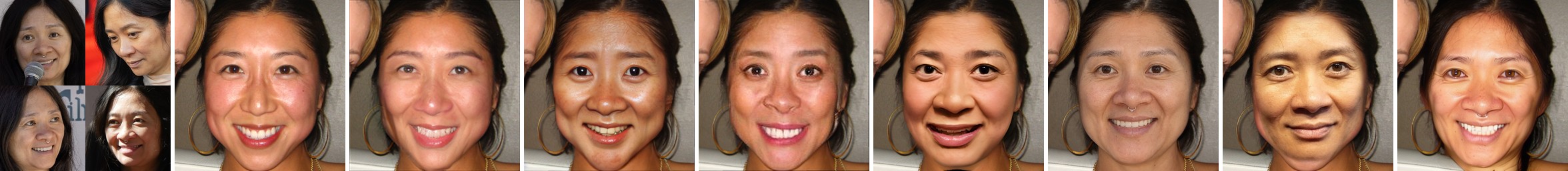}\\[\qgap]\includegraphics[width=\qrow]{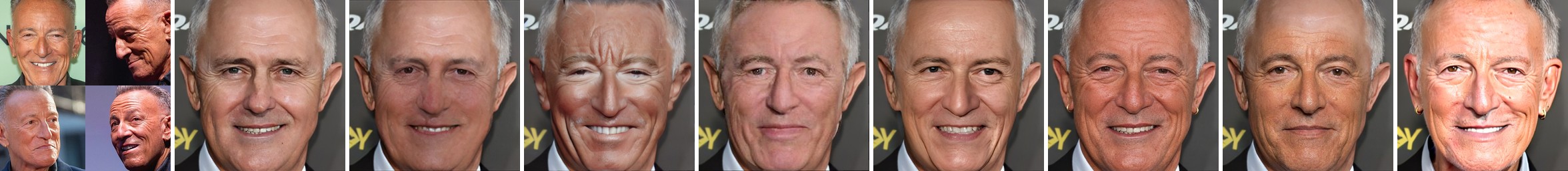}\end{minipage}\\[4pt]
  \begin{minipage}[c]{\qlab}\centering\small\textbf{B}\end{minipage}\hspace{\qgap}\begin{minipage}[c]{\qrow}\includegraphics[width=\qrow]{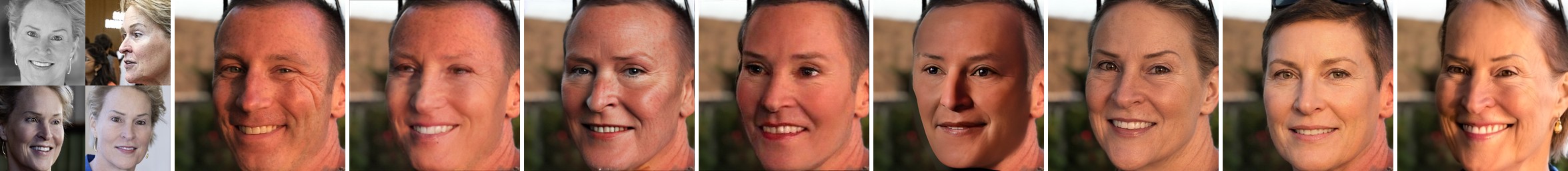}\\[\qgap]\includegraphics[width=\qrow]{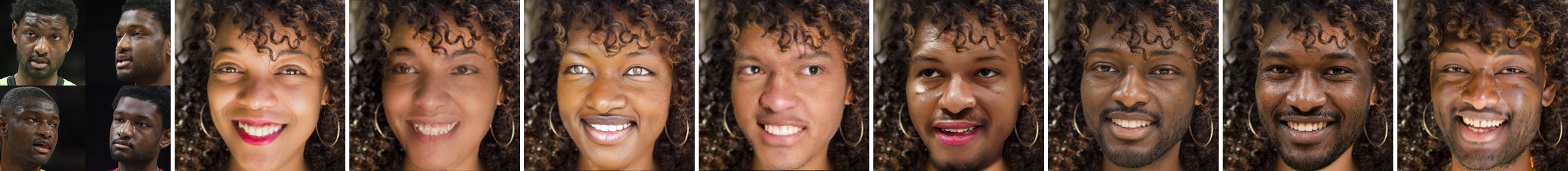}\\[\qgap]\includegraphics[width=\qrow]{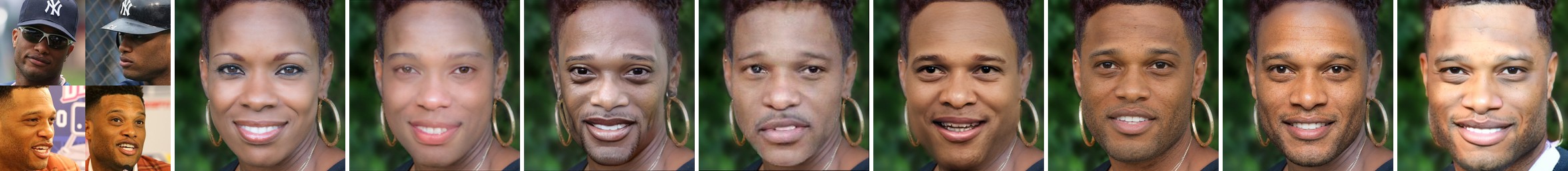}\end{minipage}\\[4pt]
  \begin{minipage}[c]{\qlab}\centering\small\textbf{C}\end{minipage}\hspace{\qgap}\begin{minipage}[c]{\qrow}\includegraphics[width=\qrow]{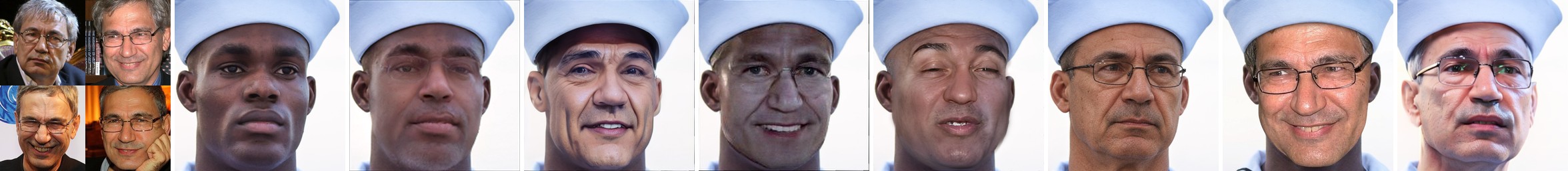}\\[\qgap]\includegraphics[width=\qrow]{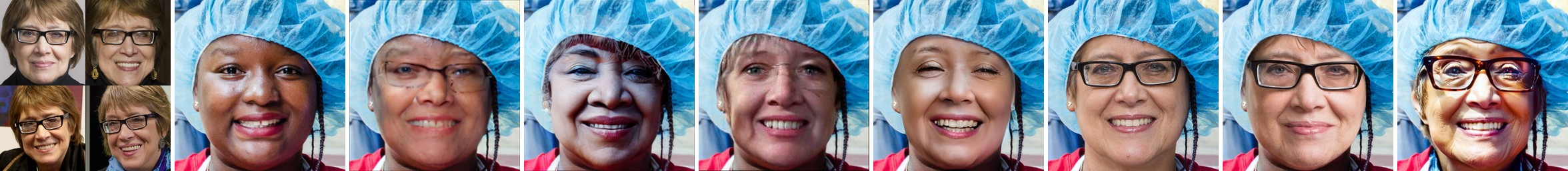}\\[\qgap]\includegraphics[width=\qrow]{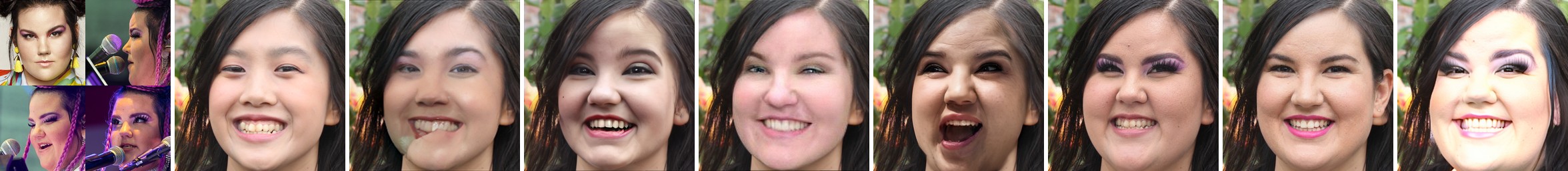}\end{minipage}\\[4pt]
  \begin{minipage}[c]{\qlab}\centering\small\textbf{D}\end{minipage}\hspace{\qgap}\begin{minipage}[c]{\qrow}\includegraphics[width=\qrow]{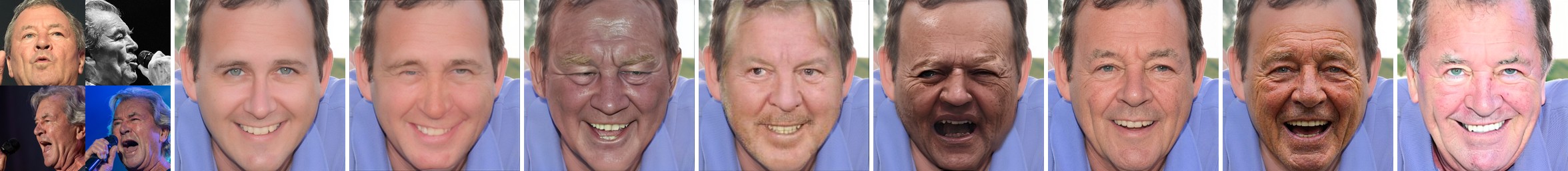}\\[\qgap]\includegraphics[width=\qrow]{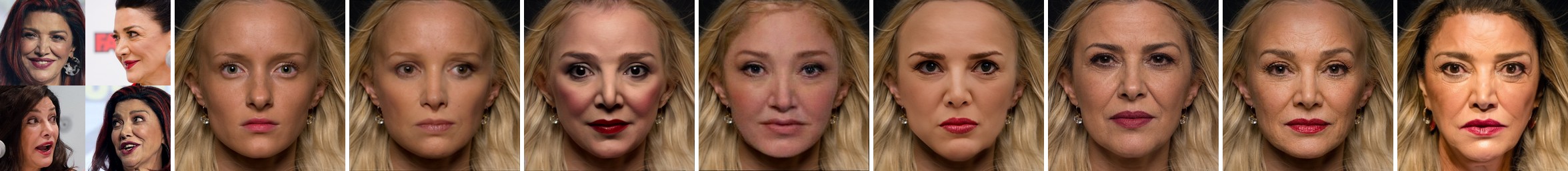}\\[\qgap]\includegraphics[width=\qrow]{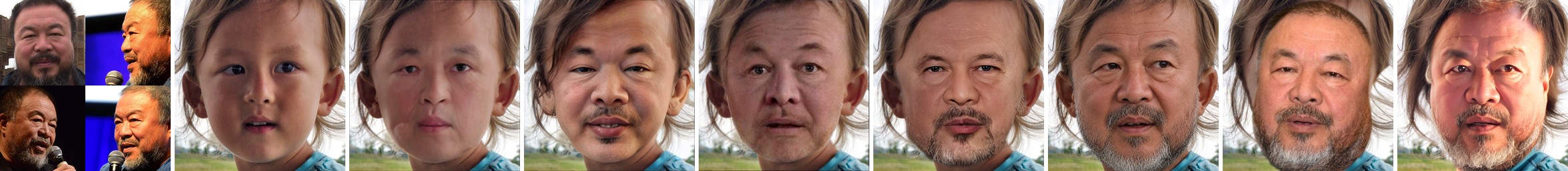}\end{minipage}\\[0pt]
  \caption{\textbf{Comparison with the other methods and the two image models}, on the same pairs as Fig.~\ref{fig:suppl:qual8}.}
  \label{fig:suppl:qual6}
\end{figure*}

Figs.~\ref{fig:suppl:qual8} and~\ref{fig:suppl:qual6} show the results of all fourteen methods on $12$ pairs of the user study, three per condition.

\section{Expression, lighting and occluders}
\label{sec:suppl:driving}

\subsection{Expression and head pose}
\label{sec:suppl:driving:expr}
We take driving images from MEAD~\cite{wang2020mead}, in which actors show
eight emotions and are filmed by seven cameras at once. We use the four subjects
that both steps of Stage~1 held out, so their faces are new to the identity
encoder, which saw MEAD in $1\%$ of the samples of Stage~1; the geometry branch
was pretrained on all MEAD subjects (Sec.~\ref{sec:suppl:impl:arch}). For each subject, we take one front-camera frame per
emotion, neutral and seven emotions at the highest intensity, picked as the frame
that EmoNet~\cite{toisoul2021emonet} assigns to the emotion most clearly, and
one frame of a surprised expression from every camera. One subject was filmed
by three of the seven cameras only, which gives $52$ driving images. The identities of four
targets of the identity benchmark, none close to an identity of our training
data, are transferred into these frames by our model, InSwapper, Face-Adapter and
DreamID-V, with the settings of Table~\ref{tab:main}.

Table~\ref{tab:suppl:mead} compares each result with its driving image. Our results
take the EmoNet class of the driving image most often, for the emotions ($0.68$,
against $0.35$ to $0.54$) and for the camera views ($1.00$, against $0.86$ to
$0.89$), and have the lowest valence-arousal distance. InSwapper has a slightly
lower SMIRK expression error on the emotions and, as in Table~\ref{tab:main},
the highest ArcFace score. Under the side and top views, our ArcFace score stays
at $0.53$.

\begin{figure*}[t]
  \centering
  \noindent\makebox[\textwidth][l]{\makebox[0.085\textwidth]{}\makebox[0.0998\textwidth][c]{\scriptsize Target}\hspace{0.0055\textwidth}\makebox[0.0998\textwidth][c]{\scriptsize Neutral}\hspace{0.0016\textwidth}\makebox[0.0998\textwidth][c]{\scriptsize Happy}\hspace{0.0016\textwidth}\makebox[0.0998\textwidth][c]{\scriptsize Sad}\hspace{0.0016\textwidth}\makebox[0.0998\textwidth][c]{\scriptsize Surprised}\hspace{0.0016\textwidth}\makebox[0.0998\textwidth][c]{\scriptsize Fear}\hspace{0.0016\textwidth}\makebox[0.0998\textwidth][c]{\scriptsize Disgusted}\hspace{0.0016\textwidth}\makebox[0.0998\textwidth][c]{\scriptsize Angry}\hspace{0.0016\textwidth}\makebox[0.0998\textwidth][c]{\scriptsize Contempt}}\\[2pt]
  \noindent\makebox[0.085\textwidth][r]{\scriptsize MEAD\hspace{3pt}}\raisebox{-0.5\height}{\includegraphics[width=0.915\textwidth]{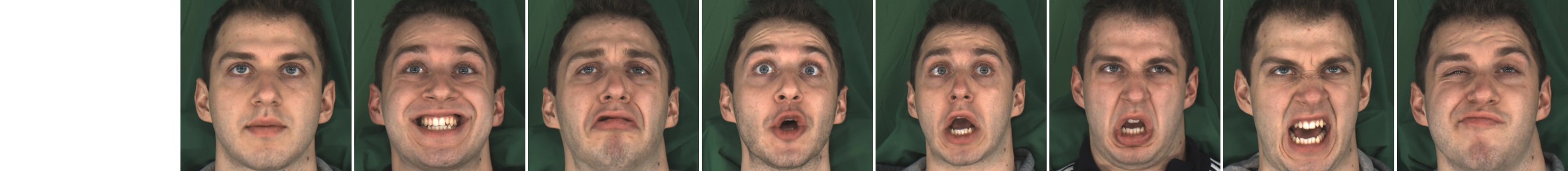}}\\[1pt]
  \noindent\makebox[0.085\textwidth][r]{\scriptsize \textbf{Ours}\hspace{3pt}}\raisebox{-0.5\height}{\includegraphics[width=0.915\textwidth]{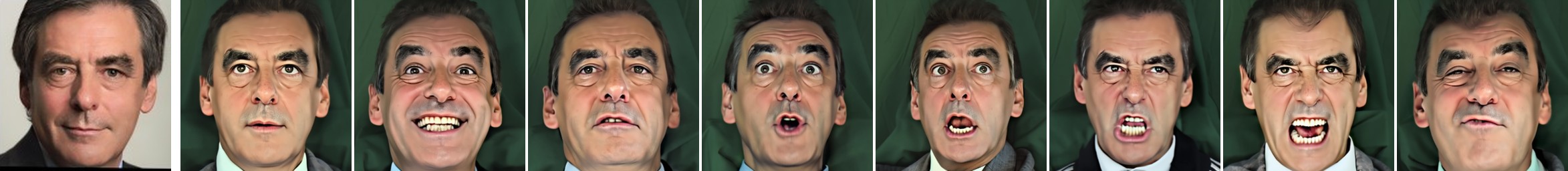}}\\[4pt]
  \noindent\makebox[0.085\textwidth][r]{\scriptsize MEAD\hspace{3pt}}\raisebox{-0.5\height}{\includegraphics[width=0.915\textwidth]{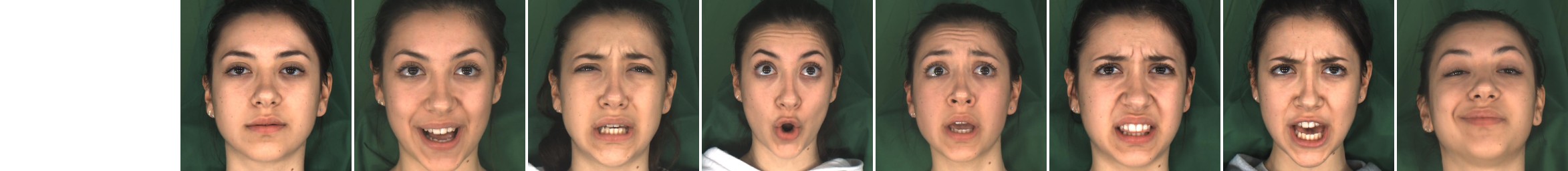}}\\[1pt]
  \noindent\makebox[0.085\textwidth][r]{\scriptsize \textbf{Ours}\hspace{3pt}}\raisebox{-0.5\height}{\includegraphics[width=0.915\textwidth]{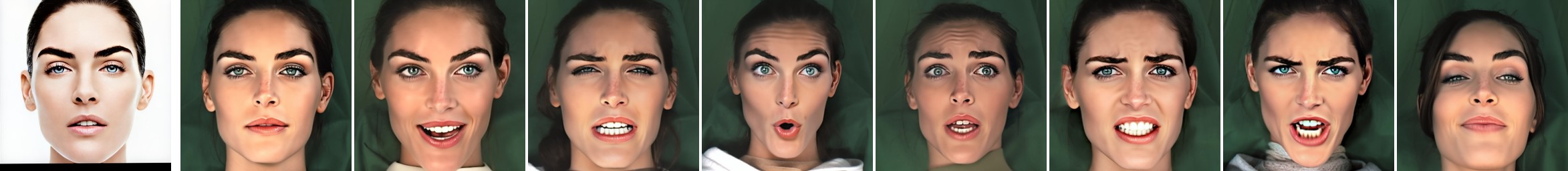}}\\[4pt]
  \noindent\makebox[0.085\textwidth][r]{\scriptsize MEAD\hspace{3pt}}\raisebox{-0.5\height}{\includegraphics[width=0.915\textwidth]{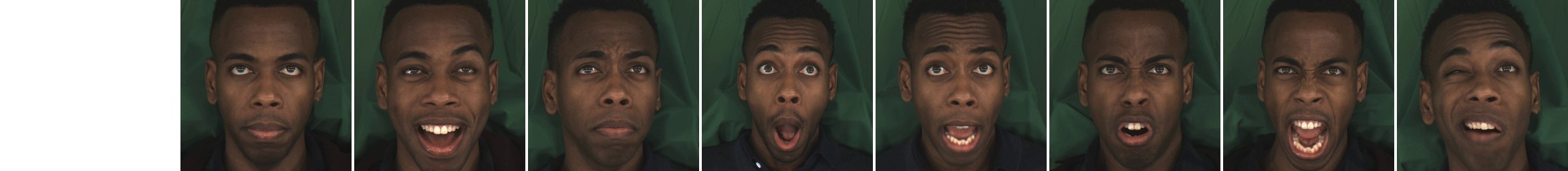}}\\[1pt]
  \noindent\makebox[0.085\textwidth][r]{\scriptsize \textbf{Ours}\hspace{3pt}}\raisebox{-0.5\height}{\includegraphics[width=0.915\textwidth]{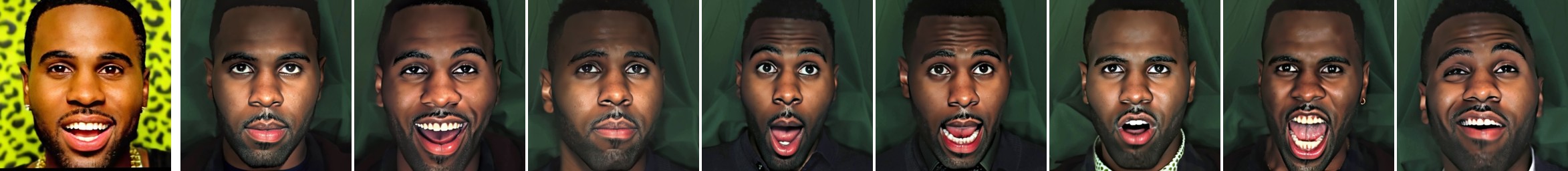}}\\[4pt]
  \noindent\makebox[0.085\textwidth][r]{\scriptsize MEAD\hspace{3pt}}\raisebox{-0.5\height}{\includegraphics[width=0.915\textwidth]{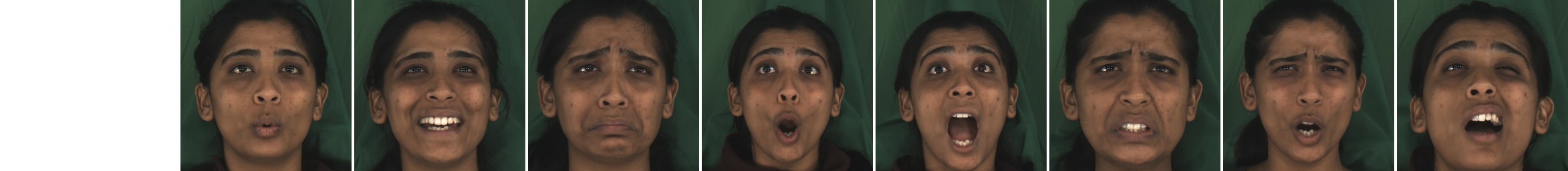}}\\[1pt]
  \noindent\makebox[0.085\textwidth][r]{\scriptsize \textbf{Ours}\hspace{3pt}}\raisebox{-0.5\height}{\includegraphics[width=0.915\textwidth]{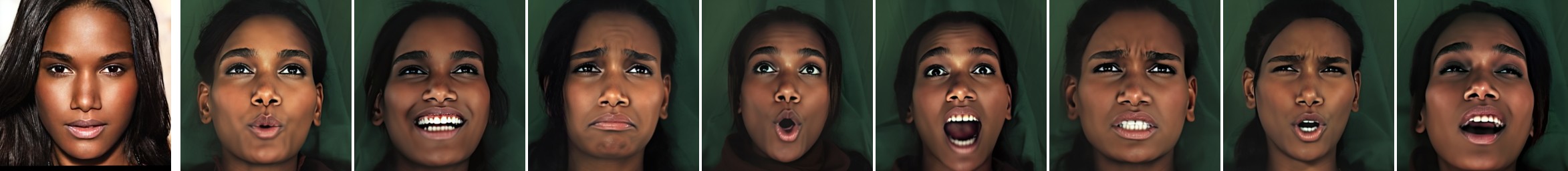}}
  \caption{\textbf{Our results follow the emotions of MEAD.} Four MEAD subjects (from top to bottom: a man and a woman with light skin, a man and a woman with dark skin), each paired with a target of the identity benchmark of the same sex and a similar skin tone. For each subject, the upper row shows the frames of the front camera, and the lower row our results; the first column is a photo of the target.}
  \label{fig:suppl:meadexpr}
\end{figure*}

\begin{figure*}[t]
  \centering
  \noindent\makebox[\textwidth][l]{\makebox[0.085\textwidth]{}\makebox[0.1123\textwidth][c]{\scriptsize Target}\hspace{0.0061\textwidth}\makebox[0.1123\textwidth][c]{\scriptsize Left $60^\circ$}\hspace{0.0018\textwidth}\makebox[0.1123\textwidth][c]{\scriptsize Left $30^\circ$}\hspace{0.0018\textwidth}\makebox[0.1123\textwidth][c]{\scriptsize Front}\hspace{0.0018\textwidth}\makebox[0.1123\textwidth][c]{\scriptsize Right $30^\circ$}\hspace{0.0018\textwidth}\makebox[0.1123\textwidth][c]{\scriptsize Right $60^\circ$}\hspace{0.0018\textwidth}\makebox[0.1123\textwidth][c]{\scriptsize Above}\hspace{0.0018\textwidth}\makebox[0.1123\textwidth][c]{\scriptsize Below}}\\[2pt]
  \noindent\makebox[0.085\textwidth][r]{\scriptsize MEAD\hspace{3pt}}\raisebox{-0.5\height}{\includegraphics[width=0.915\textwidth]{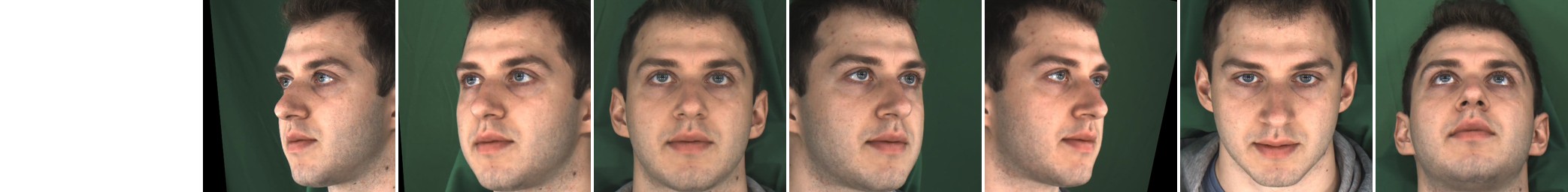}}\\[1pt]
  \noindent\makebox[0.085\textwidth][r]{\scriptsize \textbf{Ours}\hspace{3pt}}\raisebox{-0.5\height}{\includegraphics[width=0.915\textwidth]{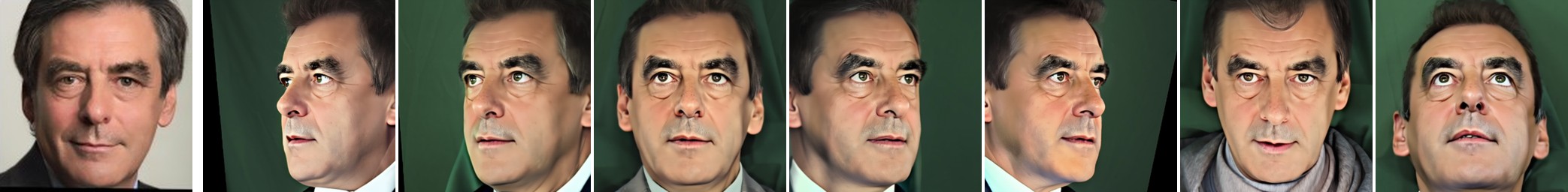}}\\[4pt]
  \noindent\makebox[0.085\textwidth][r]{\scriptsize MEAD\hspace{3pt}}\raisebox{-0.5\height}{\includegraphics[width=0.915\textwidth]{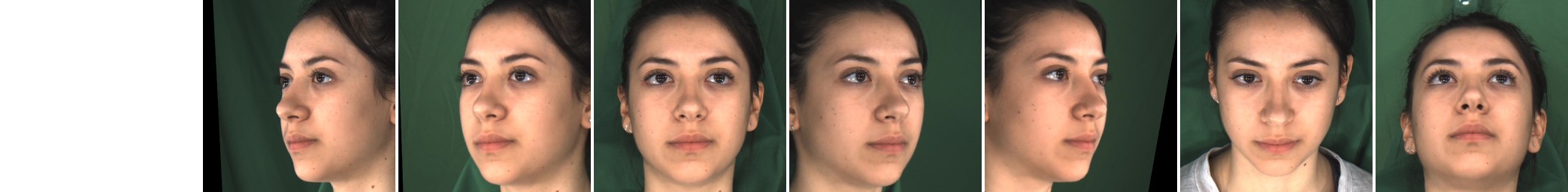}}\\[1pt]
  \noindent\makebox[0.085\textwidth][r]{\scriptsize \textbf{Ours}\hspace{3pt}}\raisebox{-0.5\height}{\includegraphics[width=0.915\textwidth]{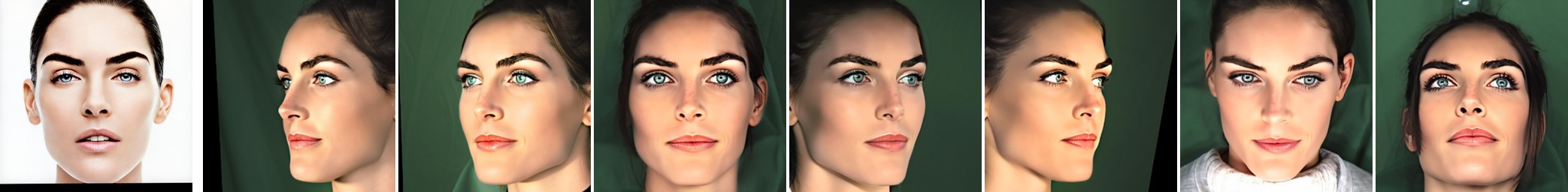}}\\[4pt]
  \noindent\makebox[0.085\textwidth][r]{\scriptsize MEAD\hspace{3pt}}\raisebox{-0.5\height}{\includegraphics[width=0.915\textwidth]{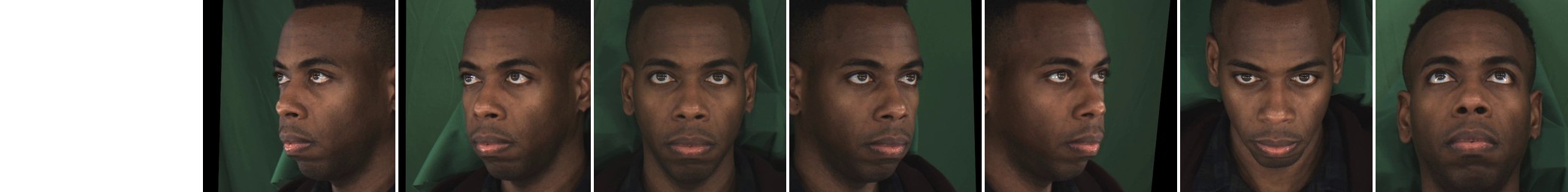}}\\[1pt]
  \noindent\makebox[0.085\textwidth][r]{\scriptsize \textbf{Ours}\hspace{3pt}}\raisebox{-0.5\height}{\includegraphics[width=0.915\textwidth]{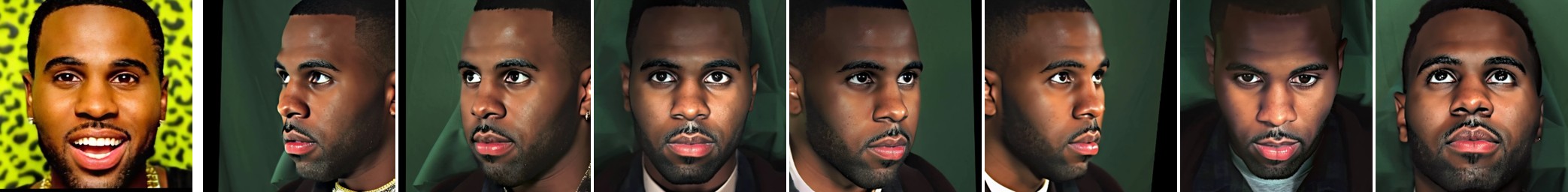}}\\[4pt]
  \noindent\makebox[0.085\textwidth][r]{\scriptsize MEAD\hspace{3pt}}\raisebox{-0.5\height}{\includegraphics[width=0.915\textwidth]{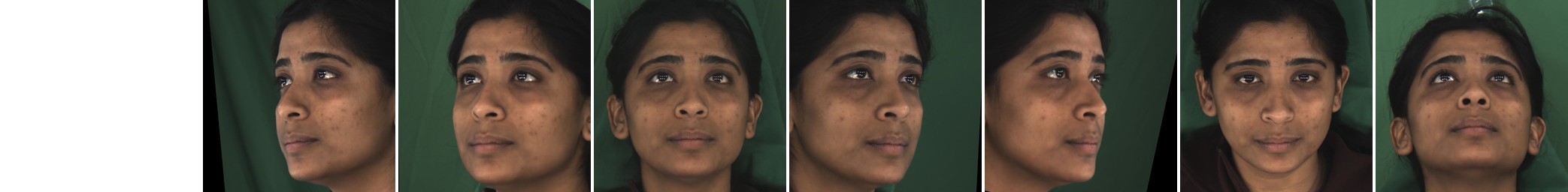}}\\[1pt]
  \noindent\makebox[0.085\textwidth][r]{\scriptsize \textbf{Ours}\hspace{3pt}}\raisebox{-0.5\height}{\includegraphics[width=0.915\textwidth]{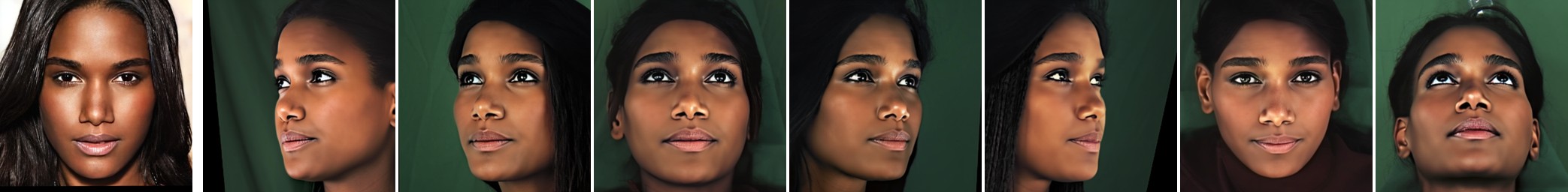}}
  \caption{\textbf{Our results under the seven cameras of MEAD.} The subjects and targets of Fig.~\ref{fig:suppl:meadexpr}, with a neutral face filmed by the seven cameras at the same moment.}
  \label{fig:suppl:meadpose}
\end{figure*}

\begin{figure*}[t]
  \centering
  \noindent\makebox[\textwidth][l]{\makebox[0.085\textwidth]{}\makebox[0.0998\textwidth][c]{\scriptsize Target}\hspace{0.0055\textwidth}\makebox[0.0998\textwidth][c]{\scriptsize Neutral}\hspace{0.0016\textwidth}\makebox[0.0998\textwidth][c]{\scriptsize Happy}\hspace{0.0016\textwidth}\makebox[0.0998\textwidth][c]{\scriptsize Sad}\hspace{0.0016\textwidth}\makebox[0.0998\textwidth][c]{\scriptsize Surprised}\hspace{0.0016\textwidth}\makebox[0.0998\textwidth][c]{\scriptsize Fear}\hspace{0.0016\textwidth}\makebox[0.0998\textwidth][c]{\scriptsize Disgusted}\hspace{0.0016\textwidth}\makebox[0.0998\textwidth][c]{\scriptsize Angry}\hspace{0.0016\textwidth}\makebox[0.0998\textwidth][c]{\scriptsize Contempt}}\\[2pt]
  \noindent\makebox[0.085\textwidth][r]{\scriptsize MEAD\hspace{3pt}}\raisebox{-0.5\height}{\includegraphics[width=0.915\textwidth]{figures/suppl/mead2/expr_W015_mead.jpg}}\\[1pt]
  \noindent\makebox[0.085\textwidth][r]{\scriptsize \textbf{Ours}\hspace{3pt}}\raisebox{-0.5\height}{\includegraphics[width=0.915\textwidth]{figures/suppl/mead2/expr_W015_ours.jpg}}\\[1pt]
  \noindent\makebox[0.085\textwidth][r]{\scriptsize InSwapper\hspace{3pt}}\raisebox{-0.5\height}{\includegraphics[width=0.915\textwidth]{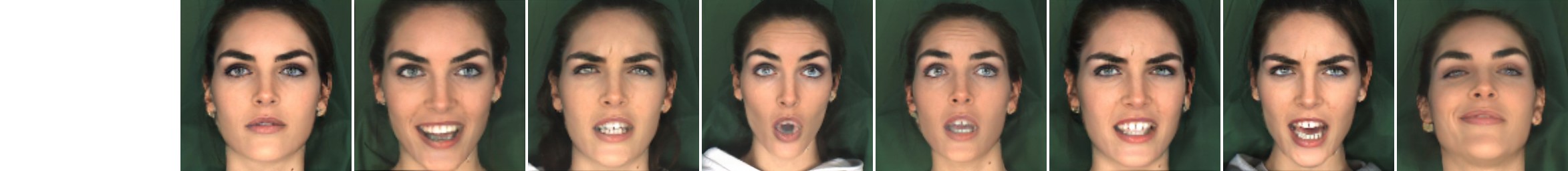}}\\[1pt]
  \noindent\makebox[0.085\textwidth][r]{\scriptsize Face-Adapter\hspace{3pt}}\raisebox{-0.5\height}{\includegraphics[width=0.915\textwidth]{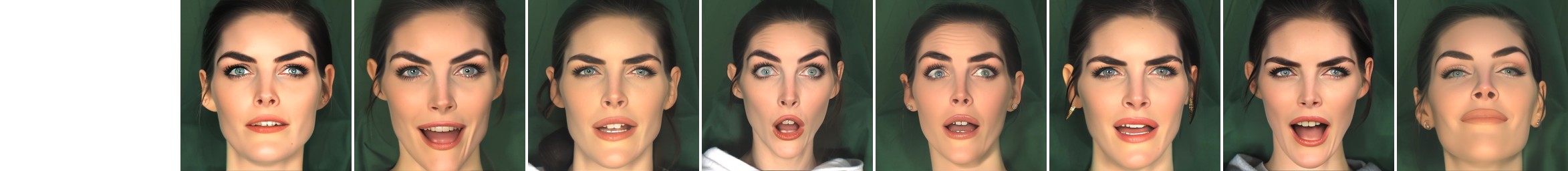}}\\[1pt]
  \noindent\makebox[0.085\textwidth][r]{\scriptsize DreamID-V\hspace{3pt}}\raisebox{-0.5\height}{\includegraphics[width=0.915\textwidth]{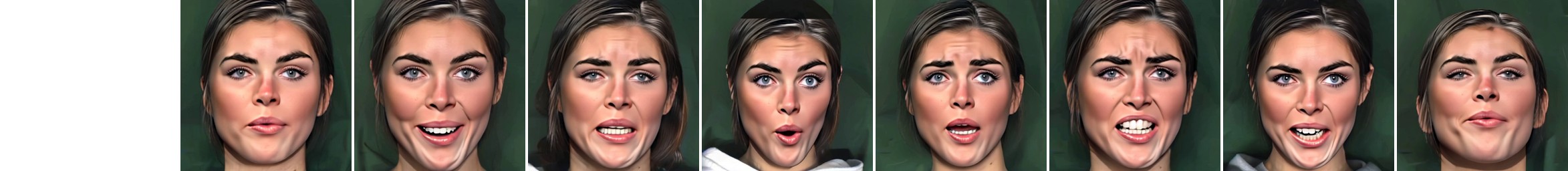}}\\[4pt]
  \noindent\makebox[0.085\textwidth][r]{\scriptsize MEAD\hspace{3pt}}\raisebox{-0.5\height}{\includegraphics[width=0.915\textwidth]{figures/suppl/mead2/expr_M033_mead.jpg}}\\[1pt]
  \noindent\makebox[0.085\textwidth][r]{\scriptsize \textbf{Ours}\hspace{3pt}}\raisebox{-0.5\height}{\includegraphics[width=0.915\textwidth]{figures/suppl/mead2/expr_M033_ours.jpg}}\\[1pt]
  \noindent\makebox[0.085\textwidth][r]{\scriptsize InSwapper\hspace{3pt}}\raisebox{-0.5\height}{\includegraphics[width=0.915\textwidth]{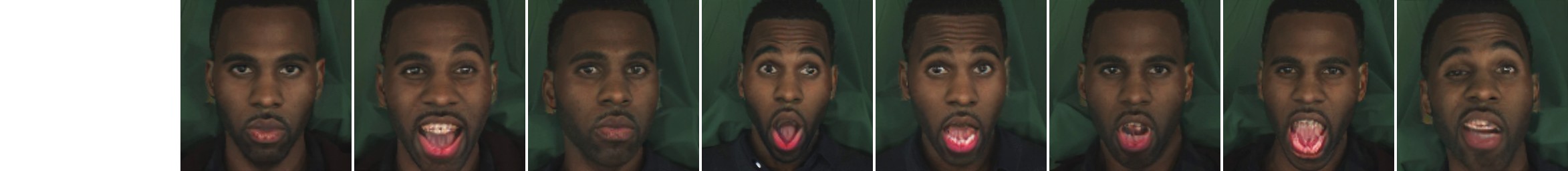}}\\[1pt]
  \noindent\makebox[0.085\textwidth][r]{\scriptsize Face-Adapter\hspace{3pt}}\raisebox{-0.5\height}{\includegraphics[width=0.915\textwidth]{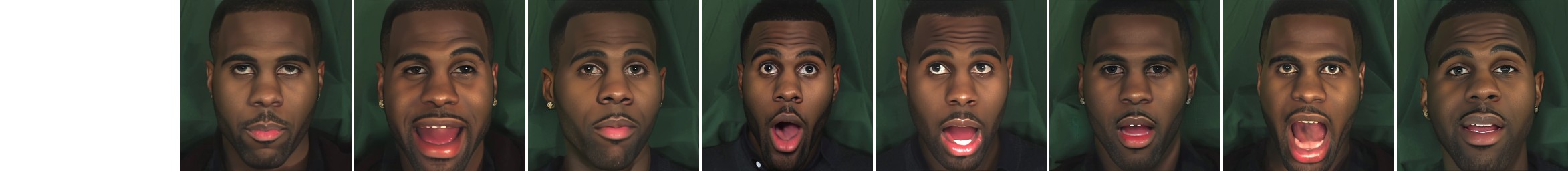}}\\[1pt]
  \noindent\makebox[0.085\textwidth][r]{\scriptsize DreamID-V\hspace{3pt}}\raisebox{-0.5\height}{\includegraphics[width=0.915\textwidth]{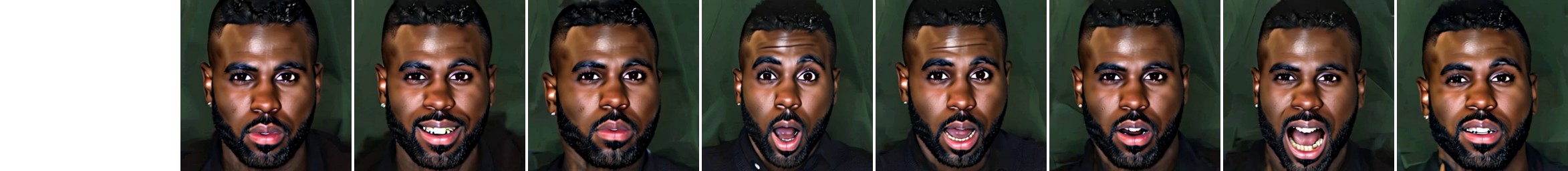}}
  \caption{\textbf{Emotions of MEAD, compared with prior work.} Two of the subjects of Fig.~\ref{fig:suppl:meadexpr} (top: a woman with light skin; bottom: a man with dark skin). Each method runs with the settings of Table~\ref{tab:main}.}
  \label{fig:suppl:meadcmp}
\end{figure*}

Figs.~\ref{fig:suppl:meadexpr} and~\ref{fig:suppl:meadpose} show four other MEAD
subjects, two with light and two with dark skin, each paired with a target of the
identity benchmark of the same sex and a similar skin tone. In
Fig.~\ref{fig:suppl:meadexpr}, the subjects show the eight emotions; in
Fig.~\ref{fig:suppl:meadpose}, a neutral face under the seven cameras. Our results
take on the open mouth of surprise and fear, the bared teeth of disgust and anger
and the frown of sadness, and they keep the face of the target from one profile
to the other and from above and below. Fig.~\ref{fig:suppl:meadcmp} compares the
methods on two of these subjects: the results of InSwapper and Face-Adapter show
weaker expressions, and DreamID-V generates the hair and the head anew. Over the
emotions of the four subjects, our results take the EmoNet class of the driving
image in $66\%$ of the cases, against $50\%$ for InSwapper and DreamID-V and
$30\%$ for Face-Adapter.
\begin{table*}[t]
  \centering\footnotesize
  \setlength{\tabcolsep}{5pt}
  \caption{\textbf{Results driven by MEAD frames.} Four targets, four held-out
  MEAD subjects. Match: share of results whose EmoNet class equals that of the
  driving image. VA: EmoNet valence-arousal distance. SMIRK: SMIRK expression
  error. ArcFace: mean cosine to the $24$ held-out photos of the target. Means
  over the results where all estimators find a face ($105$ to $116$ per method
  for the emotions, $65$ to $80$ for the views).}
  \label{tab:suppl:mead}
  \begin{tabular}{@{}l cccc cccc@{}}
    \toprule
    & \multicolumn{4}{c}{Emotions (front camera)} & \multicolumn{4}{c}{Camera views} \\
    \cmidrule(lr){2-5} \cmidrule(l){6-9}
    Method & Match$\uparrow$ & VA$\downarrow$ & SMIRK$\downarrow$ & ArcFace$\uparrow$ & Match$\uparrow$ & VA$\downarrow$ & SMIRK$\downarrow$ & ArcFace$\uparrow$ \\
    \midrule
    InSwapper & 0.54 & 0.34 & \textbf{0.59} & \textit{0.718} & 0.87 & 0.17 & 0.66 & \textit{0.708} \\
    Face-Adapter & 0.35 & 0.35 & 0.73 & 0.464 & 0.89 & 0.21 & 0.72 & 0.443 \\
    DreamID-V & 0.41 & 0.39 & 0.90 & 0.510 & 0.86 & 0.30 & 0.78 & 0.509 \\
    \midrule
    \textbf{Ours} & \textbf{0.68} & \textbf{0.20} & 0.63 & \textbf{0.567} & \textbf{1.00} & \textbf{0.12} & \textbf{0.62} & \textbf{0.527} \\
    \bottomrule
  \end{tabular}
\end{table*}

\subsection{Lighting}
\label{sec:suppl:driving:light}
\begin{figure*}[p]
  \centering
  \edef\lcell{\the\dimexpr 0.1260\textwidth\relax}\edef\lgap{\the\dimexpr 0.0030\textwidth\relax}%
  \makebox[0.90\textwidth][l]{\makebox[\lcell][c]{\scriptsize\shortstack{Target}}\hspace{\lgap}\makebox[\lcell][c]{\scriptsize\shortstack{Driving}}\hspace{\lgap}\makebox[\lcell][c]{\scriptsize\shortstack{InSwapper}}\hspace{\lgap}\makebox[\lcell][c]{\scriptsize\shortstack{DreamID-V}}\hspace{\lgap}\makebox[\lcell][c]{\scriptsize\shortstack{Face-Adapter}}\hspace{\lgap}\makebox[\lcell][c]{\scriptsize\shortstack{\textbf{Ours}}}\hspace{\lgap}\makebox[\lcell][c]{\scriptsize\shortstack{Ours, lighting\\with ears, neck}}}\\[2pt]
  \includegraphics[width=0.90\textwidth]{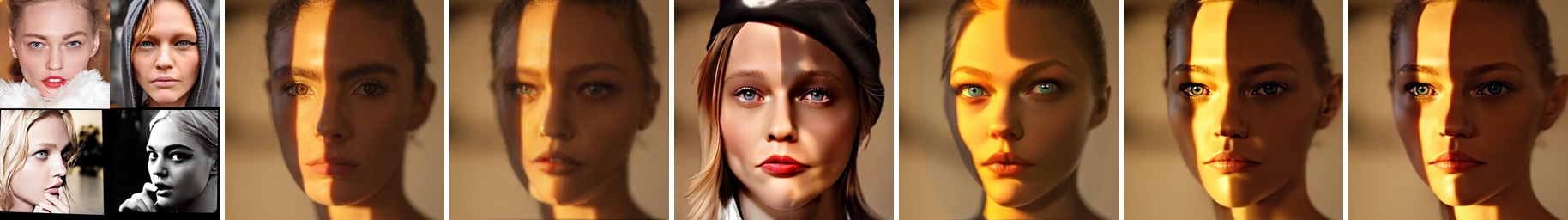}\\[\lgap]
  \includegraphics[width=0.90\textwidth]{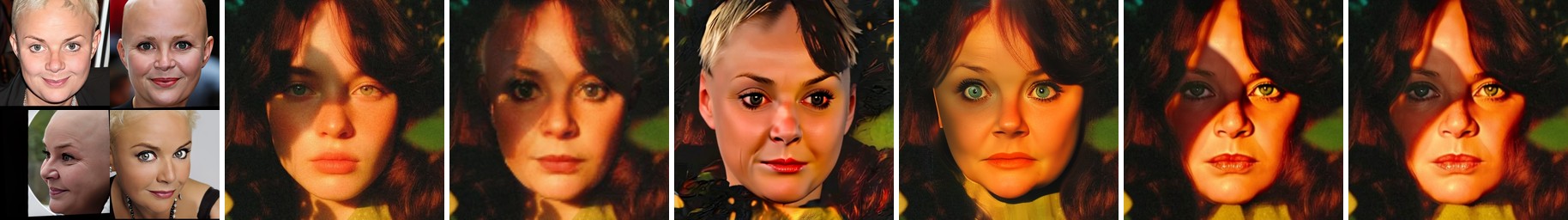}\\[\lgap]
  \includegraphics[width=0.90\textwidth]{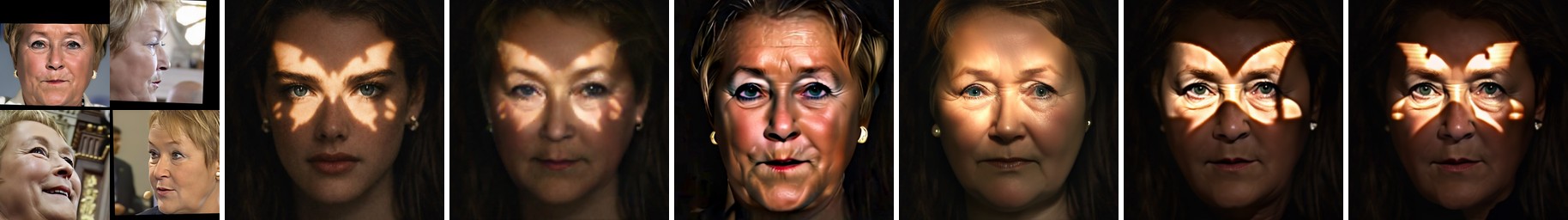}\\[\lgap]
  \includegraphics[width=0.90\textwidth]{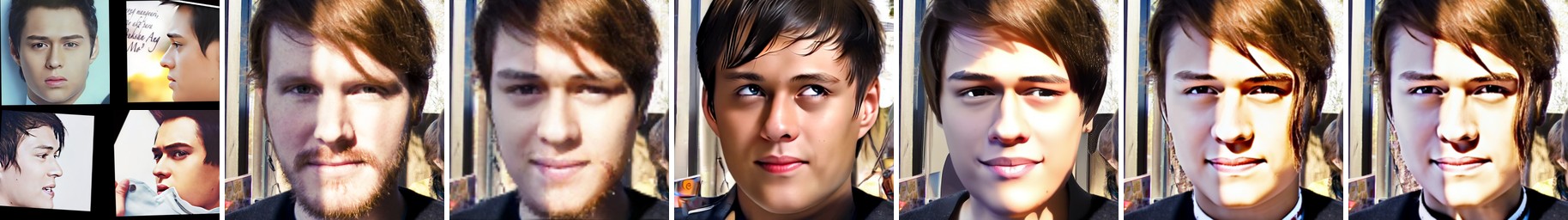}\\[\lgap]
  \includegraphics[width=0.90\textwidth]{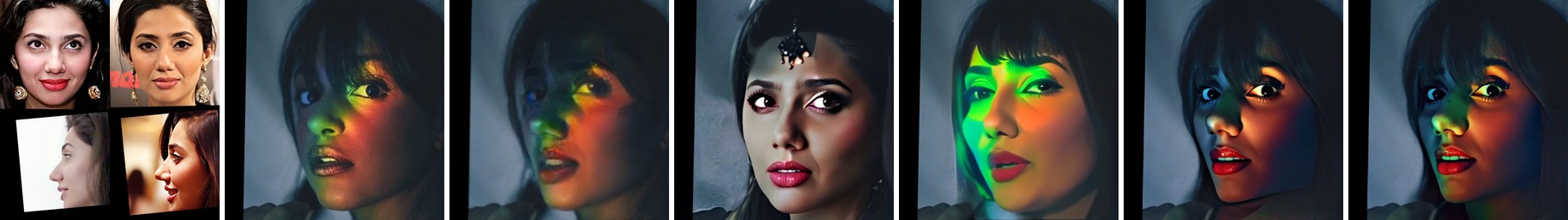}\\[\lgap]
  \includegraphics[width=0.90\textwidth]{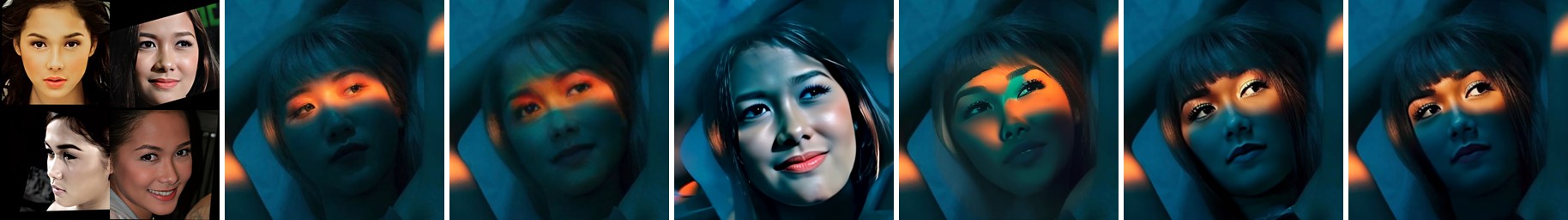}\\[\lgap]
  \includegraphics[width=0.90\textwidth]{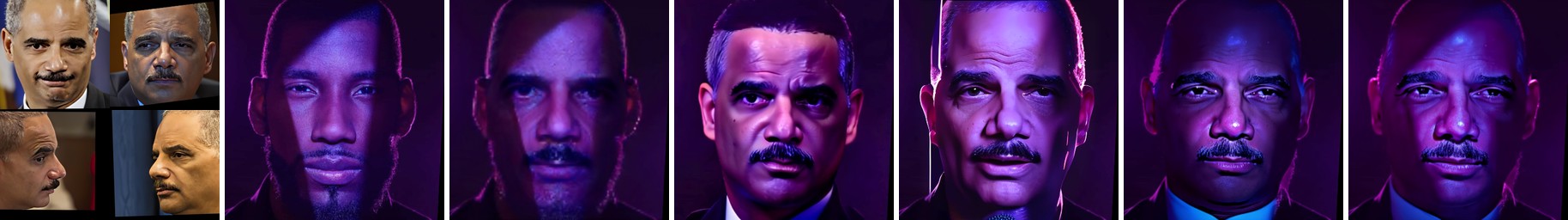}\\[\lgap]
  \includegraphics[width=0.90\textwidth]{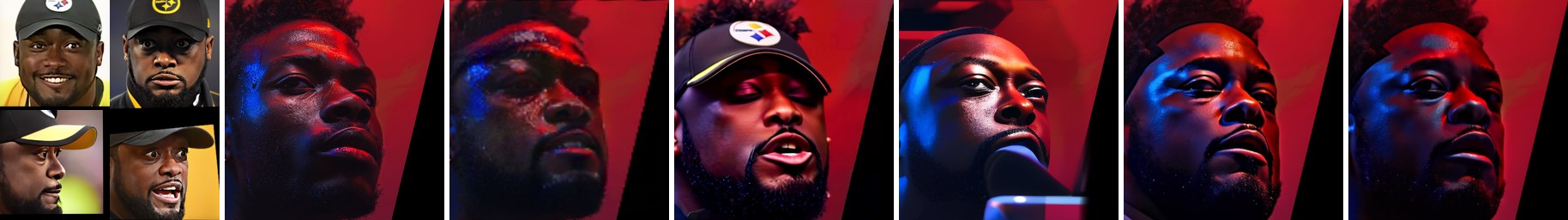}
  \caption{\textbf{Results under strong lighting.} Rows 1--4: a stripe of sunlight, sunlight through leaves, a projected pattern and a hard side light; rows 5--8: colored light. Ours: the default model, two passes with the warp; last column: the lighting stream whose input also covers the ears and neck. Target: four of its $24$ photos. Each method runs with the settings of Table~\ref{tab:main}.}
  \label{fig:suppl:light}
\end{figure*}
In film, faces are often lit by hard or colored light: a stripe of sunlight,
light through leaves, a projected pattern or a colored side light.
Fig.~\ref{fig:suppl:light} shows results under such lighting, with driving
images from the web and from FFHQ~\cite{karras2019stylegan} and targets from
VGGFace2-HQ that are not targets of the identity benchmark and not close to an
identity of our training data (Sec.~\ref{sec:suppl:bench}). Our model runs with
its default setting, two passes with the warp (Sec.~\ref{sec:method:inference}).
InSwapper, DreamID-V and Face-Adapter run with the settings of
Table~\ref{tab:main}.

Our results take on the
stripe of sunlight, the light through leaves, the projected pattern and the
colored light on the face. InSwapper also keeps the light of the
driving image, Face-Adapter keeps part of it, and DreamID-V generates the face
under a light of its own.

\begin{figure}[t]
  \centering
  \makebox[\linewidth][l]{\makebox[0.2470\linewidth][c]{\scriptsize Target}\hspace{0.0058\linewidth}\makebox[0.2470\linewidth][c]{\scriptsize Driving}\hspace{0.0058\linewidth}\makebox[0.2470\linewidth][c]{\scriptsize Ours, face}\hspace{0.0058\linewidth}\makebox[0.2470\linewidth][c]{\scriptsize Ours, ears and neck}}\\[1pt]
  \includegraphics[width=\linewidth]{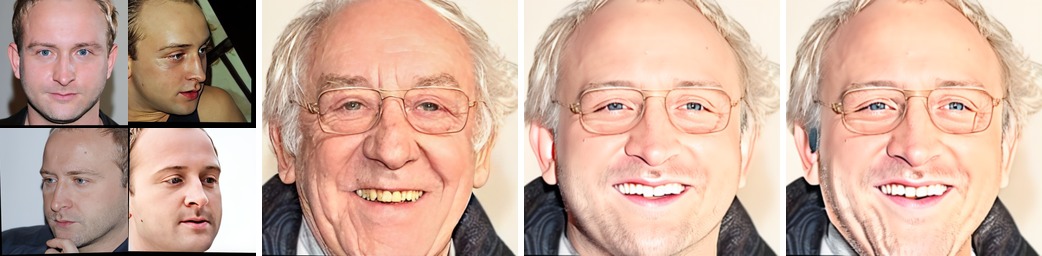}\\[2pt]
  \includegraphics[width=\linewidth]{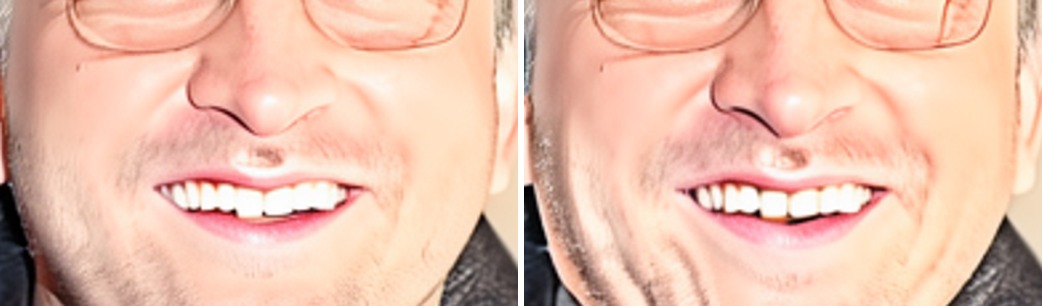}
  \caption{\textbf{Texture from the driving image.} A younger target with an older
  driving image, with the two lighting streams. Bottom: the cheeks of
  the two results. With the stream that also reads the ears and neck, folds of
  the older face appear on the cheek and the jaw.}
  \label{fig:suppl:lightleak}
\end{figure}

The last column uses a second lighting stream, which offers a trade-off between
closer lighting and a cleaner identity; the user can choose between the two
streams for each task. It reads a larger input region that covers the ears and
neck of the driving image as well as the face. It was trained in Stage~3 on another Stage-2 model and is used with ours
unchanged. It follows the light more closely, as in the projected pattern, the
rainbow and the colored light on dark skin (rows 3, 5, 7 and 8). On the identity
benchmark, its lighting correlation rises from $0.83$ to $0.84$, while ArcFace
falls from $0.601$ to $0.589$ (Table~\ref{tab:suppl:lightstream}). The face
randomization of Stage~3 replaces the face of the lighting input with that of
another person, but the model it uses leaves the ears and neck unchanged. In
training, these regions therefore show the person of the ground truth, which
gives the stream a path to copy from its input. When the target and the person
in the driving image differ in fine detail, some texture of the driving image
can then appear in the result, such as the wrinkles of an older face
(Fig.~\ref{fig:suppl:lightleak}). The lighting examples of Fig.~\ref{fig:teaser} (top left) use this
stream.
\begin{table*}[t]
  \centering\footnotesize
  \setlength{\tabcolsep}{6pt}
  \caption{\textbf{The two lighting streams} on the identity benchmark ($240$
  pairs) and the shape benchmark ($348$ pairs). Measures as in
  Tables~\ref{tab:main} and~\ref{tab:shape}; same seeds and settings, only the
  lighting stream differs.}
  \label{tab:suppl:lightstream}
  \begin{tabular}{@{}l cccccc cc@{}}
    \toprule
    Input of the stream & ArcFace$\uparrow$ & AdaFace$\uparrow$ & Expr.$\downarrow$ & Pose$\downarrow$ & Light$\uparrow$ & FaceScore$\uparrow$ & $s_\mathrm{gt}\downarrow$ & Leak$\downarrow$ \\
    \midrule
    Face (default) & \textbf{0.601} & \textbf{0.573} & 0.253 & 3.99 & 0.83 & \textbf{4.62} & \textbf{0.083} & 0.34 \\
    Face, ears and neck & 0.589 & 0.562 & 0.245 & 3.85 & \textbf{0.84} & 4.49 & 0.086 & 0.34 \\
    \bottomrule
  \end{tabular}
\end{table*}

\begin{figure}[t]
  \centering
  \edef\wcell{\the\dimexpr 0.2460\linewidth\relax}\edef\wgap{\the\dimexpr 0.0054\linewidth\relax}%
  \makebox[\linewidth][l]{\makebox[\wcell][c]{\scriptsize\shortstack{Target}}\hspace{\wgap}\makebox[\wcell][c]{\scriptsize\shortstack{Driving}}\hspace{\wgap}\makebox[\wcell][c]{\scriptsize\shortstack{One pass\\no warp}}\hspace{\wgap}\makebox[\wcell][c]{\scriptsize\shortstack{\textbf{Two passes}\\\textbf{with warp}}}}\\[2pt]
  \includegraphics[width=\linewidth]{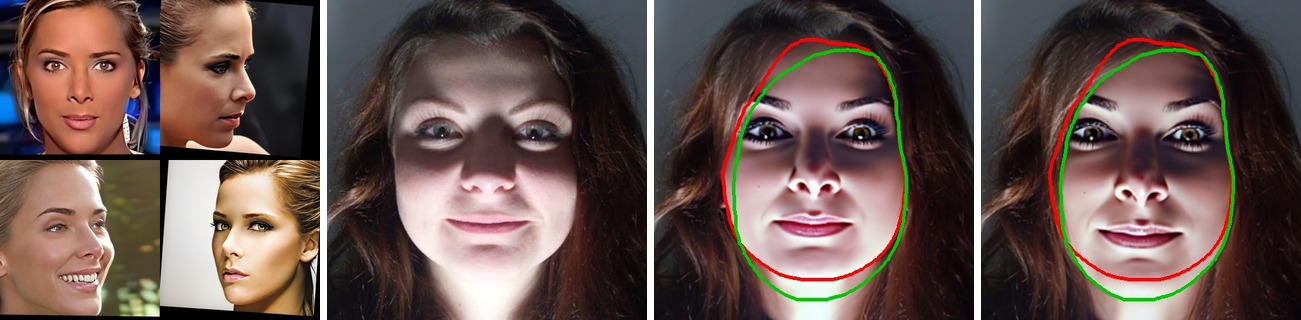}\\[\wgap]
  \includegraphics[width=\linewidth]{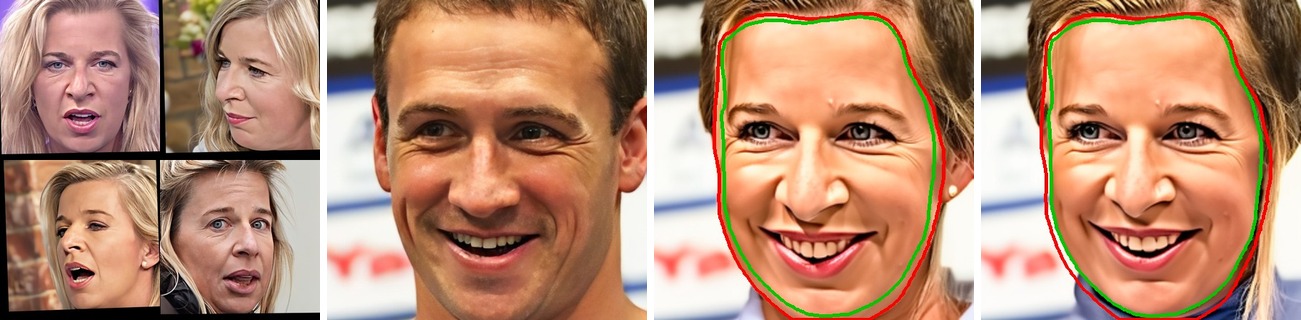}\\[\wgap]
  \includegraphics[width=\linewidth]{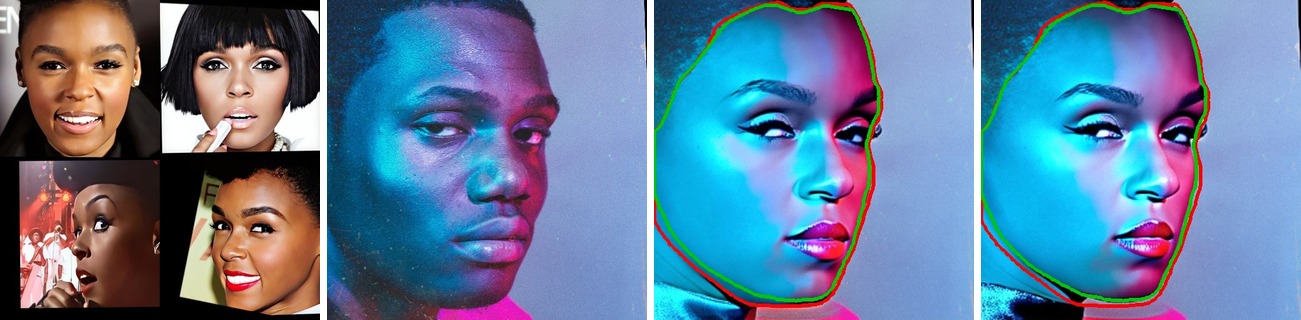}
  \caption{\textbf{The warp of the second pass.} Our results in one pass without the warp and in two passes with the warp (our default). Both images of a row carry the same two outlines of the face: one pass (red) and two passes (green). Top: a driving image lit from below. Middle and bottom: a female target with a male driving image. Without the warp, the result takes on part of the face shape of the driving image.}
  \label{fig:suppl:lightwarp}
\end{figure}

Fig.~\ref{fig:suppl:lightwarp} shows why the second pass warps the lighting
input. In one pass, the lighting stream reads the face of the driving image, and
the result takes on part of its shape: the round face lit from below in the top row
and the wide jaws of the men in the two rows below it. With the warp, the lighting
input follows the face of the first pass, and the face of the result stays closer to
the target's under the same light. The outlines in the figure show the change:
with the warp, the jaw is narrower in all three rows.

\subsection{Occluders}
\label{sec:suppl:driving:occ}
\begin{figure*}[p]
  \centering
  \edef\ocell{\the\dimexpr 0.1260\textwidth\relax}\edef\ogap{\the\dimexpr 0.0030\textwidth\relax}%
  \makebox[0.90\textwidth][l]{\makebox[\ocell][c]{\scriptsize Target}\hspace{\ogap}\makebox[\ocell][c]{\scriptsize Driving}\hspace{\ogap}\makebox[\ocell][c]{\scriptsize Occluder map}\hspace{\ogap}\makebox[\ocell][c]{\scriptsize InSwapper}\hspace{\ogap}\makebox[\ocell][c]{\scriptsize DreamID-V}\hspace{\ogap}\makebox[\ocell][c]{\scriptsize Face-Adapter}\hspace{\ogap}\makebox[\ocell][c]{\scriptsize \textbf{Ours}}}\\[2pt]
  \includegraphics[width=0.90\textwidth]{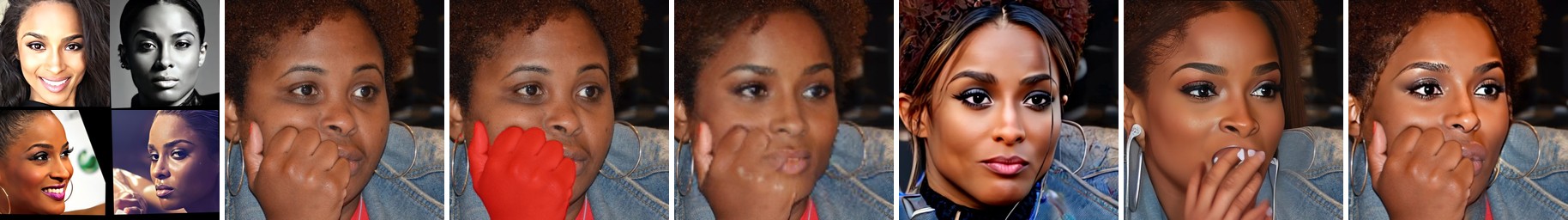}\\[\ogap]
  \includegraphics[width=0.90\textwidth]{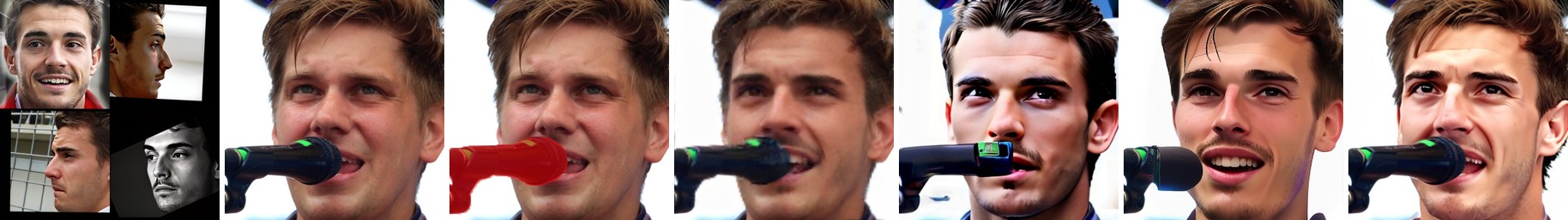}\\[\ogap]
  \includegraphics[width=0.90\textwidth]{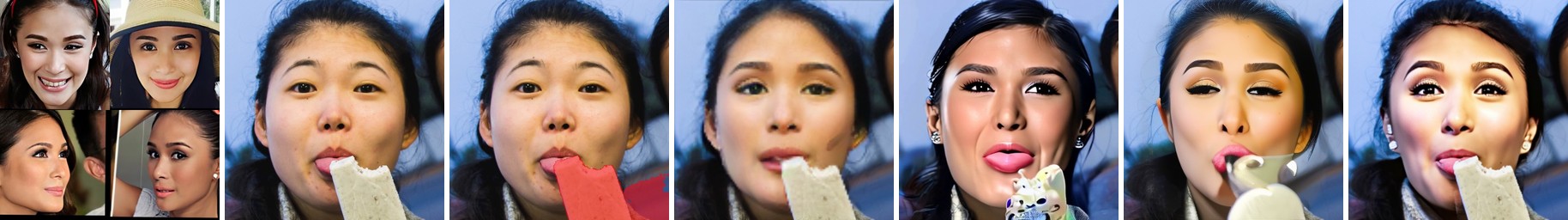}\\[\ogap]
  \includegraphics[width=0.90\textwidth]{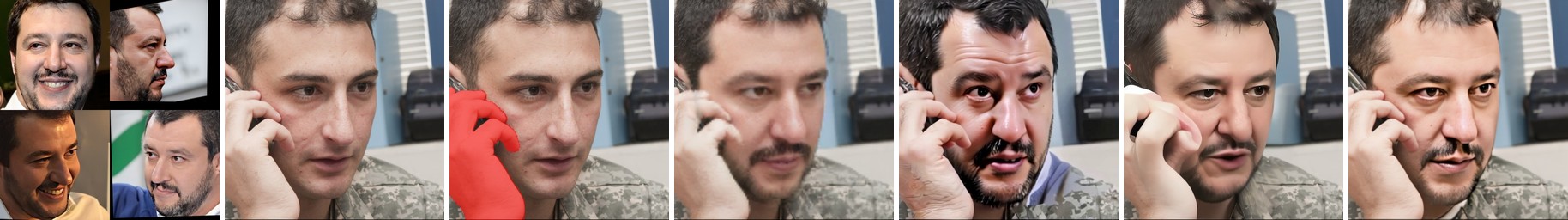}\\[\ogap]
  \includegraphics[width=0.90\textwidth]{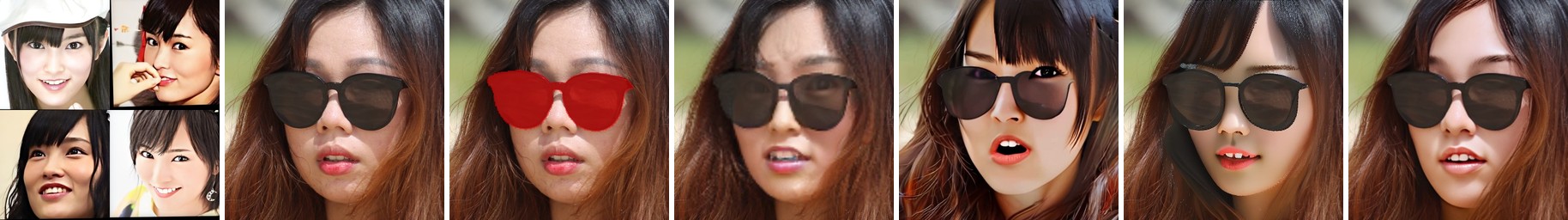}\\[\ogap]
  \includegraphics[width=0.90\textwidth]{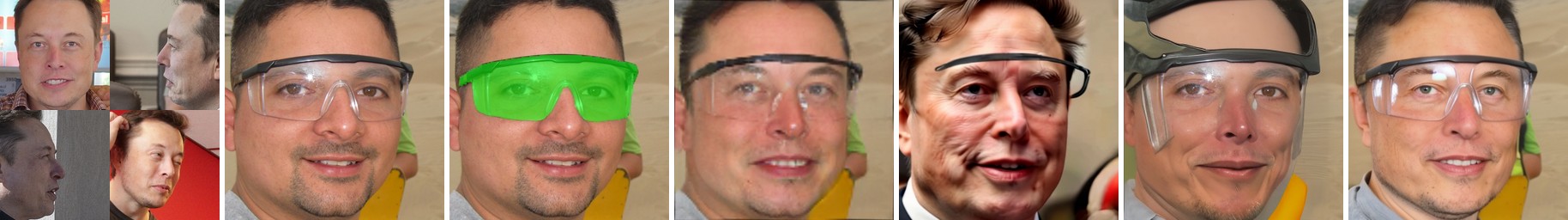}\\[\ogap]
  \includegraphics[width=0.90\textwidth]{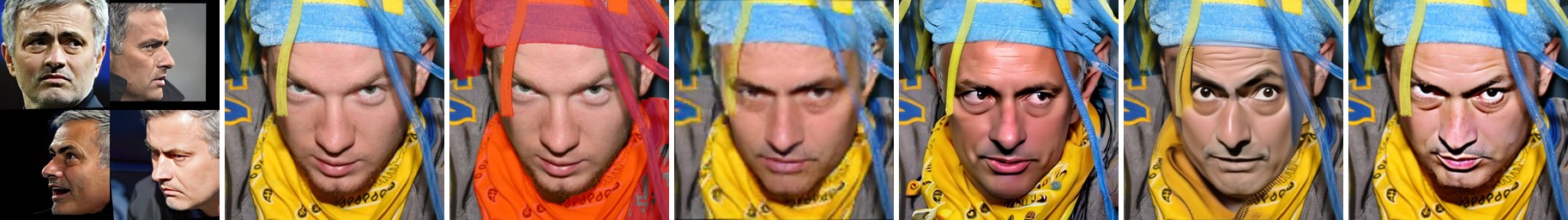}\\[\ogap]
  \includegraphics[width=0.90\textwidth]{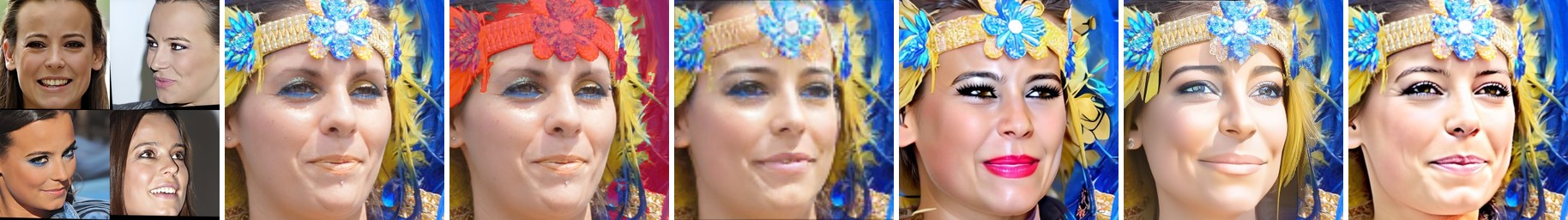}\\[\ogap]
  \includegraphics[width=0.90\textwidth]{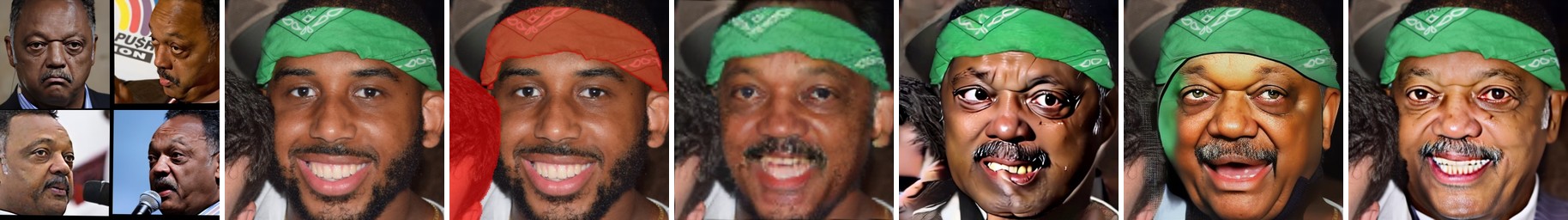}
  \caption{\textbf{Results with occluders.} Rows: a hand, a microphone, food, a phone and a hand, sunglasses, clear safety glasses, a scarf and a headband, a headdress, and a bandana. Occluder map: the occluders found in the driving image (red) and clear glasses (green), as in Sec.~\ref{sec:suppl:driving:occ}. The clear glasses of row~6 are not an occluder; they follow the rule for glasses (Sec.~\ref{sec:suppl:impl:inference}), and the eyes behind them are generated again. Target: four of its photos. Each method runs with the settings of Table~\ref{tab:main}.}
  \label{fig:suppl:occ}
\end{figure*}

In VFX, the mask that decides what stays from the driving image is an input that
artists often refine for each shot. Our default finds the occluders with a
general detection chain (Sec.~\ref{sec:suppl:impl:inference}); for thin or
unusual objects, a finer detector, another mask design or a person in the loop
can refine it, without any change to the model.
Fig.~\ref{fig:suppl:occ} shows results with driving images from
FFHQ~\cite{karras2019stylegan} in which something covers part of the face, such
as a hand, a microphone, food, a phone, a scarf or headwear, with targets from
VGGFace2-HQ that are not targets of the identity benchmark and not close to an
identity of our training data (Sec.~\ref{sec:suppl:bench}). Row~6 is a celebrity
pair of the user study whose driving image shows clear safety glasses. Our model
runs with its default setting; as an example of such a refinement, in rows~7
and~8 the detection also receives the names of the thin objects (a headband,
strings). InSwapper, DreamID-V and Face-Adapter run with the settings of
Table~\ref{tab:main}.

Our results keep the occluders of the driving image in front
of the face of the target. InSwapper
keeps them too, as it changes the inner face only. DreamID-V removes the hand and
replaces the microphone and the food, and Face-Adapter loses most of the hand and
changes the shape of the food.

\section{Face shape}
\label{sec:suppl:shape}
\label{tab:suppl:shape}%

\begin{table}[t]
  \centering\footnotesize
  \setlength{\tabcolsep}{4pt}
  \caption{\textbf{Face shape on the shape benchmark, all methods} ($348$
  pairs). $s_\mathrm{gt}$: distance from the jawline of the result to that of the ground
  truth. Leak: $0$ when the result ignores the jawline of the driving image, $1$
  when it copies it. 3D: distance between the reconstructed 3D face shapes of
  the result and the ground truth, outside the hair. The ground truth under one
  other lighting scores $0.024$ ($s_\mathrm{gt}$) and $1.14$ (3D). Each method receives $24$ renders of the target and runs at its
  setting of Table~\ref{tab:suppl:allmetrics}; GPT-Image uses the first $8$ and Gemini-Pro the first one, on $347$ pairs each, as their safety filters blocked one. Rows of ours: resolution and inference mask; the face mask covers the face, ears and neck, and the face+hair mask also the hair, which the model then generates. Bold: best.}
  \label{tab:suppl:shapeall}
  \begin{tabular}{@{}l c c c@{}}
    \toprule
    Method & $s_\mathrm{gt}\downarrow$ & Leak$\downarrow$ & 3D$\downarrow$ \\
    \midrule
    SimSwap & 0.125 & 0.95 & 1.83 \\
    FaceDancer & 0.122 & 0.91 & 1.85 \\
    BlendFace & 0.119 & 0.86 & 1.99 \\
    GHOST & 0.132 & 0.90 & 2.21 \\
    InSwapper & 0.116 & 0.81 & 1.48 \\
    HifiFace & 0.138 & 0.78 & 2.03 \\
    Face-Adapter & 0.138 & 0.60 & 1.86 \\
    DreamID-V & 0.149 & 0.52 & 1.67 \\
    DiffSwap & 0.125 & 0.92 & 2.53 \\
    REFace & 0.135 & 0.94 & 2.18 \\
    DiffFace & 0.120 & 0.84 & 1.92 \\
    E4S & 0.131 & 0.77 & 1.93 \\
    GPT-Image & 0.127 & 0.75 & 1.62 \\
    Gemini-Pro & 0.103 & 0.47 & 1.49 \\
    \midrule
    \textbf{Ours}, 256\,px, face & 0.083 & 0.34 & 1.35 \\
    \quad 512\,px, face & \textbf{0.082} & \textbf{0.30} & \textbf{1.34} \\
    \quad 256\,px, face+hair & \textbf{0.082} & 0.32 & 1.35 \\
    \bottomrule
  \end{tabular}
\end{table}

\begin{figure*}[t]
  \centering
  \edef\ucell{\the\dimexpr 0.0816\textwidth\relax}\edef\ugap{\the\dimexpr 0.0019\textwidth\relax}%
  \makebox[1.00\textwidth][l]{\makebox[\ucell][c]{\scriptsize\shortstack{Target}}\hspace{\ugap}\makebox[\ucell][c]{\scriptsize\shortstack{Driving}}\hspace{\ugap}\makebox[\ucell][c]{\scriptsize\shortstack{Ground\\truth}}\hspace{\ugap}\makebox[\ucell][c]{\scriptsize\shortstack{SimSwap}}\hspace{\ugap}\makebox[\ucell][c]{\scriptsize\shortstack{FaceDancer}}\hspace{\ugap}\makebox[\ucell][c]{\scriptsize\shortstack{BlendFace}}\hspace{\ugap}\makebox[\ucell][c]{\scriptsize\shortstack{GHOST}}\hspace{\ugap}\makebox[\ucell][c]{\scriptsize\shortstack{InSwapper}}\hspace{\ugap}\makebox[\ucell][c]{\scriptsize\shortstack{HifiFace}}\hspace{\ugap}\makebox[\ucell][c]{\scriptsize\shortstack{Face-Adapter}}\hspace{\ugap}\makebox[\ucell][c]{\scriptsize\shortstack{DreamID-V}}\hspace{\ugap}\makebox[\ucell][c]{\scriptsize\shortstack{\textbf{Ours}}}}\\[2pt]
  \includegraphics[width=1.00\textwidth]{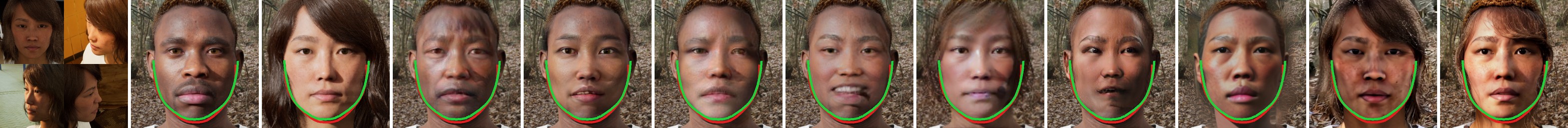}\\[\ugap]
  \includegraphics[width=1.00\textwidth]{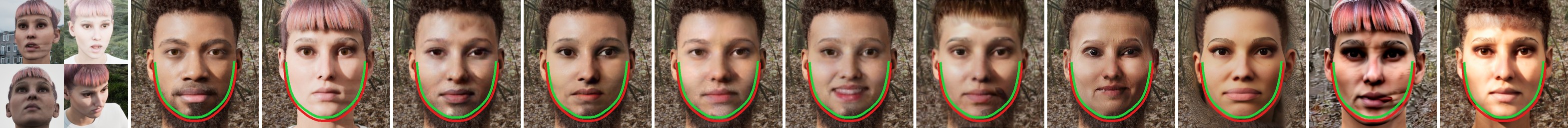}\\[\ugap]
  \includegraphics[width=1.00\textwidth]{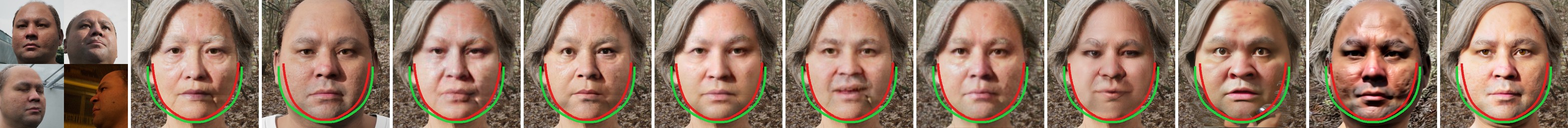}\\[\ugap]
  \includegraphics[width=1.00\textwidth]{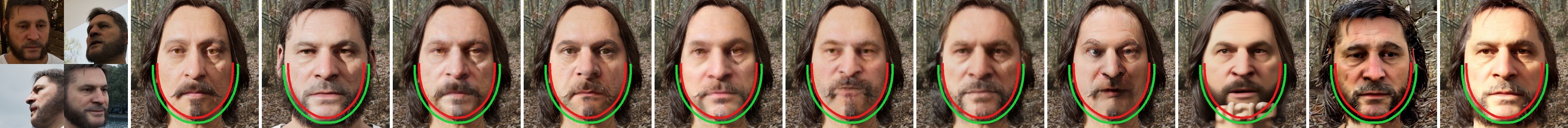}\\[\ugap]
  \includegraphics[width=1.00\textwidth]{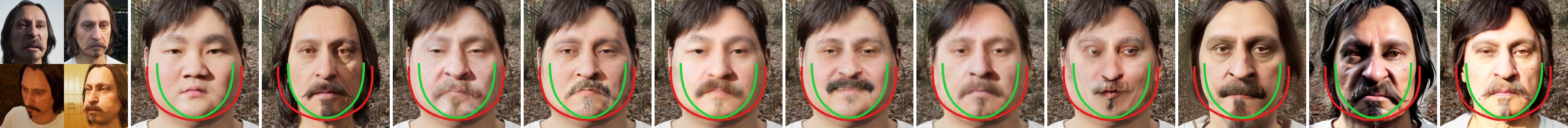}\\[\ugap]
  \includegraphics[width=1.00\textwidth]{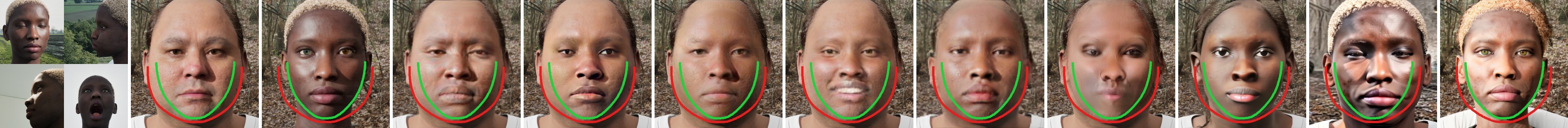}
  \caption{\textbf{Face shape on the shape benchmark.} Six pairs with the methods of Table~\ref{tab:main}, sorted by the distance between the jawlines of the ground truth and the driving image, from the smallest (top) to the largest (bottom). Target: four of its $24$ renders. Lines as in Fig.~\ref{fig:ablation}: jawline of the ground truth (green) and of the driving image (red).}
  \label{fig:suppl:ue5meth}
\end{figure*}

Fig.~\ref{fig:suppl:ue5meth} shows six pairs of the shape benchmark with the
methods of Table~\ref{tab:main}, sorted by how far the jawline of the driving
image lies from that of the ground truth. The results of the baselines stay close to
the jawline of the driving image, so they move away from the ground truth as this
distance grows: the mean $s_\mathrm{gt}$ of the eight baselines rises from
$0.08$ in the top row to $0.37$ in the bottom row, while that of our results stays
between $0.01$ and $0.07$. The same holds on the whole benchmark.
Table~\ref{tab:suppl:shapegap} splits the $348$ pairs into thirds by the distance
between the jawlines of the ground truth and the driving image. From the closest
third to the farthest, the mean $s_\mathrm{gt}$ of the eight baselines of
Table~\ref{tab:main} rises from $0.076$ to $0.116$ and $0.197$, and ours from
$0.065$ to $0.085$ and $0.099$. Where the two jawlines are close, the jawline of
the driving image is nearly correct: the driving image itself scores $0.052$, and
six of the twelve baselines and Gemini-Pro score slightly lower than ours ($0.053$
to $0.062$).

\begin{table}[t]
  \centering\footnotesize
  \setlength{\tabcolsep}{4pt}
  \caption{\textbf{Face shape by the distance between the jawlines of the ground truth and the driving image} (shape benchmark). The $348$ pairs are split into thirds by this distance (mean distance of all $21$ jaw points, in the units of $s_\mathrm{gt}$): $0.03$--$0.10$, $0.10$--$0.18$, $0.18$--$0.44$, with $116$ pairs each; GPT-Image and Gemini-Pro lack one (Table~\ref{tab:suppl:shapeall}). Mean $s_\mathrm{gt}$ per third. The driving image row scores the driving image itself. Rows of ours as in Table~\ref{tab:suppl:shapeall}. Bold: best.}
  \label{tab:suppl:shapegap}
  \begin{tabular}{@{}l cccc@{}}
    \toprule
    & \multicolumn{3}{c}{Third} & \\
    \cmidrule(lr){2-4}
    Method & Closest & Middle & Farthest & All \\
    \midrule
    Driving image & 0.052 & 0.105 & 0.226 & 0.128 \\
    \midrule
    SimSwap & \textbf{0.053} & 0.100 & 0.222 & 0.125 \\
    FaceDancer & 0.056 & 0.101 & 0.208 & 0.122 \\
    BlendFace & 0.054 & 0.098 & 0.205 & 0.119 \\
    GHOST & 0.069 & 0.108 & 0.218 & 0.132 \\
    InSwapper & 0.055 & 0.103 & 0.191 & 0.116 \\
    HifiFace & 0.088 & 0.122 & 0.203 & 0.138 \\
    Face-Adapter & 0.107 & 0.156 & 0.152 & 0.138 \\
    DreamID-V & 0.128 & 0.141 & 0.178 & 0.149 \\
    DiffSwap & 0.057 & 0.100 & 0.218 & 0.125 \\
    REFace & 0.070 & 0.107 & 0.228 & 0.135 \\
    DiffFace & 0.054 & 0.096 & 0.209 & 0.120 \\
    E4S & 0.095 & 0.111 & 0.188 & 0.131 \\
    GPT-Image & 0.072 & 0.100 & 0.208 & 0.127 \\
    Gemini-Pro & 0.062 & 0.094 & 0.153 & 0.103 \\
    \midrule
    \textbf{Ours}, 256\,px, face & 0.065 & 0.085 & 0.099 & 0.083 \\
    \quad 512\,px, face & 0.067 & \textbf{0.083} & 0.096 & \textbf{0.082} \\
    \quad 256\,px, face+hair & 0.067 & 0.084 & \textbf{0.095} & \textbf{0.082} \\
    \bottomrule
  \end{tabular}
\end{table}

\begin{table*}[t]
  \centering\footnotesize
  \setlength{\tabcolsep}{5pt}
  \caption{\textbf{Face shape by the gap between the driving image and the
  target} (shape benchmark). $s_\mathrm{gt}$ and 3D as in
  Table~\ref{tab:suppl:shapeall}. Sex and age are estimated by a face analysis
  model on the render of each character, and skin brightness is the mean
  lightness of the central face region (0--255). Different skin brightness: gap
  of at least $30$; different age: gap of at least $15$ years. Pairs per group,
  from left to right: $186$, $162$, $177$, $171$, $222$, $126$; GPT-Image and Gemini-Pro lack one pair each (Table~\ref{tab:suppl:shapeall}). The
  driving image row scores the driving image itself. Rows of ours as in Table~\ref{tab:suppl:shapeall}. Bold: best.}
  \label{tab:suppl:shapegroup}
  \begin{tabular}{@{}l cccccc cccccc@{}}
    \toprule
    & \multicolumn{6}{c}{$s_\mathrm{gt}\downarrow$} & \multicolumn{6}{c}{3D$\downarrow$} \\
    \cmidrule(lr){2-7} \cmidrule(l){8-13}
    & \multicolumn{2}{c}{Skin brightness} & \multicolumn{2}{c}{Sex} & \multicolumn{2}{c}{Age}
    & \multicolumn{2}{c}{Skin brightness} & \multicolumn{2}{c}{Sex} & \multicolumn{2}{c}{Age} \\
    \cmidrule(lr){2-3} \cmidrule(lr){4-5} \cmidrule(lr){6-7} \cmidrule(lr){8-9} \cmidrule(lr){10-11} \cmidrule(l){12-13}
    Method & sim. & diff. & same & diff. & sim. & diff. & sim. & diff. & same & diff. & sim. & diff. \\
    \midrule
    Driving image & 0.135 & 0.120 & 0.129 & 0.127 & 0.122 & 0.139 & 3.35 & 3.66 & 3.26 & 3.74 & 3.28 & 3.88 \\
    \midrule
    SimSwap & 0.132 & 0.117 & 0.129 & 0.121 & 0.116 & 0.141 & 1.73 & 1.94 & 1.78 & 1.88 & 1.72 & 2.03 \\
    FaceDancer & 0.128 & 0.114 & 0.125 & 0.118 & 0.115 & 0.134 & 1.77 & 1.93 & 1.87 & 1.82 & 1.75 & 2.01 \\
    BlendFace & 0.126 & 0.110 & 0.122 & 0.116 & 0.110 & 0.134 & 1.90 & 2.09 & 1.89 & 2.09 & 1.87 & 2.20 \\
    GHOST & 0.133 & 0.130 & 0.133 & 0.130 & 0.126 & 0.141 & 2.19 & 2.23 & 2.16 & 2.26 & 2.17 & 2.29 \\
    InSwapper & 0.120 & 0.111 & 0.119 & 0.113 & 0.105 & 0.136 & 1.43 & 1.54 & 1.44 & 1.52 & 1.42 & 1.58 \\
    HifiFace & 0.139 & 0.137 & 0.139 & 0.136 & 0.130 & 0.151 & 2.01 & 2.05 & 2.00 & 2.06 & 1.99 & 2.08 \\
    Face-Adapter & 0.138 & 0.139 & 0.148 & 0.129 & 0.131 & 0.151 & 1.83 & 1.89 & 1.93 & 1.78 & 1.81 & 1.94 \\
    DreamID-V & 0.148 & 0.151 & 0.147 & 0.152 & 0.138 & 0.170 & 1.64 & 1.71 & 1.71 & 1.63 & 1.65 & 1.71 \\
    DiffSwap & 0.132 & 0.117 & 0.126 & 0.123 & 0.118 & 0.137 & 2.43 & 2.63 & 2.51 & 2.54 & 2.45 & 2.66 \\
    REFace & 0.141 & 0.128 & 0.135 & 0.135 & 0.129 & 0.146 & 2.09 & 2.29 & 2.10 & 2.27 & 2.09 & 2.33 \\
    DiffFace & 0.127 & 0.111 & 0.123 & 0.116 & 0.111 & 0.134 & 1.86 & 2.00 & 1.91 & 1.93 & 1.90 & 1.97 \\
    E4S & 0.138 & 0.123 & 0.138 & 0.124 & 0.124 & 0.144 & 1.92 & 1.93 & 1.92 & 1.94 & 1.81 & 2.14 \\
    GPT-Image & 0.133 & 0.120 & 0.132 & 0.122 & 0.121 & 0.138 & 1.55 & 1.70 & 1.66 & 1.58 & 1.58 & 1.70 \\
    Gemini-Pro & 0.109 & 0.096 & 0.110 & 0.095 & 0.098 & 0.112 & 1.46 & 1.54 & 1.51 & 1.48 & 1.45 & 1.56 \\
    \midrule
    \textbf{Ours}, 256\,px, face & 0.084 & 0.082 & \textbf{0.085} & 0.081 & 0.078 & 0.092 & \textbf{1.27} & 1.44 & \textbf{1.32} & 1.38 & 1.33 & \textbf{1.39} \\
    \quad 512\,px, face & 0.086 & \textbf{0.078} & 0.086 & \textbf{0.078} & 0.079 & \textbf{0.088} & 1.29 & \textbf{1.40} & \textbf{1.32} & \textbf{1.36} & \textbf{1.28} & 1.44 \\
    \quad 256\,px, face+hair & \textbf{0.083} & 0.081 & \textbf{0.085} & 0.080 & \textbf{0.077} & 0.091 & 1.29 & 1.43 & 1.33 & 1.38 & 1.33 & 1.40 \\
    \bottomrule
  \end{tabular}
\end{table*}

Of the two commercial image models, Gemini-Pro has the lowest contour distance of
the baselines ($0.103$) and, with DreamID-V, the lowest leak ($0.47$ and $0.52$):
it copies about half of the jawline of the driving image. The driving images of
this benchmark face the camera with a neutral expression, and Gemini-Pro keeps
their pose: its 6DRepNet error, $2.4^\circ$, is at the level of a perfect result
(the ground truth scores $2.6^\circ$, Sec.~\ref{sec:suppl:exprfloor}). With photos as
driving images, on the identity benchmark, it has the largest EmoNet distance
and, after DreamID-V, the largest pose error (Table~\ref{tab:suppl:allmetrics}).
GPT-Image keeps most of the jawline of the driving image (leak $0.75$), and its
contour distance, $0.127$, is about that of the driving image itself ($0.128$).
Like DreamID-V, it generates the whole image and moves the head more than most
methods. The midpoint between the eyes moves from that of the driving image by
$0.07$ in the units of $s_\mathrm{gt}$ for GPT-Image and $0.14$ for DreamID-V,
against $0.02$ to $0.08$ for the other baselines and $0.04$ for ours; that of the
ground truth lies $0.03$ away, as its eyes are those of the target. To separate
the shape of the jawline from the position of the head, we also measure
$s_\mathrm{gt}$ after a shift and scale that put the eyes of each result on
those of the ground truth. Ours then scores $0.065$, and the closest baselines
$0.096$ (Gemini-Pro), $0.100$ (InSwapper) and $0.101$ (DreamID-V).

\paragraph{Mask size.}
The face mask is dilated by $12$\,px at inference
(Sec.~\ref{sec:suppl:impl:inference}), which leaves room around the face of the
driving image for a wider or longer face. Table~\ref{tab:suppl:masksize} changes
this dilation on the shape benchmark. A smaller mask ($6$\,px) leaves less room,
so more of the face shape of the driving image stays in the result (leak $0.41$
against $0.34$). A larger mask does not make the face fill it: from $12$ to
$24$\,px, the edge of the mask moves out by $12$\,px, but the jaw of the result by
only about $2$\,px. With this room, the result moves further from the jawline of
the driving image (leak $0.28$) and ArcFace to the ground truth rises, while the
jaw ends slightly outside that of the target, likely because the larger mask
also regenerates more of the hair and neck around the face, which fix the scale
of the head in the driving image. The contour distance and the 3D error
therefore rise a little. With a mask of the whole image, nothing of the hair and
neck is kept, and the face grows further ($s_\mathrm{gt}$ $0.113$), while the
leak stays at $0.28$. At every dilation, our results remain ahead of all the
baselines of Table~\ref{tab:suppl:shapeall} on all three shape measures; with the
mask of the whole image, Gemini-Pro has a lower contour distance ($0.103$ against
$0.113$). The
default of $12$\,px has the lowest contour distance and 3D error of all the
masks, and $18$\,px is close to it.
\begin{table}[t]
  \centering\footnotesize
  \caption{\textbf{Size of the inference mask} (shape benchmark, $348$ pairs).
  Dilation of the face mask at $256$\,px; whole image: the mask covers the whole
  image. $s_\mathrm{gt}$, leak and 3D as in
  Table~\ref{tab:suppl:shapeall}. ArcFace: cosine similarity between the result and
  the ground truth, a render of the target; the characters have no held-out
  photos. Bold: best.}
  \label{tab:suppl:masksize}
  \begin{tabular}{@{}l cccc@{}}
    \toprule
    Dilation & $s_\mathrm{gt}\downarrow$ & Leak$\downarrow$ & 3D$\downarrow$ & ArcFace$\uparrow$ \\
    \midrule
    $6$\,px & 0.090 & 0.41 & 1.36 & 0.761 \\
    $12$\,px (default) & \textbf{0.083} & 0.34 & \textbf{1.35} & 0.772 \\
    $18$\,px & 0.084 & 0.32 & 1.36 & 0.778 \\
    $24$\,px & 0.091 & \textbf{0.28} & 1.38 & 0.782 \\
    Whole image & 0.113 & \textbf{0.28} & 1.36 & \textbf{0.792} \\
    \bottomrule
  \end{tabular}
\end{table}

\section{More photos of the target}
\label{sec:suppl:msweep}
This section completes Sec.~\ref{sec:exp:msweep}. The protocol is the same: each
method receives the first $M$ photos of the fixed order of
Sec.~\ref{sec:suppl:idbench}, and the baselines average their identity inputs
over these photos.

\begin{figure*}[t]
  \centering
  \grpgraphics[width=0.33\linewidth]{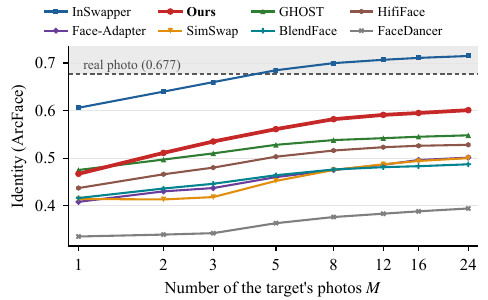}\hfill
  \grpgraphics[width=0.33\linewidth]{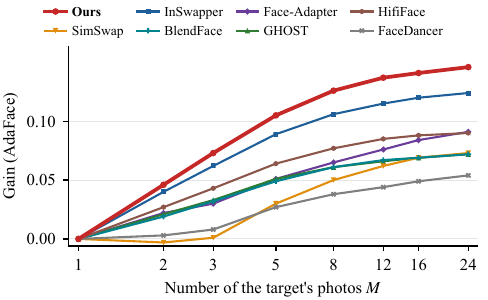}\hfill
  \grpgraphics[width=0.33\linewidth]{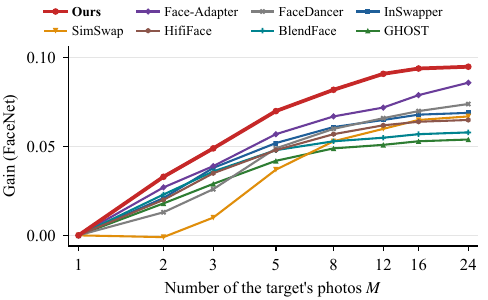}
  \caption{\textbf{Identity against the number of the target's photos.} Left:
  ArcFace score, with the level of a real photo of the target (dashed). Middle
  and right: gain over $M{=}1$ under AdaFace and FaceNet. $240$ pairs of the
  identity benchmark; the baselines average their identity inputs over the $M$
  photos.}
  \label{fig:suppl:msweep}
\end{figure*}

\paragraph{Identity.}
Fig.~\ref{fig:suppl:msweep} shows the ArcFace score itself, with the level of a
real photo of the target, and the gain over $M{=}1$ under AdaFace and FaceNet.
Under all three recognizers, our results gain the most from $M{=}1$ to $24$:
$0.146$ under AdaFace, against $0.054$ to $0.124$ for the baselines, and $0.095$
under FaceNet, against $0.054$ to $0.086$. Under ArcFace, our score rises from
$0.467$ to $0.601$. InSwapper starts higher, at $0.606$, and passes the level of a
real photo of the target from $M{=}5$ on.
InSwapper's identity encoder is not our ArcFace model but a network of the same
kind (ResNet-50, ArcFace loss) trained on other data, and the two judge
similarity alike: on $141$ photos, their similarity matrices correlate at
$0.98$. A result fitted to InSwapper's encoder can therefore score high under our
ArcFace.

\paragraph{Face shape.}
On the shape benchmark, our contour distance is $0.090$, $0.085$, $0.086$,
$0.089$ and $0.083$ at $M{=}1$, $2$, $4$, $8$ and $24$. The difference between
$M{=}1$ and $24$ is clear, $0.007$ with a $95\%$ interval of $[0.003, 0.012]$ over
the pairs, although the distance does not fall at every step. The contour distance
of InSwapper is $0.115$ at $M{=}1$ and $0.116$ at $M{=}24$. Without the
disentanglement strategies (first row of Table~\ref{tab:ablation}), our model
stays at $0.125$ and $0.124$, close to the $0.128$ of the driving image.

\begin{figure}[t]
  \centering
  \edef\acell{\the\dimexpr 0.1963\linewidth\relax}\edef\agap{\the\dimexpr 0.0046\linewidth\relax}%
  \makebox[\linewidth][l]{\makebox[\acell][c]{\scriptsize\shortstack{Target}}\hspace{\agap}\makebox[\acell][c]{\scriptsize\shortstack{Driving}}\hspace{\agap}\makebox[\acell][c]{\scriptsize\shortstack{Average}}\hspace{\agap}\makebox[\acell][c]{\scriptsize\shortstack{Average,\\rescaled}}\hspace{\agap}\makebox[\acell][c]{\scriptsize\shortstack{\textbf{Ours}}}}\\[2pt]
  \includegraphics[width=\linewidth]{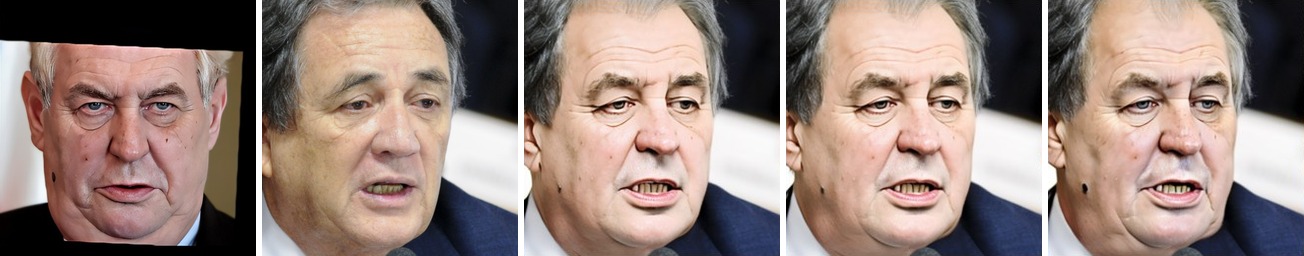}\\[\agap]
  \includegraphics[width=\linewidth]{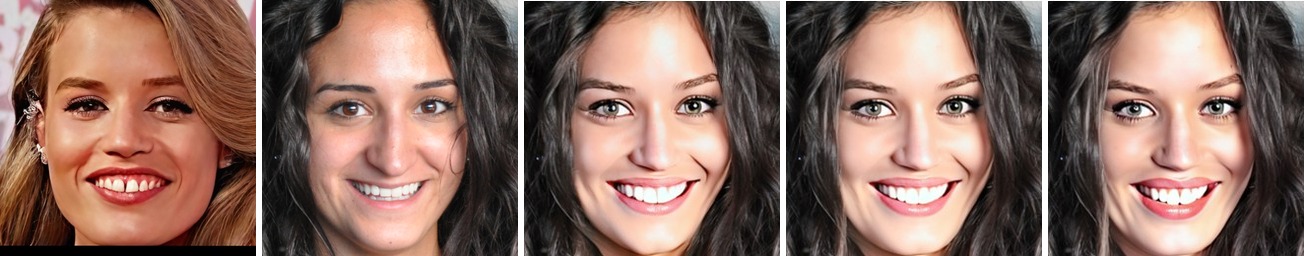}\\[\agap]
  \includegraphics[width=\linewidth]{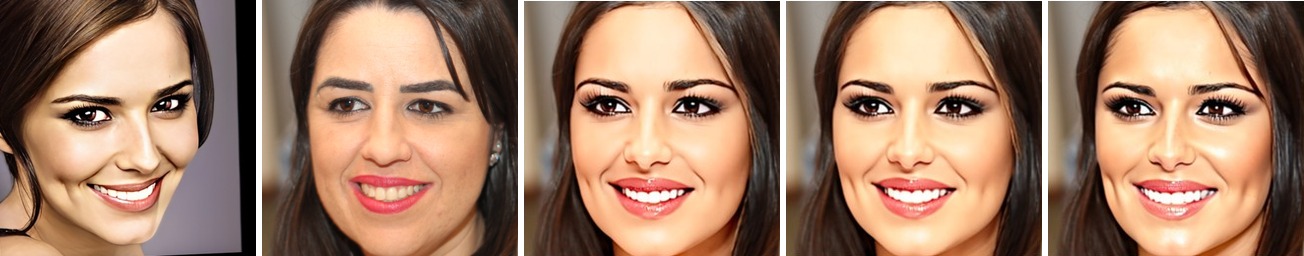}
  \caption{\textbf{Averaged identity tokens.} Ours: the identity encoder reads the $24$ photos of the target together. Average: each photo is encoded alone, and the tokens are averaged position by position. Average, rescaled: the same tokens, rescaled to the norm of the single-photo tokens. Rows: the three targets of Fig.~\ref{fig:suppl:msweepdetail}, with a mole on the cheek, a gap between the front teeth and dimples. Target: a photo that shows the detail. ArcFace of the three pairs, row by row: $0.645$, $0.668$, $0.681$; $0.440$, $0.510$, $0.616$; $0.593$, $0.629$, $0.689$.}
  \label{fig:suppl:avgtoken}
\end{figure}

\paragraph{Reading the photos together.}
The identity encoder reads all $M$ photos at once: its queries attend to the
patch tokens of every photo (Sec.~\ref{sec:suppl:impl:arch}). A simpler way to
use several photos is to encode each photo alone and average the identity tokens,
position by position. With the $24$ photos averaged this way, ArcFace falls from
$0.601$ to $0.477$ on the identity benchmark, close to the $0.467$ of a single
photo, and $99\%$ of the results score lower. The averaged tokens are also shorter
than those of the encoder (mean norm $16$ against $22$). Rescaling each of them to
the mean norm of the single-photo tokens at its position recovers about half of
the drop (ArcFace $0.534$), and $97\%$ of the results still score lower than when
the encoder reads the photos together. Face shape is also slightly worse with
averaged tokens ($s_\mathrm{gt}$ about $0.09$ against $0.083$). In
Fig.~\ref{fig:suppl:avgtoken}, the averaged tokens, rescaled or not, draw the mole
on the cheek only faintly and miss the gap between the front teeth.

\begin{figure*}[t]
  \centering
  \noindent\makebox[\textwidth][l]{\makebox[0.2948\textwidth][c]{\scriptsize Photos of the target}\hspace{0.0061\textwidth}\makebox[0.0973\textwidth][c]{\scriptsize }\hspace{0.0061\textwidth}\makebox[0.2948\textwidth][c]{\scriptsize \textbf{Ours}}\hspace{0.0061\textwidth}\makebox[0.2948\textwidth][c]{\scriptsize InSwapper}}\\[1pt]
  \noindent\makebox[\textwidth][l]{\makebox[0.0973\textwidth][c]{\scriptsize $M{=}1$}\hspace{0.0015\textwidth}\makebox[0.1960\textwidth][c]{\scriptsize added for $M{=}3$}\hspace{0.0061\textwidth}\makebox[0.0973\textwidth][c]{\scriptsize Driving}\hspace{0.0061\textwidth}\makebox[0.0973\textwidth][c]{\scriptsize $M{=}1$}\hspace{0.0015\textwidth}\makebox[0.0973\textwidth][c]{\scriptsize $M{=}3$}\hspace{0.0015\textwidth}\makebox[0.0973\textwidth][c]{\scriptsize $M{=}24$}\hspace{0.0061\textwidth}\makebox[0.0973\textwidth][c]{\scriptsize $M{=}1$}\hspace{0.0015\textwidth}\makebox[0.0973\textwidth][c]{\scriptsize $M{=}3$}\hspace{0.0015\textwidth}\makebox[0.0973\textwidth][c]{\scriptsize $M{=}24$}}\\[2pt]
  \includegraphics[width=\textwidth]{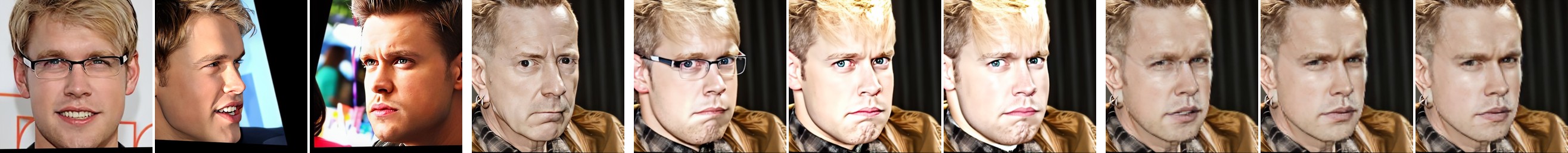}\\[0.0015\textwidth]\includegraphics[width=\textwidth]{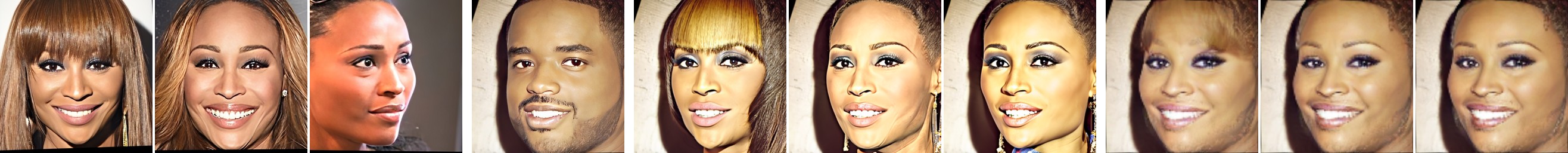}\\[0.0015\textwidth]\includegraphics[width=\textwidth]{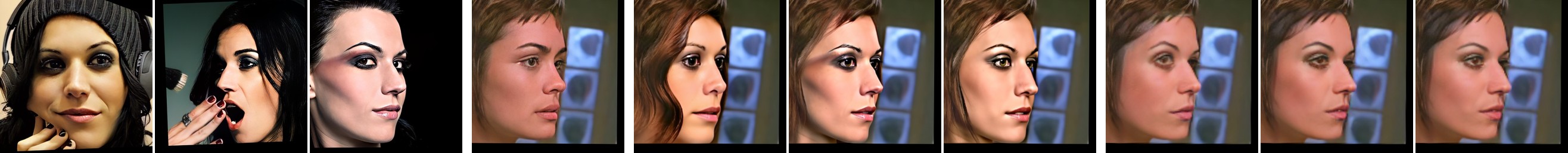}\\[0.0015\textwidth]\includegraphics[width=\textwidth]{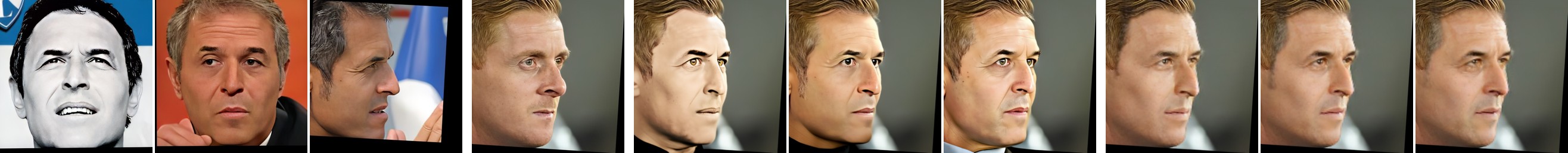}
  \caption{\textbf{Results with more photos of the target.} Four targets of the
  identity benchmark, each with one driving image. $M{=}1$ gives the first photo of
  the fixed order, $M{=}3$ adds the next two, and $M{=}24$ all of them; InSwapper
  averages its identity embedding over the same photos. The photo given at $M{=}1$
  shows, from top to bottom: glasses, a fringe, long hair under a hat, and
  black-and-white film; the photos added at $M{=}3$ do not, and these details
  leave our result. In the third row, the third photo is the only one that shows the
  side of the face seen in the driving image, and the make-up on that cheek enters
  our result at $M{=}3$.}
  \label{fig:suppl:msweepex}
\end{figure*}

\paragraph{Examples.}
Fig.~\ref{fig:suppl:msweepex} shows four targets of the identity benchmark, none
of them close to an identity of our training data
(Sec.~\ref{sec:suppl:idbench}), at $M{=}1$, $3$ and $24$. In each, the photo given
at $M{=}1$ shows something that belongs to that photo rather than to the person:
glasses, a fringe, long hair under a hat, or black-and-white film. With this photo
alone, our result takes it on as part of the identity; the two photos added at
$M{=}3$ do not show it, and it leaves the result. The added photos can also show
what the first one hides. In the third row, the third photo is the only one that
shows the side of the face seen in the driving image, and the make-up on that
cheek enters the result at $M{=}3$. InSwapper's results change less with $M$.

\begin{figure*}[t]
  \centering
  \noindent\makebox[\textwidth][l]{\makebox[0.2948\textwidth][c]{\scriptsize Photos of the target}\hspace{0.0061\textwidth}\makebox[0.0973\textwidth][c]{\scriptsize }\hspace{0.0061\textwidth}\makebox[0.2948\textwidth][c]{\scriptsize \textbf{Ours}}\hspace{0.0061\textwidth}\makebox[0.2948\textwidth][c]{\scriptsize InSwapper}}\\[1pt]
  \noindent\makebox[\textwidth][l]{\makebox[0.0973\textwidth][c]{\scriptsize $M{=}1$}\hspace{0.0015\textwidth}\makebox[0.1960\textwidth][c]{\scriptsize added for $M{=}3$}\hspace{0.0061\textwidth}\makebox[0.0973\textwidth][c]{\scriptsize Driving}\hspace{0.0061\textwidth}\makebox[0.0973\textwidth][c]{\scriptsize $M{=}1$}\hspace{0.0015\textwidth}\makebox[0.0973\textwidth][c]{\scriptsize $M{=}3$}\hspace{0.0015\textwidth}\makebox[0.0973\textwidth][c]{\scriptsize $M{=}24$}\hspace{0.0061\textwidth}\makebox[0.0973\textwidth][c]{\scriptsize $M{=}1$}\hspace{0.0015\textwidth}\makebox[0.0973\textwidth][c]{\scriptsize $M{=}3$}\hspace{0.0015\textwidth}\makebox[0.0973\textwidth][c]{\scriptsize $M{=}24$}}\\[2pt]
  \includegraphics[width=\textwidth]{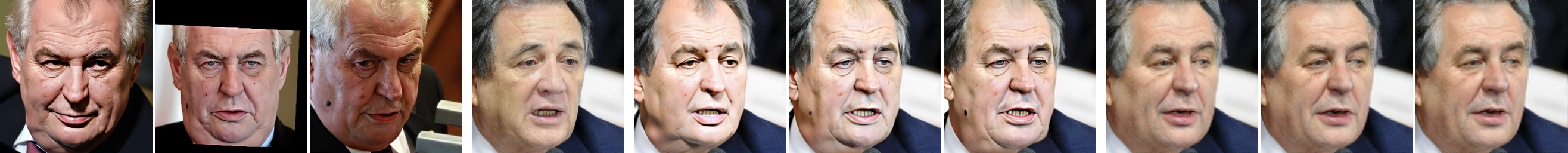}\\[0.0015\textwidth]
  \includegraphics[width=\textwidth]{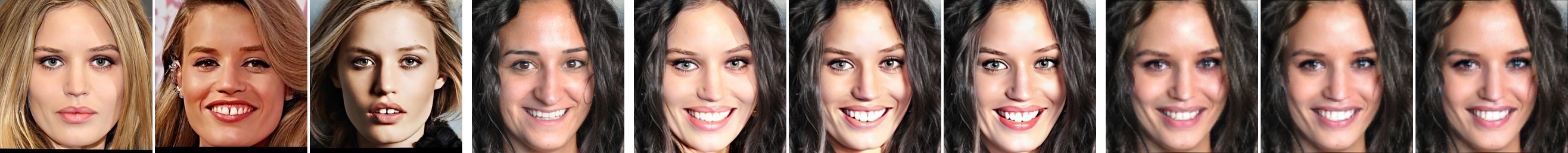}\\[0.0015\textwidth]
  \includegraphics[width=\textwidth]{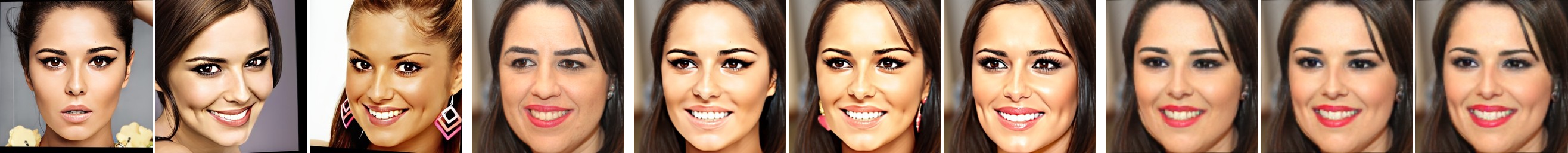}\\[0.0015\textwidth]
  \includegraphics[width=\textwidth]{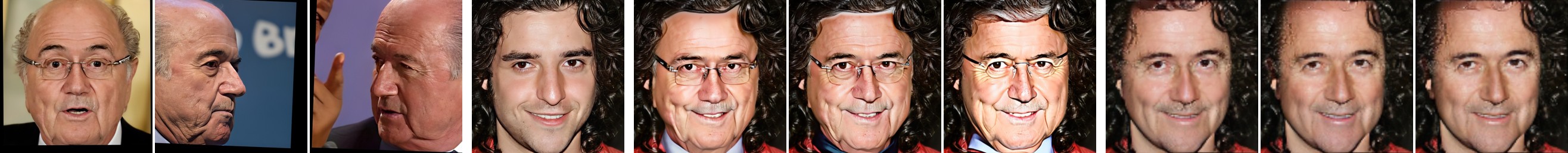}
  \caption{\textbf{Details that more photos bring.} Rows 1--3: three people of
  VGGFace2-HQ with driving images of FFHQ; their photos are ordered so that the
  photo given at $M{=}1$ hides a detail of the face and the two photos added at
  $M{=}3$ show it: a mole on the cheek, a gap between the front teeth and dimples.
  Row 4: a target of the identity benchmark, in the fixed photo order, whose
  driving image smiles. InSwapper averages its identity embedding over the same
  photos.}
  \label{fig:suppl:msweepdetail}
\end{figure*}

In Fig.~\ref{fig:suppl:msweepex}, details of the first photo leave the result as
photos are added. Fig.~\ref{fig:suppl:msweepdetail} shows details that enter it:
features of the face that the first photo hides and only the added photos show.
For three other people, we order the photos so that the first
one hides a detail of the face and the two added at $M{=}3$ show it. Our results
take on the mole and the gap between the front teeth from $M{=}3$ on, and the
dimples at $M{=}24$. InSwapper's results show the dimples at $M{=}24$ too, but not
the mole or the gap at any $M$, although its ArcFace score is higher than ours at
every $M$ in these three examples, and at $M{=}24$ higher than that of the photo
given at $M{=}1$. In the last row, an older target of the identity benchmark, the
wrinkles of the smile deepen in our results as $M$ grows, and InSwapper's results
keep a smoother face.

\section{Ablations}
\label{sec:suppl:design}
\label{tab:suppl:design}%
This section completes Sec.~\ref{sec:exp:ablations}. The first row of
Table~\ref{tab:ablation} is the model of the first phase of Stage~2; the other
Stage~2 rows continue from it for $100$k steps (Table~\ref{tab:suppl:train}).

\paragraph{Self-augmentation, split.}
Self-augmentation applies two warps to the expression input: moving least
squares on the outline of the face, then a change of the aspect ratio
(Sec.~\ref{sec:suppl:impl:strategies}). Table~\ref{tab:suppl:selfaug} trains the
full model with only one of them. Neither changes identity, expression, pose or
lighting. Each alone keeps part of the gain in face shape: the leak is $0.37$
with moving least squares only and $0.39$ with the aspect warp only, against
$0.33$ with both, and the contour distance $0.084$ and $0.085$ against $0.081$.
Both single warps differ from the full model on these two measures, and they
cannot be told apart from each other.
\begin{table}[t]
  \centering\footnotesize
  \setlength{\tabcolsep}{2.5pt}
  \caption{\textbf{Self-augmentation, split into its two warps}, on top of the
  full model, lighting off. MLS: moving least squares on the outline of the face;
  aspect: the change of the aspect ratio. Columns as in Table~\ref{tab:ablation}.}
  \label{tab:suppl:selfaug}
  \begin{tabular}{@{}l cccc ccc@{}}
    \toprule
    Warps & Id.$\uparrow$ & Expr.$\downarrow$ & Pose$\downarrow$ & Light$\uparrow$ & $s_\mathrm{gt}\downarrow$ & Leak$\downarrow$ & 3D$\downarrow$ \\
    \midrule
    Both (full model) & 0.616 & 0.256 & 3.97 & 0.66 & \textbf{0.081} & \textbf{0.33} & 1.26 \\
    MLS only & 0.615 & 0.259 & 3.96 & 0.66 & 0.084 & 0.37 & 1.27 \\
    Aspect only & 0.616 & 0.251 & 3.94 & 0.67 & 0.085 & 0.39 & 1.26 \\
    \bottomrule
  \end{tabular}
\end{table}

\paragraph{Pose and expression under a change of face shape.}
The pose and expression measures move when only the face shape of an image
changes (Sec.~\ref{sec:suppl:exprfloor}). Part of the higher pose error of the full
model than of the face mask (Table~\ref{tab:ablation}) can therefore come from the
estimator itself, as the full model changes the face shape more.

\begin{figure*}[t]
  \centering
  \edef\ucell{\the\dimexpr 0.0816\textwidth\relax}\edef\ugap{\the\dimexpr 0.0019\textwidth\relax}%
  \makebox[1.00\textwidth][l]{\makebox[\ucell][c]{\scriptsize\shortstack{Target}}\hspace{\ugap}\makebox[\ucell][c]{\scriptsize\shortstack{Driving}}\hspace{\ugap}\makebox[\ucell][c]{\scriptsize\shortstack{Ground\\truth}}\hspace{\ugap}\makebox[\ucell][c]{\scriptsize\shortstack{Face\\mask}}\hspace{\ugap}\makebox[\ucell][c]{\scriptsize\shortstack{Random\\block mask}}\hspace{\ugap}\makebox[\ucell][c]{\scriptsize\shortstack{+ self-\\augment.}}\hspace{\ugap}\makebox[\ucell][c]{\scriptsize\shortstack{+ face\\random.}}\hspace{\ugap}\makebox[\ucell][c]{\scriptsize\shortstack{+ both}}\hspace{\ugap}\makebox[\ucell][c]{\scriptsize\shortstack{Geometry\\only}}\hspace{\ugap}\makebox[\ucell][c]{\scriptsize\shortstack{Latent\\only}}\hspace{\ugap}\makebox[\ucell][c]{\scriptsize\shortstack{One pass,\\no warp}}\hspace{\ugap}\makebox[\ucell][c]{\scriptsize\shortstack{\textbf{Two passes,}\\\textbf{warp}}}}\\[2pt]
  \includegraphics[width=1.00\textwidth]{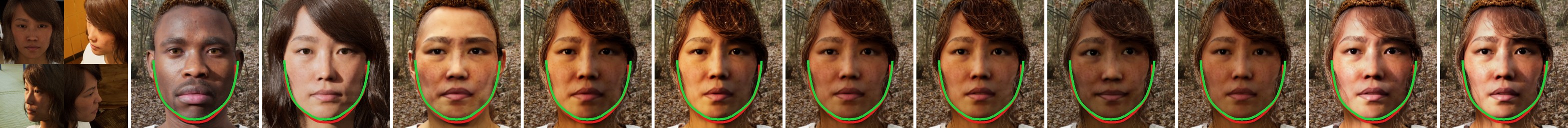}\\[\ugap]
  \includegraphics[width=1.00\textwidth]{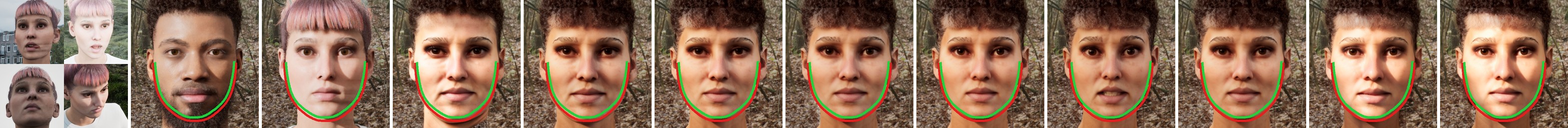}\\[\ugap]
  \includegraphics[width=1.00\textwidth]{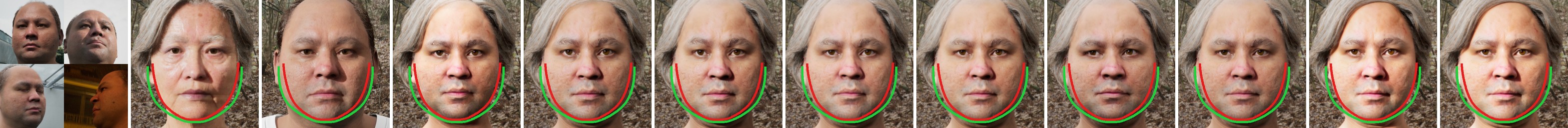}\\[\ugap]
  \includegraphics[width=1.00\textwidth]{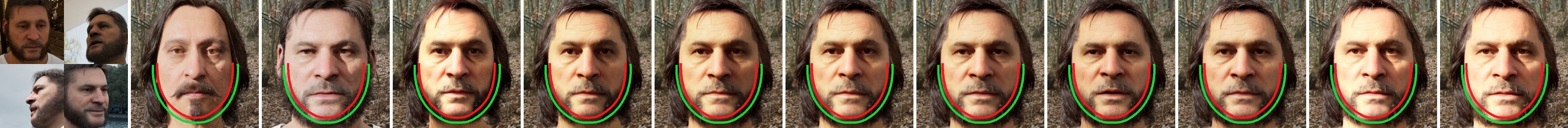}\\[\ugap]
  \includegraphics[width=1.00\textwidth]{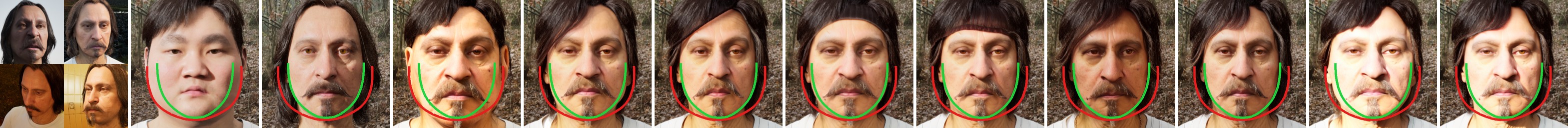}\\[\ugap]
  \includegraphics[width=1.00\textwidth]{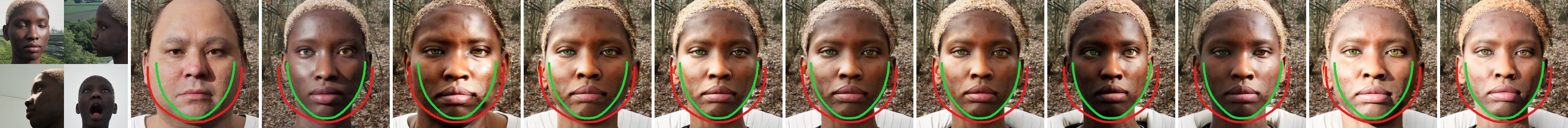}
  \caption{\textbf{Face shape in the ablations.} The pairs of Fig.~\ref{fig:suppl:ue5meth} with the variants of Table~\ref{tab:ablation}, from the fixed face mask to the full model and the two inference variants of Stage~3. Lines as in Fig.~\ref{fig:ablation}: jawline of the ground truth (green) and of the driving image (red).}
  \label{fig:suppl:ue5abl}
\end{figure*}

Fig.~\ref{fig:suppl:ue5abl} shows the pairs of Fig.~\ref{fig:suppl:ue5meth}
with the variants of Table~\ref{tab:ablation}.

\subsection{Lighting stream}
\label{sec:suppl:light}
\label{tab:suppl:light}%
The last rows of Table~\ref{tab:ablation} add the lighting stream to the full
model. In two passes with the warp, the lighting correlation rises from $0.66$
to $0.83$, ArcFace falls from $0.616$ to $0.601$, and the face shape stays about
the same (leak $0.34$ against $0.33$). In one pass without the warp, the result
follows the light a little more ($0.87$) but takes on more of the face shape of
the driving image (leak $0.60$; Fig.~\ref{fig:suppl:lightwarp}).
Sec.~\ref{sec:suppl:driving:light} shows results under strong lighting and
compares a second lighting stream that also reads the ears and neck
(Table~\ref{tab:suppl:lightstream}).

\section{Applications}
\label{sec:suppl:apps}

\subsection{Identity traversal}
\label{sec:suppl:traversal}
The identity tokens $\mC_\mathrm{id}$ carry the target's identity into the
result, so changing them changes the identity while the driving image stays
fixed. We use this to vary the apparent age of one person. On a fixed driving
image, we compute $\mC_\mathrm{id}^{A}$ from photos of the person at a younger
age and $\mC_\mathrm{id}^{B}$ from photos taken decades later, and interpolate
them linearly:
\begin{equation}
  \mC_\mathrm{id}(\tau)=(1-\tau)\,\mC_\mathrm{id}^{A}+\tau\,\mC_\mathrm{id}^{B},
  \qquad \tau\in[0,1].
\end{equation}
Every frame uses the same driving image, seed and two-pass inference
(Sec.~\ref{sec:method:inference}), so only the identity tokens change from
frame to frame.

\begin{figure*}[t]
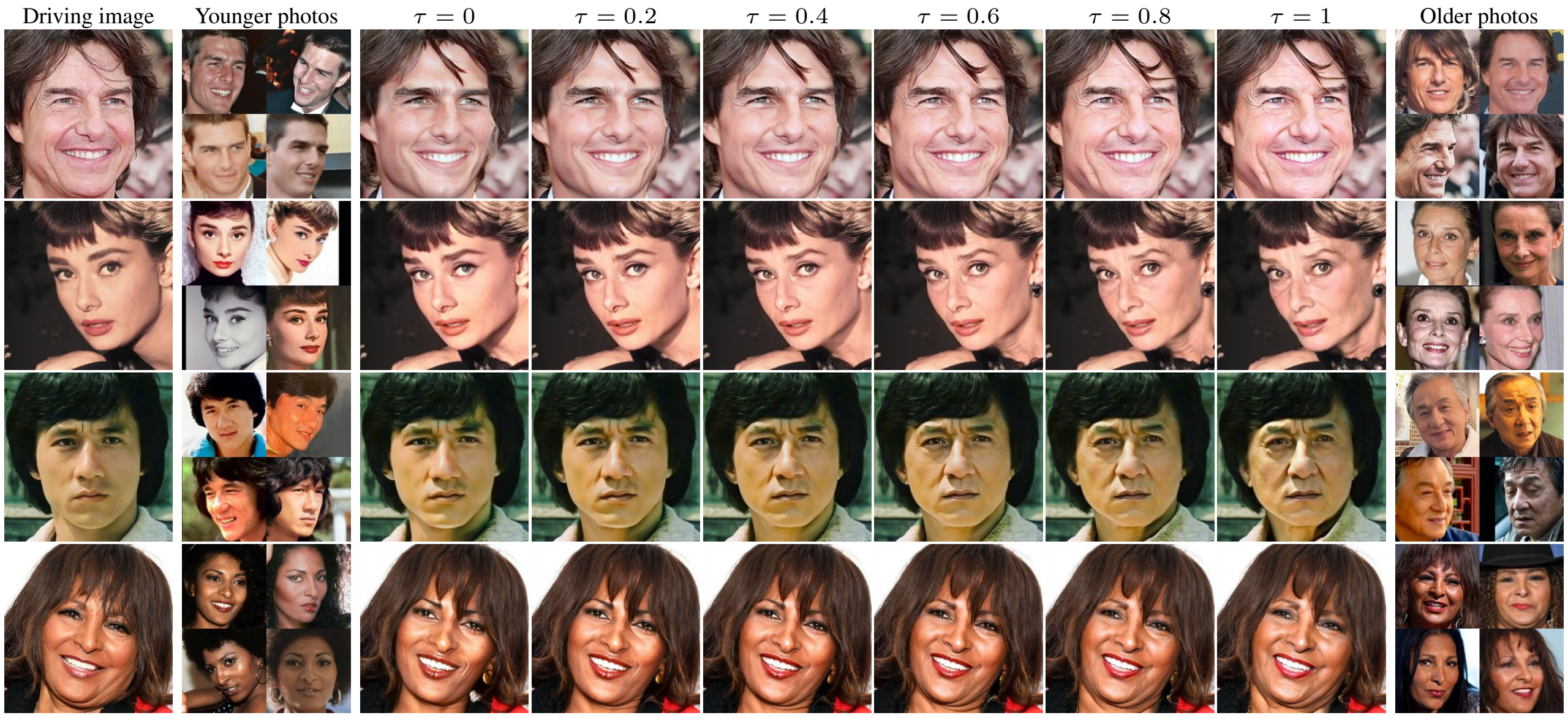

  \centering
  \def\trvw{0.104\linewidth}
  \newcommand{\trvcell}[2]{\includegraphics[width=\trvw]{figures/traversal/#1/#2.jpg}}
  \newcommand{\trvrow}[1]{\trvcell{#1}{driving} & \trvcell{#1}{young} &
    \trvcell{#1}{t0} & \trvcell{#1}{t1} & \trvcell{#1}{t2} & \trvcell{#1}{t3} &
    \trvcell{#1}{t4} & \trvcell{#1}{t5} & \trvcell{#1}{old}}
  \newcommand{\trvhead}[1]{\parbox[b]{\trvw}{\centering\scriptsize #1}}
  \setlength{\tabcolsep}{0.5pt}\renewcommand{\arraystretch}{0}
  \begin{tabular}{@{}c@{\hspace{3pt}}c@{\hspace{3pt}}cccccc@{\hspace{3pt}}c@{}}
    \trvhead{Driving image} & \trvhead{Younger photos} & \trvhead{$\tau=0$} &
    \trvhead{$\tau=0.2$} & \trvhead{$\tau=0.4$} & \trvhead{$\tau=0.6$} &
    \trvhead{$\tau=0.8$} & \trvhead{$\tau=1$} & \trvhead{Older photos} \\[2pt]
    \trvrow{t1} \\[1pt]
    \trvrow{t2} \\[1pt]
    \trvrow{t3} \\[1pt]
    \trvrow{t4} \\
  \end{tabular}
  \caption{\textbf{Identity traversal.} Each row interpolates the identity
  tokens of one person between a set of younger photos and a set of older
  photos (four of each shown), on a fixed driving image. Six of $100$ frames;
  the headers give $\tau$, rounded to one decimal.}
  \label{fig:suppl:traversal}
\end{figure*}

Fig.~\ref{fig:suppl:traversal} shows four people. Each photo set holds $5$ to
$11$ photos, and the driving image is one of the person's own photos, from one
of the two sets. As $\tau$ goes from $0$ to $1$, the apparent age and the face
shape change gradually, while the expression, head pose, hair and background stay close to
the driving image. The identity encoder also reads the ears, so earrings that
appear in several photos of one set can appear near that end (second and
fourth rows).

\subsection{Editing with the mask and a prompt}
\label{sec:suppl:apps:hair}
\begin{figure*}[t]
  \centering
  \edef\ecell{\the\dimexpr 0.1471\textwidth\relax}\edef\egap{\the\dimexpr 0.0034\textwidth\relax}%
  \makebox[0.90\textwidth][l]{\makebox[\ecell][c]{\scriptsize Target}\hspace{\egap}\makebox[\ecell][c]{\scriptsize Driving}\hspace{\egap}\makebox[\ecell][c]{\scriptsize Face}\hspace{\egap}\makebox[\ecell][c]{\scriptsize Face+hair}\hspace{\egap}\makebox[\ecell][c]{\scriptsize Hair}\hspace{\egap}\makebox[\ecell][c]{\scriptsize Whole image}}\\[2pt]
  \includegraphics[width=0.90\textwidth]{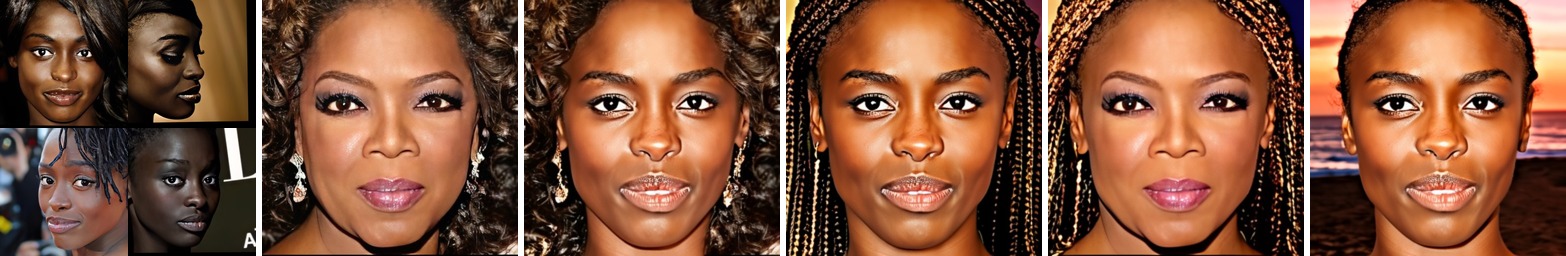}\\[\egap]
  \includegraphics[width=0.90\textwidth]{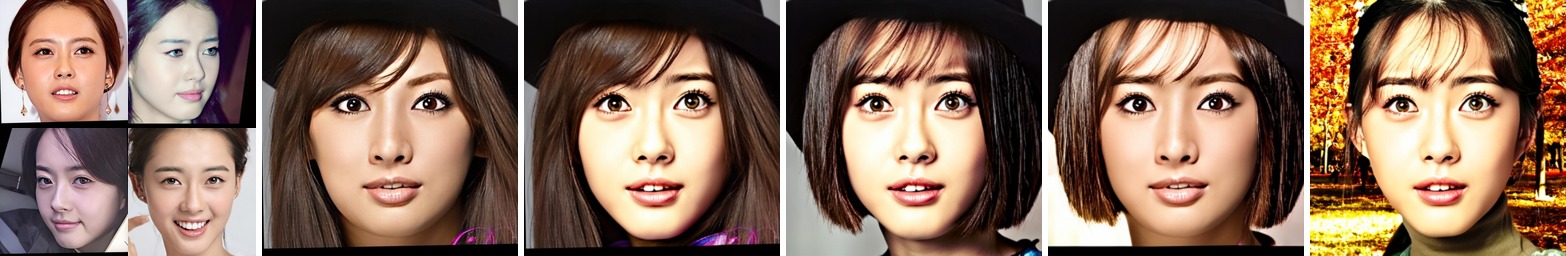}\\[\egap]
  \includegraphics[width=0.90\textwidth]{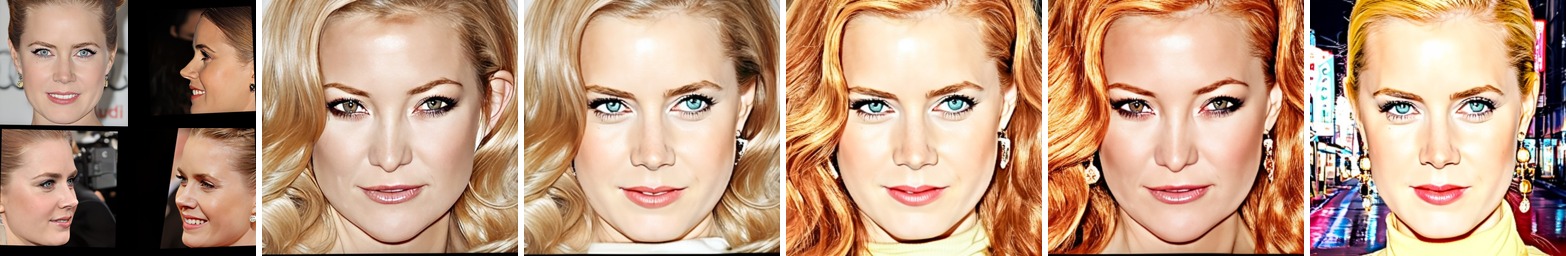}\\[\egap]
  \includegraphics[width=0.90\textwidth]{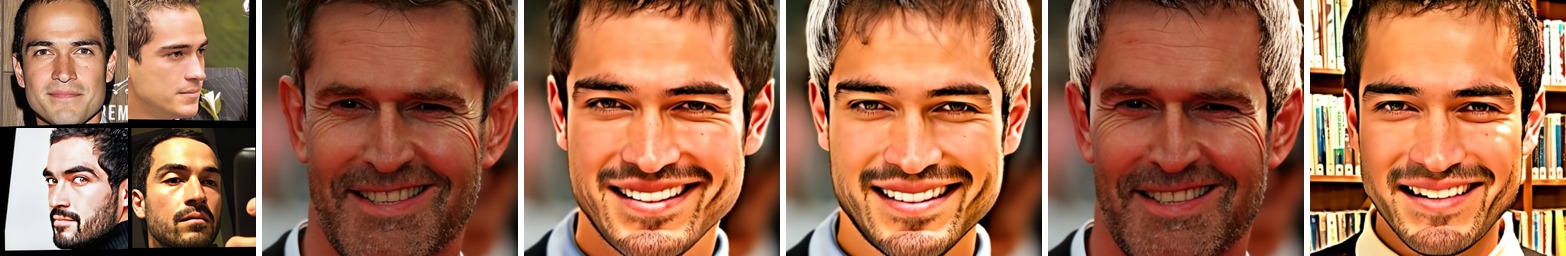}
  \caption{\textbf{Editing with the mask and a prompt.} One model, four masks. Face: the default face mask. Face+hair: with a prompt for the hair. Hair: a mask of the hair alone, with the same prompt; the face stays that of the driving image. Whole image: with a prompt for the background. Prompts, rows 1--4: braids and a beach at sunset; a short bob and an autumn park; wavy red hair and a neon-lit street at night; silver gray hair and a library. Target: four of its $24$ photos.}
  \label{fig:suppl:edits}
\end{figure*}
\begin{figure*}[t]
  \centering
  \edef\scell{\the\dimexpr 0.1167\textwidth\relax}\edef\sgap{\the\dimexpr 0.0018\textwidth\relax}\edef\sbig{\the\dimexpr 0.0055\textwidth\relax}%
  \makebox[0.95\textwidth][l]{\makebox[\scell][c]{\scriptsize Target}\hspace{\sgap}\makebox[\scell][c]{\scriptsize Driving}\hspace{\sbig}\makebox[\scell][c]{\scriptsize $1$}\hspace{\sgap}\makebox[\scell][c]{\scriptsize $2$}\hspace{\sgap}\makebox[\scell][c]{\scriptsize $4$ (default)}\hspace{\sgap}\makebox[\scell][c]{\scriptsize $6$}\hspace{\sgap}\makebox[\scell][c]{\scriptsize $12$}\hspace{\sgap}\makebox[\scell][c]{\scriptsize $24$}}\\[2pt]
  \includegraphics[width=0.95\textwidth]{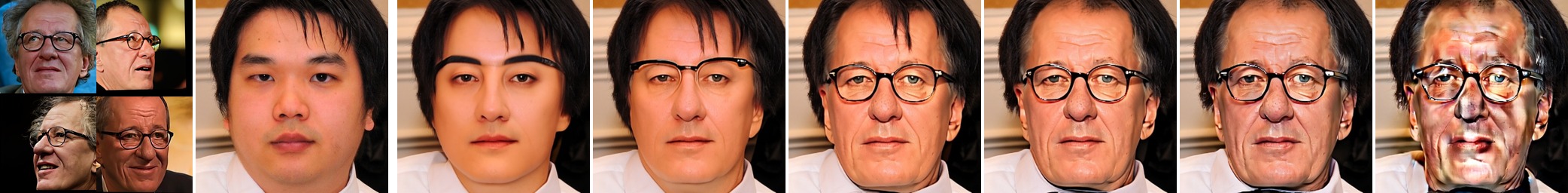}\\[\sgap]
  \includegraphics[width=0.95\textwidth]{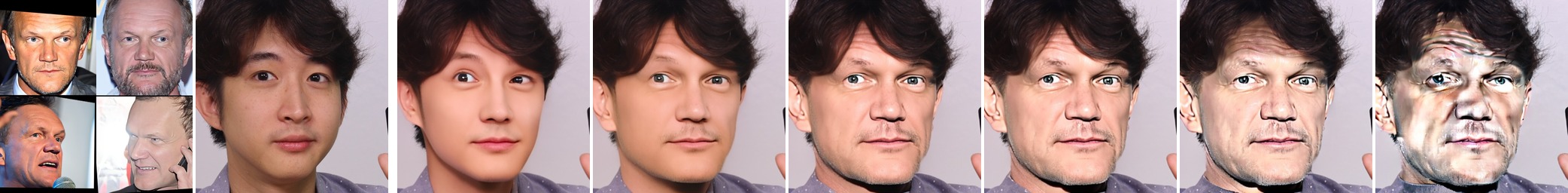}\\[\sgap]
  \includegraphics[width=0.95\textwidth]{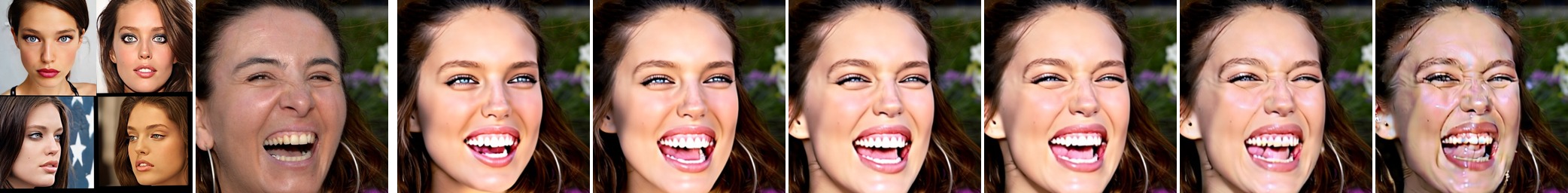}\\[\sgap]
  \includegraphics[width=0.95\textwidth]{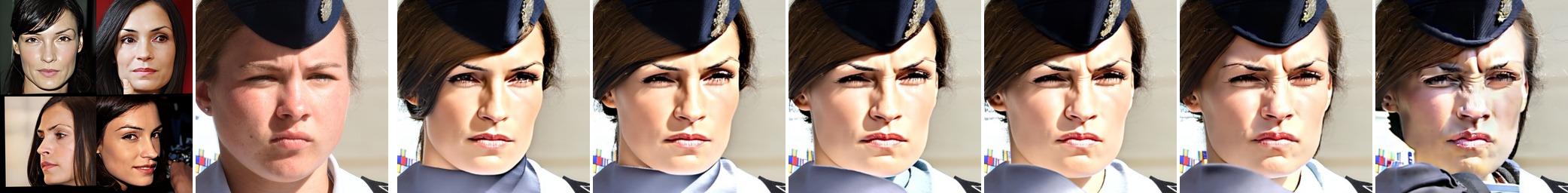}\\[\sgap]
  \includegraphics[width=0.95\textwidth]{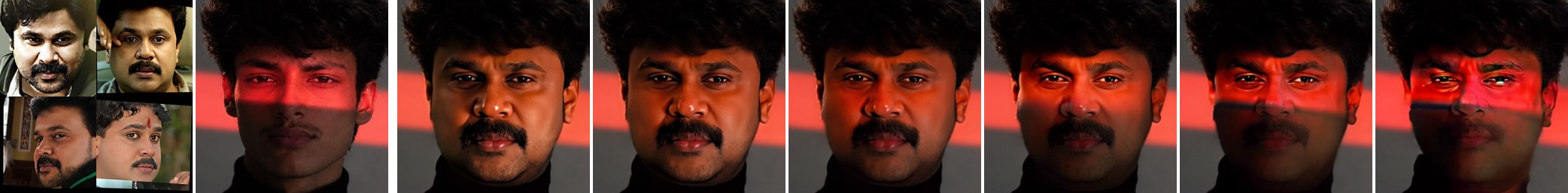}\\[\sgap]
  \includegraphics[width=0.95\textwidth]{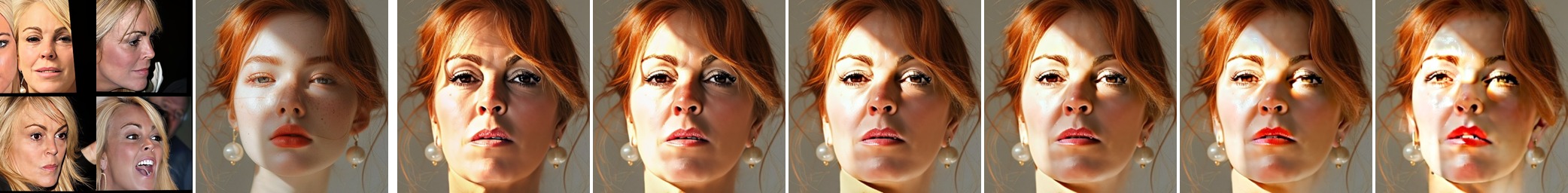}
  \caption{\textbf{Control over each stream.} Each row changes the scale of one stream, with the other two at $4$. Rows~1 and~2: identity; row~1 uses the $13$ photos of the target that show the glasses. Rows~3 and~4: expression. Rows~5 and~6: lighting. Target: four of its photos.}
  \label{fig:suppl:scale}
\end{figure*}
\begin{figure*}[t]
  \centering
  \grpgraphics[width=0.9\textwidth]{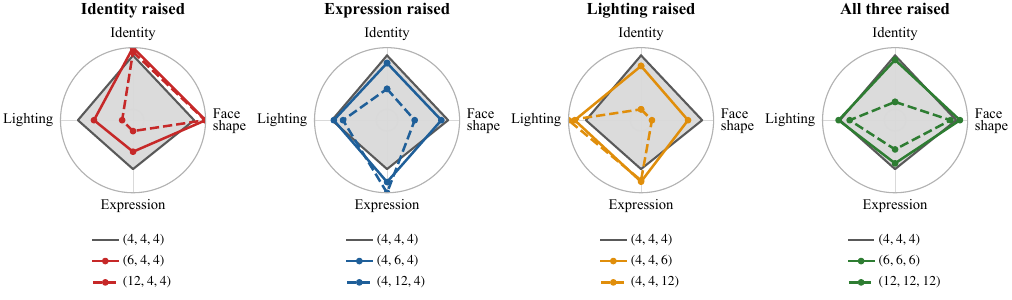}
  \caption{\textbf{Control over each stream on the benchmarks.} Each panel raises
  the scale of one stream to $6$ (solid) or $12$ (dashed), with the other two at
  $4$; the last panel raises all three. Gray: $(4,4,4)$, our default. Identity: ArcFace; face shape: contour
  distance $s_\mathrm{gt}$ on the shape benchmark; expression: EmoNet error;
  lighting: lighting correlation. Each axis is scaled linearly over the nine
  settings, from the worst value (small inner circle) to the best (outer
  circle), so small differences look large; Table~\ref{tab:suppl:scale} gives
  the values.}
  \label{fig:suppl:radar}
\end{figure*}
\begin{table*}[t]
  \centering
  \caption{\textbf{Control over each stream, measured.} Scales:
  $(w_\mathrm{id},w_\mathrm{exp},w_\mathrm{light})$. Means over the $240$ pairs
  of the identity benchmark and the $348$ pairs of the shape benchmark. Below
  each mean: the $95\%$ bootstrap interval of its difference from $(4,4,4)$,
  paired over the pairs (for Leak, over the $29$ targets). Bold: significantly
  better than $(4,4,4)$; italic: significantly worse. $\Delta L^*$: skin
  brightness of the result minus that of the driving image; closer to $0$ is
  better (Sec.~\ref{sec:limitations}).}
  \label{tab:suppl:scale}
  \scriptsize
  \setlength{\tabcolsep}{1.5pt}
  \begin{tabular}{@{}l ccccc c ccc@{}}
  \toprule
  & \multicolumn{6}{c}{Identity benchmark} & \multicolumn{3}{c}{Shape benchmark} \\
  \cmidrule(r{4pt}){2-7} \cmidrule(l{2pt}){8-10}
  Scales & ArcFace$\uparrow$ & Expr.$\downarrow$ & Pose$\downarrow$ & Light$\uparrow$ & FaceScore$\uparrow$ & $\Delta L^*$ & $s_\mathrm{gt}\downarrow$ & Leak$\downarrow$ & 3D$\downarrow$ \\
  \midrule
  $(4,4,4)$ & 0.601 & 0.253 & 3.99 & 0.833 & 4.62 & 7.8 & 0.0830 & 0.337 & 1.35 \\
  \midrule
  $(6,4,4)$ & \textbf{0.615} & \textit{0.266} & \textit{4.18} & \textit{0.811} & 4.61 & \textit{8.2} & \textbf{0.0809} & 0.327 & \textbf{1.26} \\
  & {\tiny $[+0.011,+0.016]$} & {\tiny $[+0.003,+0.023]$} & {\tiny $[+0.07,+0.33]$} & {\tiny $[-0.027,-0.017]$} & {\tiny $[-0.04,+0.02]$} & {\tiny $[+0.2,+0.7]$} & {\tiny $[-0.0039,-0.0005]$} & {\tiny $[-0.023,+0.004]$} & {\tiny $[-0.11,-0.08]$} \\[1pt]
  $(12,4,4)$ & \textbf{0.608} & \textit{0.281} & \textit{4.47} & \textit{0.773} & \textit{4.47} & \textit{8.6} & 0.0810 & 0.333 & \textbf{1.28} \\
  & {\tiny $[+0.002,+0.011]$} & {\tiny $[+0.011,+0.045]$} & {\tiny $[+0.26,+0.69]$} & {\tiny $[-0.072,-0.050]$} & {\tiny $[-0.21,-0.09]$} & {\tiny $[+0.2,+1.3]$} & {\tiny $[-0.0052,+0.0009]$} & {\tiny $[-0.033,+0.028]$} & {\tiny $[-0.11,-0.04]$} \\[1pt]
  $(4,6,4)$ & \textit{0.589} & \textbf{0.244} & \textbf{3.87} & 0.831 & \textit{4.56} & \textbf{7.0} & 0.0843 & \textit{0.377} & \textit{1.42} \\
  & {\tiny $[-0.015,-0.011]$} & {\tiny $[-0.020,-0.0003]$} & {\tiny $[-0.22,-0.01]$} & {\tiny $[-0.007,+0.001]$} & {\tiny $[-0.09,-0.03]$} & {\tiny $[-1.0,-0.6]$} & {\tiny $[-0.0008,+0.0034]$} & {\tiny $[+0.024,+0.056]$} & {\tiny $[+0.05,+0.08]$} \\[1pt]
  $(4,12,4)$ & \textit{0.544} & \textbf{0.236} & 3.83 & \textit{0.818} & \textit{4.40} & \textbf{4.8} & \textit{0.0894} & \textit{0.418} & \textit{1.62} \\
  & {\tiny $[-0.062,-0.052]$} & {\tiny $[-0.032,-0.002]$} & {\tiny $[-0.34,+0.02]$} & {\tiny $[-0.022,-0.009]$} & {\tiny $[-0.27,-0.16]$} & {\tiny $[-3.4,-2.5]$} & {\tiny $[+0.0022,+0.0107]$} & {\tiny $[+0.047,+0.117]$} & {\tiny $[+0.24,+0.31]$} \\[1pt]
  $(4,4,6)$ & \textit{0.583} & \textbf{0.244} & \textit{4.11} & \textbf{0.849} & \textit{4.57} & 7.6 & \textit{0.0857} & 0.340 & \textit{1.50} \\
  & {\tiny $[-0.021,-0.015]$} & {\tiny $[-0.016,-0.002]$} & {\tiny $[+0.02,+0.22]$} & {\tiny $[+0.011,+0.019]$} & {\tiny $[-0.07,-0.01]$} & {\tiny $[-0.4,+0.1]$} & {\tiny $[+0.0013,+0.0041]$} & {\tiny $[-0.007,+0.014]$} & {\tiny $[+0.13,+0.17]$} \\[1pt]
  $(4,4,12)$ & \textit{0.509} & 0.245 & 4.15 & \textbf{0.857} & \textit{4.33} & \textbf{7.1} & \textit{0.0925} & 0.349 & \textit{1.85} \\
  & {\tiny $[-0.098,-0.086]$} & {\tiny $[-0.022,+0.006]$} & {\tiny $[-0.03,+0.34]$} & {\tiny $[+0.017,+0.031]$} & {\tiny $[-0.35,-0.23]$} & {\tiny $[-1.3,-0.2]$} & {\tiny $[+0.0063,+0.0128]$} & {\tiny $[-0.011,+0.035]$} & {\tiny $[+0.46,+0.54]$} \\[1pt]
  $(6,6,6)$ & \textit{0.594} & 0.257 & 4.05 & 0.835 & \textit{4.51} & 7.6 & 0.0823 & 0.352 & \textit{1.40} \\
  & {\tiny $[-0.010,-0.005]$} & {\tiny $[-0.007,+0.015]$} & {\tiny $[-0.10,+0.23]$} & {\tiny $[-0.003,+0.006]$} & {\tiny $[-0.14,-0.07]$} & {\tiny $[-0.4,+0.02]$} & {\tiny $[-0.0033,+0.0019]$} & {\tiny $[-0.007,+0.041]$} & {\tiny $[+0.02,+0.06]$} \\[1pt]
  $(12,12,12)$ & \textit{0.522} & 0.268 & \textit{4.35} & \textit{0.820} & \textit{4.18} & \textbf{7.1} & 0.0841 & 0.364 & \textit{1.59} \\
  & {\tiny $[-0.085,-0.074]$} & {\tiny $[-0.003,+0.032]$} & {\tiny $[+0.12,+0.60]$} & {\tiny $[-0.021,-0.006]$} & {\tiny $[-0.49,-0.38]$} & {\tiny $[-1.1,-0.3]$} & {\tiny $[-0.0029,+0.0050]$} & {\tiny $[-0.012,+0.067]$} & {\tiny $[+0.20,+0.27]$} \\[1pt]
  \bottomrule
\end{tabular}

\end{table*}
The mask decides which parts of the image are generated
(Sec.~\ref{sec:method:inference}), and an edit prompt guides them
(Sec.~\ref{sec:suppl:impl:inference}). Fig.~\ref{fig:suppl:edits} uses one model
with four masks. With the face mask, our default, only the face changes. With the
face+hair mask and a prompt for the hair, the hair also follows the prompt, as in
Fig.~\ref{fig:teaser}. With a mask of the hair alone, only the hair is generated,
and the face stays that of the driving image. With a mask of the whole image and a
prompt for the background, the scene is generated anew, while the expression and
pose still follow the driving image. The hair and clothes then come from the photos
of the target, and the light on the face still follows the driving image, so the
new background changes it only a little. The prompts take the form of
Sec.~\ref{sec:suppl:impl:inference}, for example ``a photorealistic portrait of a
human face with braids'' or ``a photorealistic portrait of a human face on a beach
at sunset''.

\subsection{Control over each stream}
\label{sec:suppl:apps:scale}
Identity, expression and lighting enter the model through separate streams, and
each stream is dropped on its own in training (Sec.~\ref{sec:suppl:impl:train}).
Each stream can therefore have its own guidance scale at inference. With the
scales $w_\mathrm{id}$, $w_\mathrm{exp}$ and $w_\mathrm{light}$, the prediction is
\begin{equation}
  \vv=\vv_\mathrm{u}+4\,(\vv_\mathrm{c}-\vv_\mathrm{u})
  +\sum_{s}(w_s-4)\,(\vv_\mathrm{c}-\vv_{\mathrm{c}\setminus s}),
  \label{eq:suppl:scale}
\end{equation}
where $s$ runs over the three streams and $\vv_{\mathrm{c}\setminus s}$ is the
conditional prediction with stream $s$ dropped, as in training. With all three
scales at $4$, this is the guidance of Sec.~\ref{sec:suppl:impl:inference}. The
first pass has no lighting stream, so only the identity and expression terms
apply there. The three scales give an artist one control per stream. A VFX shot
may call for more of the actor's facial detail, a stronger performance or a
closer match to the light on set, and the artist can weigh these against each
other for each shot.

Fig.~\ref{fig:suppl:scale} changes one scale at a time, with the other two at
$4$. A higher identity scale makes the features of the target more pronounced,
such as the glasses and the wrinkles in row~1 and the lines on the forehead in
row~2. In row~2, the results at $4$ and $6$ both look like the target, and the
artist can choose how strongly these features show. A higher expression scale
makes the laugh stronger (row~3). In row~4, the driving image squints with
knitted brows: the lines between the brows appear at $4$ and deepen at $6$, and
at $12$ the lines and the squint are stronger than in the driving image. A
higher lighting scale makes the red band of light across the eyes sharper
(row~5), while in row~6 the default already follows the sunlight. At $24$, the
results degrade in every row.

Table~\ref{tab:suppl:scale} and Fig.~\ref{fig:suppl:radar} measure these
controls on the identity benchmark ($240$ pairs) and the shape benchmark ($348$
pairs). Each scale moves the measures of its own stream. At $6$, ArcFace rises
from $0.601$ to $0.615$ and the contour distance falls from $0.0830$ to
$0.0809$. The expression error falls from $0.253$ to $0.244$, and the skin
brightness moves closer to that of the driving image
(Sec.~\ref{sec:limitations}). The lighting correlation rises from $0.833$ to
$0.849$. Each gain costs some of the other streams. At $12$, the costs grow,
and the identity gain falls back (ArcFace $0.608$). Raising all three scales
together does not add up the gains. At $(6,6,6)$, the results stay close to
those of $(4,4,4)$, with small losses such as ArcFace $0.594$. At
$(12,12,12)$, most measures are worse (ArcFace $0.522$). None of the eight
settings is at least as good as $(4,4,4)$ on every measure. Our default of $4$
for all three streams keeps the measures balanced, and the artist can move away
from it where a shot needs it.

\section{Limitations and Broader Impact}
\label{sec:limitations}
\begin{figure}[t]
  \centering
  \includegraphics[width=\linewidth]{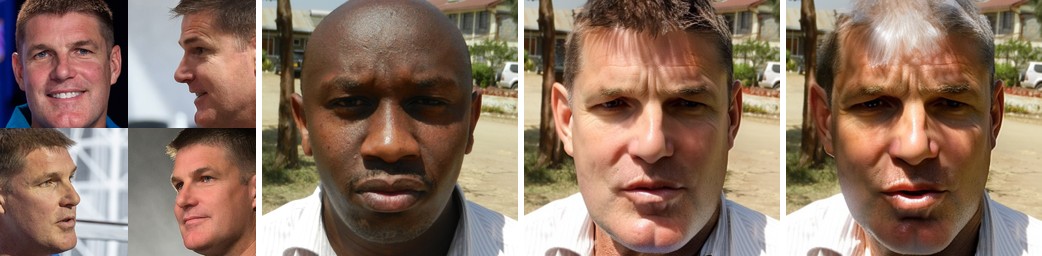}\\[1pt]
  \makebox[0.25\linewidth][c]{\scriptsize Target}%
  \makebox[0.25\linewidth][c]{\scriptsize Driving}%
  \makebox[0.25\linewidth][c]{\scriptsize Ours, lighting off}%
  \makebox[0.25\linewidth][c]{\scriptsize Ours}\\[4pt]
  \includegraphics[width=\linewidth]{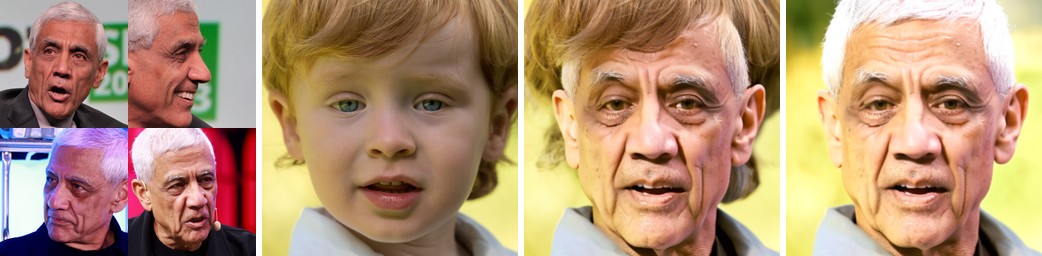}\\[1pt]
  \makebox[0.25\linewidth][c]{\scriptsize Target}%
  \makebox[0.25\linewidth][c]{\scriptsize Driving}%
  \makebox[0.25\linewidth][c]{\scriptsize Ours, face mask}%
  \makebox[0.25\linewidth][c]{\scriptsize Ours, face+hair mask}\\[4pt]
  \includegraphics[width=\linewidth]{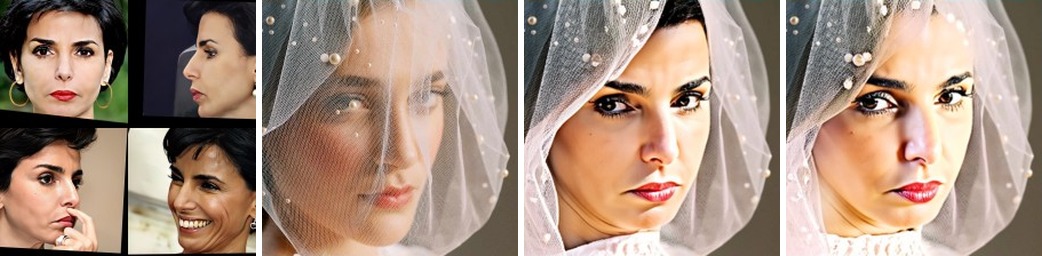}\\[1pt]
  \makebox[0.25\linewidth][c]{\scriptsize Target}%
  \makebox[0.25\linewidth][c]{\scriptsize Driving}%
  \makebox[0.25\linewidth][c]{\scriptsize Ours, lighting off}%
  \makebox[0.25\linewidth][c]{\scriptsize Ours}
  \caption{\textbf{Limitations.} \emph{Top:} skin tone and lighting. A target
  with light skin and a driving image of a person with dark skin in sunlight.
  Without the lighting stream (pass~1 of the inference), our result keeps the
  target's skin tone under flat light. With it, the result takes on the light of
  the driving image, and also part of its skin tone. \emph{Middle:} hair comes
  from the driving image. With the face mask, our result keeps the child's hair of
  the driving image under the target's face. With the face+hair mask, the hair is
  generated: here it is white like the target's, but the identity stream does not
  see hair, so it is not transferred. \emph{Bottom:} a semi-transparent veil. It
  is not detected as an occluder, so it lies inside the mask. Our model generates
  the face without the veil, already in pass~1, while the part of the veil outside
  the mask stays.}
  \label{fig:suppl:limits}
\end{figure}

\paragraph{Limitations.}
MYSTIQUE processes each driving image on its own: it has no temporal model, and
extending it to video is future work.
Our results make the skin lighter than in the driving
image in $88\%$ of the pairs of the identity benchmark, by $7.8$ in $L^*$
(CIELAB) on average, while those of InSwapper, SimSwap and Face-Adapter stay
within $1$ of it. The offset is about the same for every skin tone, so the results
of dark-skinned targets end up furthest from the target ($+21$ in Individual
Typology Angle for the darkest third of the targets, against $+7$ for the
lightest). It is also present without the lighting stream. A lower guidance
scale reduces it: on $63$ of these pairs, from a scale of $4$ to $2$, it falls
from $7.6$ to $4.0$ at a small cost in identity (ArcFace $-0.014$). In addition,
the
lighting stream cannot fully separate skin tone from lighting
(Fig.~\ref{fig:suppl:limits}, top), and at times it over-exposes the face and hides
fine texture details.
The identity stream
never sees hair. Hair is therefore kept from the driving image, even where it
does not suit the target (Fig.~\ref{fig:suppl:limits}, middle), or, with the
face+hair mask, generated rather than transferred.
A semi-transparent veil over the face is neither an opaque occluder nor glasses.
If it is not detected as an occluder, our model generates the face without the
veil, while the part of the veil outside the mask stays
(Fig.~\ref{fig:suppl:limits}, bottom).
Finally, the shape benchmark
scores only the jawline of rendered characters, which is one aspect of
face shape on faces that are not real people.

\paragraph{Broader impact.}
Identity transfer can be misused for non-consensual imagery, disinformation and the
misappropriation of a person's likeness, and a method that renders identity
more faithfully increases these risks. MYSTIQUE is intended for visual effects in
film and television production with the consent of the people depicted.

\end{document}